\documentclass[10pt]{article} % For LaTeX2e
\usepackage[accepted]{tmlr}
\usepackage[utf8]{inputenc} % allow utf-8 input
\usepackage[T1]{fontenc}    % use 8-bit T1 fonts
\usepackage{hyperref}       % hyperlinks
\usepackage{url}            % simple URL typesetting
\usepackage{booktabs}       % professional-quality tables
\usepackage{amsfonts}       % blackboard math symbols
\usepackage{nicefrac}       % compact symbols for 1/2, etc.
\usepackage{microtype}      % microtypography
\usepackage{colortbl}
\usepackage[table,xcdraw]{xcolor}
\usepackage{subcaption}
\usepackage{wrapfig}
\usepackage{multicol}
\usepackage{multirow}
\usepackage{soul}

\usepackage{bbm}
\usepackage{pifont}
\newcommand{\cmark}{\ding{51}}%
\newcommand{\xmark}{\ding{55}}%

\usepackage{array}
\usepackage{listings}
\definecolor{vscodebg}{rgb}{0.15, 0.16, 0.18}      % Background (dark gray)
\definecolor{vscodefg}{rgb}{0.85, 0.85, 0.85}      % Foreground (light gray)
\definecolor{vscodekeyword}{rgb}{0.86, 0.43, 0.84} % Keywords (purple)
\definecolor{vscodecomment}{rgb}{0.13, 0.55, 0.13} % Comments (green)
\definecolor{vscodestring}{rgb}{0.94, 0.81, 0.54}  % Strings (yellowish)
\definecolor{vscodenumber}{rgb}{0.56, 0.83, 0.98}  % Numbers (light blue)
\usepackage{amsmath}
\usepackage{amssymb}
\usepackage{mathtools}
\usepackage{amsthm}

\usepackage{enumitem}
\usepackage{wrapfig}
\usepackage{amsmath,amsfonts,bm}

\def\eqref#1{equation~\ref{#1}}
\def\1{\bm{1}}

\DeclareMathAlphabet{\mathsfit}{\encodingdefault}{\sfdefault}{m}{sl}
\SetMathAlphabet{\mathsfit}{bold}{\encodingdefault}{\sfdefault}{bx}{n}

\usepackage{url}

\title{ToMoE: Converting Dense Large Language Models to Mixture-of-Experts
through Dynamic Structural Pruning}

\author{\name Shangqian Gao
      \thanks{Corresponding author.} 
      \email sg24bi@fsu.edu \\
      \addr Department of Computer Science, Florida State University
      \AND
      \name Hua Ting \email thua@nd.edu \\
      \addr Department of Computer Science and Engineering, University of Notre Dame
      \AND 
      \name Reza Shirkavand \email rezashkv@cs.umd.edu\\
      \addr Department of Computer Science, University of Maryland, College Park
      \AND 
      \name Chi-Heng Lin \email chiheng.lin@samsung.com\\
      \addr Samsung Research America
      \AND 
      \name Zheng Tang \email zheng.tang@samsung.com\\
      \addr Samsung Research America
      \AND 
      \name Zhengao Li \email zl23i@fsu.edu\\
      \addr Department of Computer Science, Florida State University
      \AND 
      \name Longge Yuan \email ly23a@fsu.edu\\
      \addr Department of Computer Science, Florida State University
      \AND 
      \name Fangyi Li \email fangyili@seas.upenn.edu\\
      \addr School of Engineering and Applied Science, University of Pennsylvania
      \AND 
      \name Zeyu Zhang \email zeyzhan@amazon.com\\
      \addr Amazon AGI
      \AND 
      \name Alireza Ganjdanesh \email aliganj@umd.edu\\
      \addr Department of Computer Science, University of Maryland, College Park
      \AND
      \name Lou Qian \email qian.lou@ucf.edu\\
      \addr Department of Computer Science, University of Central Florida
      \AND
      \name Xu Jie \email xujie@ufl.edu\\
      \addr Department of Health Outcomes and Biomedical Informatics, University of Florida
      \AND 
      \name Yen-Chang Hsu \email yenchang.hsu@samsung.com\\
      \addr Samsung Research America      
}
\newcommand{\revise}[1]{\textcolor{black}{#1}}
\def\month{01}  % Insert correct month for camera-ready version
\def\year{2026} % Insert correct year for camera-ready version
\def\openreview{\url{https://openreview.net/forum?id=RFHq46pjb6}} % Insert correct link to OpenReview for camera-ready version

\begin{document}
\definecolor{color0}{HTML}{fbb4ae}
\definecolor{color1}{HTML}{b3cde3}
\definecolor{color2}{HTML}{ccebc5}
\definecolor{color3}{HTML}{decbe4}
\definecolor{color4}{HTML}{fed9a6}
\definecolor{color5}{HTML}{ffffcc}
\definecolor{color6}{HTML}{e5d8bd}
\definecolor{color7}{HTML}{fddaec}
\definecolor{tabcolor}{RGB}{224, 224, 255}
\maketitle
%{\noindent\footnotesize $^{\ast}$Corresponding author.}
\begin{abstract}
Large Language Models (LLMs) demonstrate remarkable capabilities but face deployment challenges due to their high computational demands. Traditional pruning methods reduce these costs by permanently removing parameters, which inevitably leads to performance degradation. To mitigate this issue, we propose ToMoE, a method that transforms dense LLMs into Mixture-of-Experts (MoE) models by uncovering experts inherently present within dense models, without requiring any weight updates. ToMoE leverages dynamic structural pruning to unify expert construction and router training in a single stage, achieving consistently strong performance. Remarkably, even without fine-tuning \revise{the model weights}, ToMoE consistently outperforms state-of-the-art pruning and MoE techniques across Phi-2, LLaMA-2, LLaMA-3, and Qwen-2.5 models. The code for this paper is available at \url{https://github.com/gaosh/ToMoE}.
\end{abstract}

\section{Introduction}
\label{sec:intro}
% Large Language Models (LLMs) have emerged as powerful tools in artificial intelligence, exhibiting remarkable capacity in natural language understanding and generation. By leveraging advanced neural architectures~\cite{vaswani2017attention}, and pretraining on vast corpora of text, these models have demonstrated a remarkable ability to perform diverse tasks, ranging from question answering and machine translation to content creation and scientific discovery~\cite{brown2020language, kenton2019bert,raffel2020exploring}. Examples of popular commercial or non-commercial LLMs include ChatGPT~\cite{openai2022chatgpt}, Claude~\cite{anthropic2023claude}, and LLaMA \cite{touvron2023llama}. The core strength of LLMs lies in their ability to capture complicated relationships within the context, enabling them to generalize across tasks with in-context learning~\cite{dong2022survey} or minimal fine-tuning~\cite{radford2019gpt2, kaplan2020scaling}.
% \ting{1st para is optional, we can start directly from 2nd}

Although LLMs demonstrate remarkable capacity to perform diverse tasks~\citep{brown2020language, kenton2019bert,raffel2020exploring,openai2022chatgpt,anthropic2023claude,dong2022survey,radford2019gpt2,kaplan2020scaling}, their huge model size often limits their usability on devices with limited resources. As a result, considerable efforts~\citep{ma2023llm,ashkboos2023slicegpt, frantar2022gptq} are focused on minimizing the computational and memory costs of these models. Structural pruning~\citep{ma2023llm} has emerged as a promising solution to this challenge because, unlike unstructured pruning, it achieves compression without the need for specialized implementations. However, the problem with structural pruning methods is that they will substantially reduce the model capacity, resulting in an obvious performance gap compared to the dense model. The fine-tuning cost for even partially recovering this gap is tremendous.

To achieve a better trade-off between the number of parameters and performance, sparse Mixture of Experts (MoE) models~\citep{Shazeer2017, lepikhingshard} are designed to activate only a subset of the model's parameters, corresponding to the selected experts. Recently proposed MoE models, such as DeepseekMoE~\citep{dai2024deepseekmoe}, demonstrated that they can match the performance of dense models with a similar total parameter count while using a small number of active parameters. Following this lead, transforming dense models into MoE models could offer a promising approach to bridging the performance gap left by structural pruning methods. Unlike prior efforts to construct MoE models from dense models~\citep{zhang2022moefication,lee2024breaking,zhu2024llama}, our findings reveal that MoE inherently exists within dense models and can be uncovered without updating model weights (continue pretraining). Specifically, we show that these experts can be identified through dynamic structural pruning. These results represent a novel contribution that has not been demonstrated in previous studies. 
%\ting{Let ToMoE to be our goal, not pruning. Pruning is our method to achieve our goal. }

\begin{figure*}[t]
	\subfloat[Original LLMs]{
	\begin{minipage}[b]{.3\linewidth}
			\centering
			\includegraphics[width=.99\textwidth]{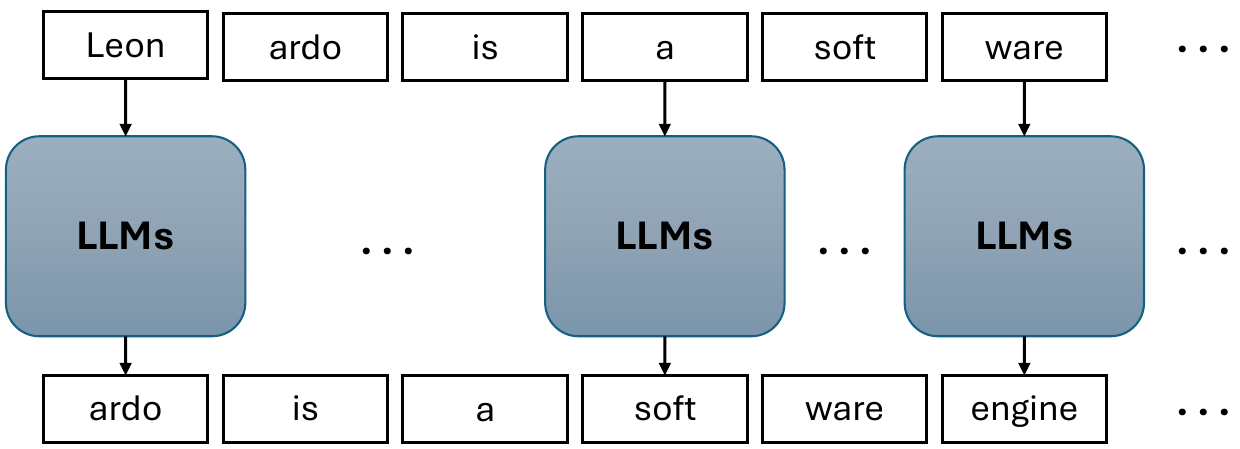}
	\end{minipage}}
        \hfill
	\subfloat[LLMs with structural pruning.]{
		\begin{minipage}[b]{.3\linewidth}
			\centering
			\includegraphics[width=.99\textwidth]{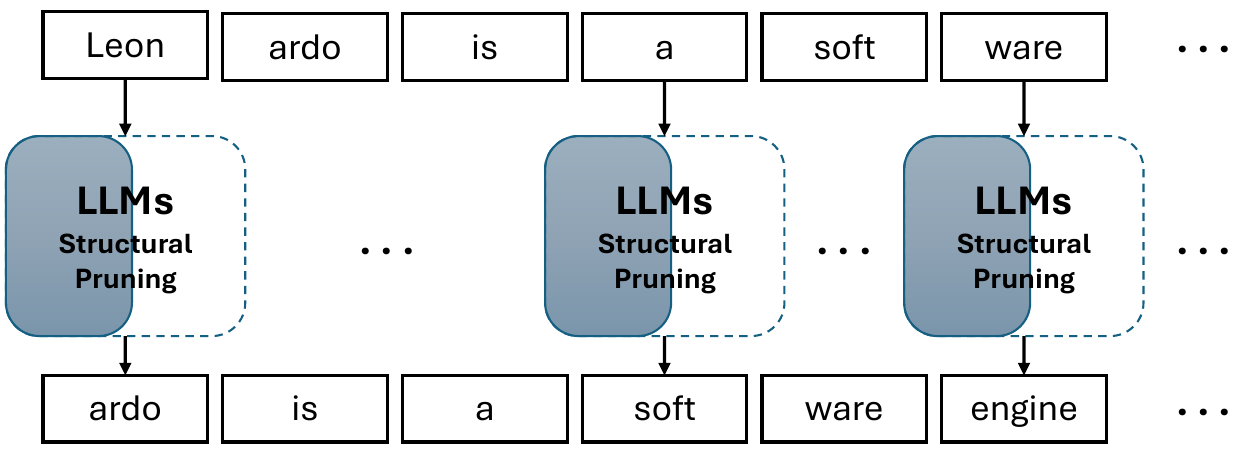}
	\end{minipage}}
        \hfill
	\subfloat[LLMs with dynamic pruning for MoE.]{
		\begin{minipage}[b]{.3\linewidth}
			\centering
			\includegraphics[width=.99\textwidth]{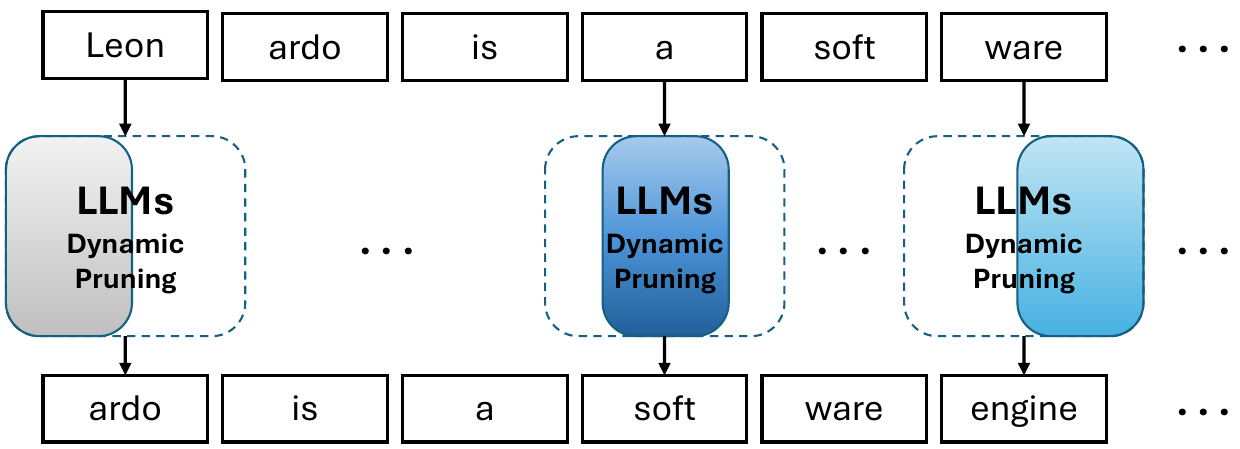}
	\end{minipage}}
	\caption{\textbf{(a):} The original LLM uses all parameters to process all the input text. \textbf{(b):} The static pruning for LLMs permanently removes model parameters, and the rest of the parameters are used to process all the input text. \textbf{Our approach (c):} LLMs with dynamic pruning use different sub-networks (illustrated by different colors) to process different tokens. We incorporate MoE to achieve a fixed expected budget in inference. }
    \label{fig:tomoe-concept}
    \vspace{-10pt}
\end{figure*}

The core idea of MoE models is conditional computation, where experts are dynamically selected based on input tokens. This concept aligns closely with dynamic pruning methods~\citep{gao2018dynamic}, which make pruning decisions given input features. Leveraging this connection, we propose to construct MoE models from dense models by using dynamic structural pruning. Specifically, for Multi-Head self-Attention (MHA) layers, we apply top-K routing and static pruning for compression, while for MLP layers, we transform them into MoE layers using top-1 expert routing. The routing mechanism learned for dynamic structural pruning can be directly applied to serve as the routing module for MoE layers. With differentiable discrete operations, the MoE conversion process can be formulated as a differentiable dynamic pruning problem. With this formulation, we can efficiently convert a dense model to an MoE model at a cost similar to or lower than regular structural pruning methods. The comparison between our method, static pruning, and the original LLM is shown in Fig.~\ref{fig:tomoe-concept}. 

Built upon the above findings and techniques, we proposed ToMoE to effectively convert dense LLMs to MoE models with dynamic pruning. The contributions of this work can be summarized as follows:
\begin{itemize}[nosep]
\item \textbf{Dense-to-MoE Conversion Through Dynamic Pruning:} 
    We introduce a novel approach to convert dense models into MoE models through dynamic pruning. Specifically, we implement top-K routing and static pruning for MHA layers along the head dimension and top-1 routing for MLP layers across the learned experts. This formulation ensures sparse and efficient computation while retaining model capacity.

 \item \textbf{Joint Optimization for Routing and Experts:} 
    The proposed method involves jointly optimizing routing modules and expert configurations by solving a regularized optimization problem. Our approach leverages differentiable operations to enable efficient and flexible MoE constructions.

%  \revise{by only training the router} without fine-tuning the model weights. 
\item \textbf{Consistent Performance Improvements:} 
    Our method consistently outperforms state-of-the-art structural pruning and MoE construction techniques on various tasks \revise{while training only the router}, without fine-tuning the model weights. This performance improvement is demonstrated across widely used public models such as Phi-2, LLaMA-2, LLaMA-3, and Qwen-2.5.
\item \revise{\textbf{Detailed Analysis:} We extensively analyze the resulting model from {ToMoE} across multiple perspectives, including parameter allocation, router behavior, and the ablation of different design components. We hope these analyses provide valuable insights and guidance for future research in this area.
}
\end{itemize}
%\vspace{-5pt}
\section{Related Works}
%\vspace{-5pt}

%pruning
%losparse,dejavu,h2O
%\ting{Any related work trying to convert dense to MOE? if not, add we are the first} there are previous methods shown in second paragraph of the related works\\
\textbf{Pruning:} Structural pruning~\citep{li2017pruning, bert-surgeon, ma2023llm} is an attractive technique for real-world deployments since it removes redundant parameters to reduce model size without requiring specialized implementations. 
%%% Static & Dynamic Pruning (RS)
Structural pruning methods fall into two main categories: static pruning~\citep{cnn-static-pruning,molchanov2019importance,fang2023depgraph} and dynamic pruning~\citep{gao2018dynamic,chen2020dynamic,dynamic-context-pruning,kv-cache-compression}. Static pruning removes parameters based on input-agnostic importance metrics. For example, LLM-Pruner~\citep{ma2023llm} eliminates non-essential coupled structures using gradient-based criteria. The problem with structural pruning is that it often creates a noticeable performance gap relative to dense models~\citep{ma2023llm,ashkboos2023slicegpt}. In contrast, dynamic pruning removes weights based on input-dependent metrics. Early attempts for dynamic pruning~\citep{gao2018dynamic,chen2020dynamic} focus on Convolutional Neural Networks, where channels are selectively activated for input samples. Recent works, such as D-LLM~\citep{d-llm}, incorporate the concept of conditional computation into LLMs by selectively skipping layers based on input tokens. The problem with dynamic pruning methods is that they do not have a fixed budget given different inputs, which creates problems when serving LLMs in a mini-batch setting or in the prefilling stage. Our method, on the other hand, converts the dense LLM to a sparse MoE model with a fixed per-token budget.

Another line of research applies contextual sparsity for LLMs~\citep{pmlr-v202-liu23am,zheng2024learn,lee2024cats}, where neurons are selectively activated given certain conditions. Although there are some promising results, they are generally more difficult to achieve better inference efficiency due to their irregular sparsity patterns. In contrast, MoE models have more comprehensive support from the system side, making them a popular choice for scaling up the model. Thus, our method mainly focuses on converting dense models to MoE models.

%For instance, Wanda~\citep{sun2024a} prunes LLMs by considering the product of weight magnitudes and input activations, while D-LLM~\citep{d-llm} selectively skips layers depending on input tokens.
%%%

%However, by removing entire components or connected structures, pruning often creates a noticeable performance gap relative to dense models~\citep{ma2023llm,ashkboos2023slicegpt}. The LLM Surgeon~\citep{van2023llm} attempts to mitigate this issue by periodically updating both the model's weights and structures during training, but it still does not fully bridge the gap with dense models. Instead of simply discarding parameters, our method repurposes pruned structures as experts, effectively merging the pruning and expert construction into a single step, and performs better than existing structural pruning approaches.

%MoE

\textbf{MoE:} Sparse Mixture-of-Experts (MoE) models improve upon pure structural pruning by maintaining or even enhancing model capacity without a proportional increase in computational cost. For instance, Sparsely-Gated MoE~\citep{Shazeer2017} employs a trainable gating network to select a small subset of experts for each input, enabling the model to scale to thousands of experts efficiently~\citep{lepikhingshard}. More recent methods like DeepSeekMoE~\citep{dai2024deepseekmoe} further address expert specialization, matching dense-model performance with a similar number of activated parameters. Previous methods constructing MoE from the dense model~\citep{zhang2022moefication,lee2024breaking,zhu2024llama} separate the expert construction and router training into two distinct stages, often leading to sub-optimal performance. In contrast, our method integrates expert construction directly into the pruning process, treating it as a unified step with router learning and thereby largely improving the performance without fine-tuning.

\vspace{-5pt}
\section{ToMoE}
\vspace{-5pt}
\begin{figure*}[t]
	\centering
    \includegraphics[width=.9\textwidth]{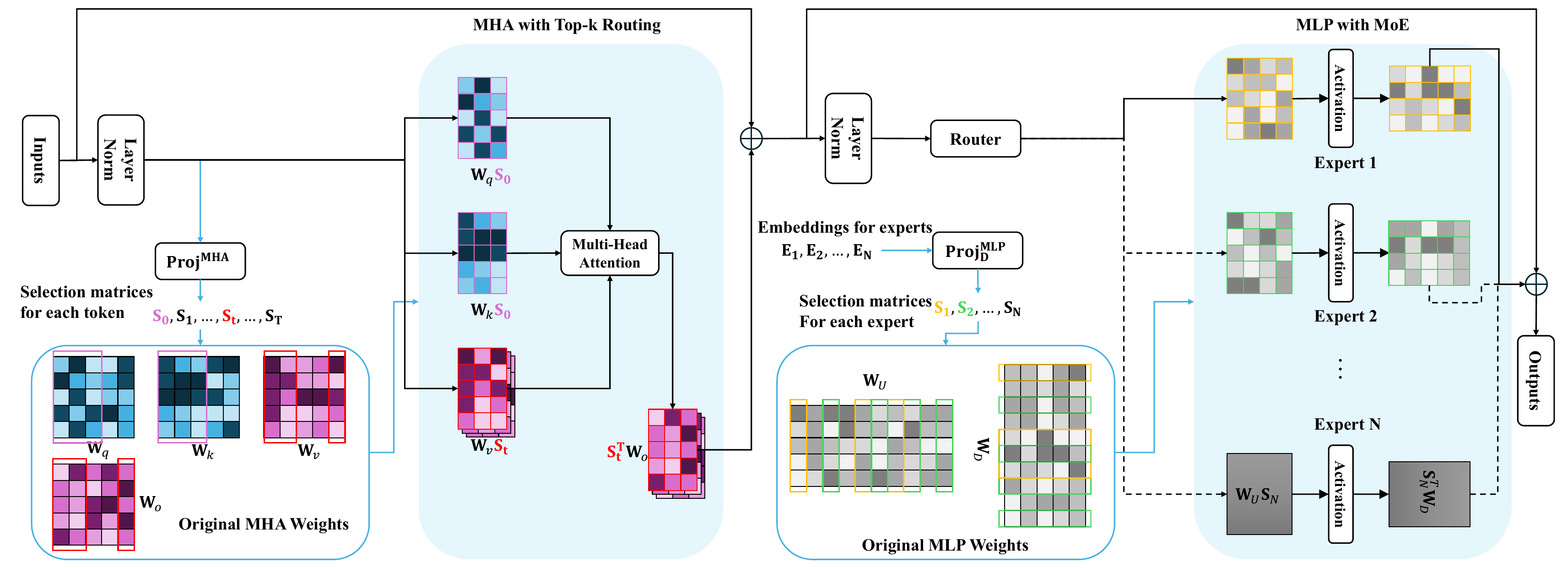}
    \caption{\textbf{ToMoE} uses top-1 routing for MLP layers, and static and dynamic pruning along the head dimension for MHA layers.
    }
  \label{fig:tomoe}
  \vspace{-5pt}
\end{figure*}

{Most recent LLMs, like GPT~\citep{radford2018gpt}, LLaMA~\citep{touvron2023llama}, etc., adapt decoder-only architectures and thus our method focuses on decoder-only architectures. A typical decoder block consists of Multi-Head Attention (MHA) and Multi-Layer Perceptron (MLP) layers. %For clarity, we let $T$, $d$, $d_{\text{mid}}$, and $H$ denote the sequence length, hidden dimension,  MLP middle layer dimension, and the number of attention heads, respectively. 
For clarity, we denote the sequence length by $T$, the hidden dimension by $d$, the MLP intermediate dimension by $d_{\text{mid}}$, and the number of attention heads by $H$.

To reduce the computational costs of the decoder-only architecture, we propose to convert the original model into MoE models. For MHA layers, we utilize top-K routing and static pruning along the head dimension $\frac{d}{H}$. Top-K routing and static pruning for MHA layers ensure that, during prefilling or model serving, all tokens maintain the same head dimension, enabling parallel processing. For MLP layers, our approach transforms them into MoE layers along the MLP middle dimension $d_{\text{mid}}$ and employs top-1 routing.
\label{sec:arch_basics}
}A key distinction between our method and previous dynamic pruning approaches is that the converted model maintains consistent computational costs for all inputs. This property could be crucial for efficient processing.

\subsection{Expert Embeddings}
Inspired by the recent success of using hypernetworks~\citep{ha2016hypernetworks,ganjdanesh2024not,gaodisp} to generate pruning decisions, we adopt a hypernetwork to generate expert embeddings:
\begin{equation}~\label{eq:hn}
    \mathbf{E}_{\text{all}} = \text{HN(z)},
\end{equation}
where $z$ is the input to the hypernetwork drawn from a random distribution, and $\mathbf{E}_{\text{all}} = [\mathbf{E}_1,\cdots,\mathbf{E}_l, \cdots, \mathbf{E}_L]$ contains embeddings for all layers and $\mathbf{E}_l \in \mathbb{R}^{N\times d_e}$, where $N$ is the number of experts and $d_e$ is the expert embedding dimension. Each embedding $\mathbf{E}_{l,i}$ will then be used to generate the configurations of experts. The purpose of having the hypernetwork to generate $\mathbf{E}_{\text{all}}$ is to introduce inter-layer dependencies across different layers and operations. This design has been shown to accelerate the learning process in practice~\citep{gaodisp}. More details are given in the Appendix~\ref{sec:app-modules}.

\subsection{Expert Construction \label{subsec:EC}}
In this section, we will talk about how to construct experts from MLP layers. In a decoder layer, the formulation of MLP is: $f_{\text{MLP}}(\mathbf{X}) = 
\sigma(\mathbf{X}\mathbf{W}_G)\odot (\mathbf{X}\mathbf{W}_U) \mathbf{W}_D$, where matrices $\mathbf{W}_U \in \mathbb{R}^{d \times d_{\text{mid}}}$, $\mathbf{W}_G \in \mathbb{R}^{d \times d_{\text{mid}}}$ and $\mathbf{W}_D \in \mathbb{R}^{d_{\text{mid}} \times d}$ denote up, gated, and down projection matrices. In addition, $\sigma$ denotes nonlinear activation functions and $\odot$ denotes the Hadamard product (element-wise product). 

Assume the target is to use $N$ experts, under the setting of structural pruning, each expert can be represented by:
\begin{equation}~\label{eq:mlp-moe}
f_{\text{MLP}}^{i}(\mathbf{X}_t)=\sigma(\mathbf{X}_t\mathbf{W}_G \mathbf{S}_i)\odot (\mathbf{X}_t\mathbf{W}_U\mathbf{S}_i) \mathbf{S}_i^{\top}\mathbf{W}_D,
\end{equation}
where $i = 1,\cdots,N$, and $\mathbf{S}_i  = \text{Diag}(\mathbf{s}_i)$ ($\mathbf{s}_i\in \mathbb{R}^{d_{\text{mid}}}$, $\mathbf{S}_i\in \mathbb{R}^{d_{\text{mid}}\times d_{\text{mid}}}$), is a binary diagonal matrix selecting a subset of weight vectors for the $i$th expert. $\mathbf{X}_t$ is the $t$th token, which is assumed to be routed to the $i$th expert. 
Once each expert is formulated, its configuration is learned as follows:
\begin{equation}~\label{eq:mlp}
    \mathbf{s} = {\text{ST‑GSig}}(\text{Proj}_{\text{D}}^{\text{\tiny MLP}}(\mathbf{G}\mathbf{E})), \ \mathbf{G} = {\text{ST‑GSmax}}(\text{Router}(\mathbf{X})),
\end{equation}
where $\mathbf{G} \in \mathbb{R}^{T \times N}$ is the output of the router module, $\mathbf{E}$ ($l$ is omitted for clarity) is the expert embeddings, $\text{Proj}_{\text{D}}^{\text{\tiny MLP}}: \mathbb{R}^{d_e} \to \mathbb{R}^{d_{\text{mid}}} $ is a projection module to project the latent embedding to the MLP middle dimension, \revise{$\text{Router}(\cdot): \mathbb{R}^{d_e} \to \mathbb{R}^{N}$ is the router module that maps the inputs to an $N$-dimensional routing score vector for expert selection}, and $\text{ST‑GSig}$ and $\text{ST-GSmax}$ are Straight-Through Gumbel-Sigmoid and Gumbel-Softmax functions respectively~\citep{jang2016categorical}. Under this setting, $\mathbf{s}_i$ will contain retained positions (represented by $1$) for the $i$th expert, and $\mathbf{G}$ contains one-hot routing decisions for tokens in $\mathbf{X}$.  

\subsection{MHA top-K Routing\label{subsec:top-k}}
% An MHA layer can be represented as:
% $f_{\text{MHA}}(\mathbf{X}) = \sum_{i=1}^H \sigma_s\left({e}(\mathbf{X}\mathbf{W}_{Q,i}){e}^\top(\mathbf{X}\mathbf{W}_{K,i})\right)\mathbf{X}\mathbf{W}_{V,i}{\mathbf{W}_{O,i}}$, where $\mathbf{W}_{Q,i}, \mathbf{W}_{K,i}, \mathbf{W}_{V,i} \in \mathbb{R}^{d\times \frac{d}{H}}, \mathbf{W}_{O,i} \in \mathbb{R}^{\frac{d}{H}\times d}$ are the query, key, value, and output matrices for each attention head, and $\mathbf{X}\in\mathbb{R}^{T\times d}$ is the input hidden states.
An MHA layer can be represented as 
$f_{\text{MHA}}(\mathbf{X}) = \sum_{i=1}^H \sigma_s\left(e(\mathbf{X}\mathbf{W}_{Q,i}) e^\top(\mathbf{X}\mathbf{W}_{K,i})\right) \mathbf{X}\mathbf{W}_{V,i} \mathbf{W}_{O,i}$, 
where $\mathbf{W}_{Q,i}, \mathbf{W}_{K,i}, \mathbf{W}_{V,i} \in \mathbb{R}^{d \times \frac{d}{H}}$, $\mathbf{W}_{O,i} \in \mathbb{R}^{\frac{d}{H} \times d}$ are the query, key, value, and output matrices for each attention head, and $\mathbf{X} \in \mathbb{R}^{T \times d}$ is the input hidden states.
${e}$ and $\sigma_s$ denote positional embedding and the softmax function.

For MHA layers, we perform two kinds of pruning: dynamic top-K pruning and static pruning, both along the head dimension. Justifications regarding the design choice are provided in the Appendix~\ref {sec:app-mha}. Like MLP layers, we also insert selection matrices:
% \begin{equation}~\label{eq:MHA}
% \begin{split}
%     f_{\text{MHA}}(\mathbf{X}_t) &= \sum_{i=1}^H [\sigma_s\left({e}(\mathbf{X}_t\mathbf{W}_{Q,i})\mathbf{S}_0\mathbf{S}_0^{\top}{e}^\top(\mathbf{X}_t\mathbf{W}_{K,i})\right)\\
%     &\mathbf{X}_t\mathbf{W}_{V,i}\mathbf{S}_t]\mathbf{S}_t^{\top}{\mathbf{W}_{O,i}}
% \end{split}
% \vspace{-5pt}
% \end{equation}
\begin{equation}\label{eq:MHA}
f_{\text{MHA}}(\mathbf{X}_t) = \sum_{i=1}^H \left[\sigma_s\left(e(\mathbf{X}_t\mathbf{W}_{Q,i})\mathbf{S}_0\mathbf{S}_0^{\top}e^\top(\mathbf{X}_t\mathbf{W}_{K,i})\right)\mathbf{X}_t\mathbf{W}_{V,i}\mathbf{S}_t\right]\mathbf{S}_t^{\top}\mathbf{W}_{O,i},
%\vspace{-5pt}
\end{equation}
%where $\mathbf{S}_0\in \mathbb{R}^{\frac{d}{H}\times\frac{d}{H}}$ is the shared selection matrix for static pruning of query and key matrices and $\mathbf{S}_t\in \mathbb{R}^{\frac{d}{H}\times\frac{d}{H}}$ is the selection matrix for the value and output matrices of the $t$th token.
where $\mathbf{S}_0, \mathbf{S}_t \in \mathbb{R}^{\frac{d}{H}\times\frac{d}{H}}$ are selection matrices. $\mathbf{S}_0$ is the shared selection matrix for static pruning of query and key matrices, while $\mathbf{S}_t$ is the token-specific selection matrix for the value and output matrices of the $t$-th token. We apply the same selection matrix across all heads, ensuring that all heads have the same head dimensions at inference time.
To generate the selection matrix, we calculate its diagonal vector $\mathbf{s}_t$ as:
\begin{equation}~\label{eq:MHA-STGS}
    \mathbf{s}_t = {\text{ST‑GSig}}(\text{Proj}_{\text{D}}^{\text{\tiny MHA}}(\text{Proj}_{\text{E}}^{\text{\tiny MHA}}(\mathbf{X}_t)+\frac{1}{N}\mathbf{1}^{\top}\mathbf{E})),
\end{equation}
where $\mathbf{1} \in \mathbb{R}^{N}$ is an all-one vector, $\frac{1}{N}\mathbf{1}^{\top}\mathbf{E}$ represents the average expert embedding of size $d_e$, $\text{Proj}_{\text{D}}^{\text{\tiny MHA}}: \mathbb{R}^{d_e} \to \mathbb{R}^{\frac{d}{H}} $ is a projection module to map the latent embedding to the head dimension and $\text{Proj}_{\text{E}}^{\text{\tiny MHA}}: \mathbb{R}^{d} \to \mathbb{R}^{d_e} $ is also a projection module to project input tokens to the space of expert embeddings, and $\text{ST‑GSig}$ is defined in Sec.~\ref{subsec:EC}. When $t=0$, we initialize $\mathbf{X}=0$, and set $\mathbf{s}_0 = {\text{ST‑GSig}}(\text{Proj}_{\text{D}}^{\text{\tiny MHA}}(\frac{1}{N}\mathbf{1}^{\top}\mathbf{E}))$, since it is input independent. 

During training, the number of ones in $\mathbf{s}$ can vary freely. After training is complete, we compute $K=\text{round}(\frac{1}{T}\sum_{t=1}^T\sum_{i=1}^{\frac{d}{H}}\mathbf{s}_{t,i})$ for a subset of tokens and use it for top-K routing during inference. Note that the $k$ in top-K for $\mathbf{s}_0$ and $\mathbf{s}_t$ ($t\geq1$) can be different, allowing for larger flexibility. Also, note that $\mathbf{s}_0$ must follow specific structural constraints to be compatible with the position embedding $e(\cdot)$, and more details can be found in Appendix~\ref{sec:app-rope}.

\subsection{Regularizations for MoE Constructions}~\label{sec:reg}
In Sec.~\ref{subsec:EC} and Sec.~\ref{subsec:top-k}, we briefly introduced the design space for constructing MoE models using dynamic structural pruning. In this subsection, we will introduce regularizations customized to the characteristics of MoE models.
%to address the specific requirements of MoE models.

\textbf{Union of Experts Regularization.} An ideal sparse MoE model converted from a dense model should maximize parameter utilization, which means that the total number of parameters in the MoE model should closely approximate the dense model. To add this regularization to our learning process, we push the union of experts to be closer to the original model. More specifically, we use MHA layers as an example:
\begin{equation}~\label{eq:union}
    \mathbf{u} = \bigcup_{i=1}^T \mathbf{s}_i = 1- \prod_{i = 1}^{T}(1-\mathbf{s}_i),
\end{equation}
where $\bigcup$ is the union operator, and $\mathbf{u}$ is the union of all kept positions for each token. For MLP layers, it can be calculated similarly. We then push $ \frac{\sum\mathbf{u}}{|\mathbf{u}|}$ ($|\mathbf{u}|$ represents the size of $\mathbf{u}$) to 1:
\begin{equation}~\label{eq:ur}
    \mathcal{R}_{\text{U}} = \frac{1}{L}\sum_{l=1}^L f_{\text{reg}}( \frac{\sum\mathbf{u}_l}{|\mathbf{u}_l|}, 1),
\end{equation}
where $f_{\text{reg}}(\cdot,\cdot)$ can be any regression loss functions, and we will choose $f_{\text{reg}}$ later.

\textbf{Parameter Regularization.} For a sparse MoE model, we also need to control the number of active parameters given the provided budget. To achieve this goal, we can directly penalize the maximum width across different experts. We choose the maximum width over experts instead of the mean, median, or other alternatives because the maximum provides precise control over the upper bound of the number of active parameters.

Denote the width of a layer  as $d_l^{*}$, where $*\in\{\text{MLP},\text{MHA}\}$. For MLP layers, it can be calculated by $d_l^{\text{ MLP}} = \max(\mathbf{s}\mathbf{1}_{d_{\text{mid}}})$, where $\mathbf{1}_{d_{\text{mid}}} \in \mathbb{R}^{d_{\text{mid}}}$ is an all-one vector of size $d_{\text{mid}}$. $\mathbf{s}\mathbf{1}_{\text{mid}}$ produces the width of all experts, and $d_l^{\text{ MLP}}$ represents the maximum width across all experts. The width of MHA layers can be calculated similarly. Based on $d_l^*$, we can calculate the number of active parameters in the model $\text{T}(\mathbf{d_{\text{MoE}}})$, where $\mathbf{d_{\text{MoE}}} = [d_1^{*},\cdots,d_L^{*}]$. To push the number of active parameters to a predefined rate $p$, the following objective is applied:
\begin{equation}~\label{eq:pr}
    \mathcal{R}_{\text{P}} = f_{\text{reg}}(\text{T}(\mathbf{d_{\text{MoE}}}), p\text{T}_{\text{total}}),
\end{equation}
where $\text{T}_{\text{total}}$ is the total number of parameters, and $p\in (0,1]$ represents the ratios of the active parameters. For $f_{\text{reg}}$ in Eq.~\ref{eq:pr} and Eq.~\ref{eq:ur},  the following function $f_{\text{reg}}$ is used:
\begin{equation*}
    f_{\text{reg}}(x,y) = \log(\max(x ,y) / \min(x ,y)).
\end{equation*}

\textbf{Load Balancing Regularization.} When determining the configurations of experts, we also apply the load balancing regularization to encourage a balanced load across experts~\citep{lepikhin2021gshard,fedus2022switch}. 
The load balancing loss from the Switch Transformer~\citep{fedus2022switch} is adopted:
\begin{equation}~\label{eq:lb}
    \mathcal{R}_{\text{L}} = N\sum_{i=1}^NF_iP_i,
\end{equation}
where $F_i=\frac{1}{T}\sum_{t=1}^T\mathbbm{1}(\mathbf{G}_{t,i}=1)$. The indicator function $\mathbbm{1}(\cdot)$ returns
$1$ if the condition is true and $0$ otherwise. $F_i$ represents the fraction of tokens assigned to the $i$-th expert. $P_i =\frac{1}{T}\sum \text{GSmax}({G}_{t,i})$, where ${G} = \text{Router}(\mathbf{X})$ is the router output \textbf{before ST‑GSmax}. $P_i$ is the fraction of the router probability allocated for the $i$-th expert. $\mathcal{R}_{\text{L}}$ will encourages uniform routing across different experts as shown in~\citep{fedus2022switch}. The combination of Eq.~\ref{eq:pr} and Eq.~\ref{eq:lb} creates an interesting phenomenon where they encourage uniform allocation of width among experts.
\begin{table*}[t]
\centering
\caption{
    Perplexity comparisons of structured pruning methods and ToMoE for LLaMA-2 7B and 13B models on WikiText-2.
    \label{tab:lm}
}
\setlength{\extrarowheight}{2pt} % Adds space between rows
\resizebox{0.7\linewidth}{!}{
\begin{tabular}{lcccccc}
\toprule
\multirow{2}{*}{\textbf{Method}} 
& \multicolumn{3}{c}{LLaMA-2 7B (ppl: 5.12 \textcolor{red}{$\downarrow$})} 
& \multicolumn{3}{c}{LLaMA-2 13B (ppl: 4.57 \textcolor{red}{$\downarrow$})} \\
\cmidrule(lr){2-4} \cmidrule(lr){5-7}
& \textbf{70\%} & \textbf{60\%} & \textbf{50\%} 
& \textbf{70\%} & \textbf{60\%} & \textbf{50\%} \\
\midrule
LLM-Pruner \citep{ma2023llm}
& 13.56 & 17.90 & \multicolumn{1}{c|}{31.05}
& 12.19 & 19.56 & 32.20 \\
LLM Surgeon \citep{van2023llm}
& {7.83} & {10.39} & \multicolumn{1}{c|}{{15.38}}
& {6.21} & {7.25} & {9.43} \\
%\midrule
ShortGPT \citep{men2024shortgpt}
& 33.21 & 71.04 & \multicolumn{1}{c|}{268.11}
& 30.48 & 48.83 & 187.23 \\
SLEB \citep{songsleb}
& 11.23 & 29.10 & \multicolumn{1}{c|}{103.38}
& 8.24 & 11.76 & 27.67 \\
SliceGPT \citep{ashkboos2023slicegpt}
& 10.47 & 15.19 & \multicolumn{1}{c|}{24.82}
& 8.68 & 12.56 & 20.57 \\
ModeGPT \citep{lin2024modegpt}
& 7.51 & 8.41 & \multicolumn{1}{c|}{11.88}
& 6.10 & 6.95 & 8.95 \\
DISP-LLM \citep{gaodisp}
& 6.85 & 8.11 & \multicolumn{1}{c|}{9.84}
& 5.77 & 6.59 & 7.11 \\
%\rowcolor{blue!5!white}
\rowcolor{tabcolor}\textbf{ToMoE (ours)}
& \textbf{6.41} & \textbf{7.17} & \multicolumn{1}{c|}{\textbf{8.36}}
& \textbf{5.54} & \textbf{6.06} & \textbf{6.78} \\
\bottomrule
\end{tabular}
}
%\vspace{-10pt}
\end{table*}

%ashkboos2023slicegpt
%men2024shortgpt
\subsection{Learning to Construct MoEs}
Based on the aforementioned techniques,  MoEs can be constructed from the dense LLMs by training router parameters, projection parameters, and hypernetwork parameters, while keeping all original model parameters frozen. This approach enables the rapid construction of an effective MoE model with a resource budget comparable to that of structural pruning. The overall framework of our method is shown in Fig.~\ref{fig:tomoe}, and the corresponding training objective function can be formulated as:
\begin{equation}~\label{Eq:all}
    \min_{\Theta}\ \mathcal{L}(f'(x;\mathbf{E}_{\text{all}}),f(x)) +\alpha \mathcal{R}_{\mathbf{P}} + \beta \mathcal{R}_{\mathbf{U}} + \gamma \mathcal{R}_{\mathbf{L}},
\end{equation}
% \begin{table}[t]
% \centering
% \caption{Comparisons with semi-structured pruning on LLaMA-2. \label{tab:Semi}}
% \setlength\extrarowheight{2pt} % Adds extra row height for better readability
% \resizebox{0.95\linewidth}{!}{ % Adjust table size to fit within the page
%     \renewcommand{\arraystretch}{1} % Reset row spacing
%     \begin{tabular}{|l|c|cc|}
%     \hline
%     \multicolumn{1}{|c|}{\textbf{Method}} & \textbf{Structure} & \textbf{50\% (7B)} & \textbf{50\% (13B)} \\ \hline
%     SparseGPT (2:4)       & Semi-structured & {10.17} & 8.32       \\
%     Wanda (2:4)           & Semi-structured & 11.02          & 8.27       \\
%     Pruner-Zero (2:4)      & Semi-structured & 10.52          & 7.41       \\ 
%     \hline
%     ToMoE (ours)      & Structured & \textbf{8.36}          & \textbf{6.78}       \\ 
%     \hline
%     \end{tabular}
% }
% \end{table}
\begin{wraptable}{r}{0.5\linewidth}
\centering
\caption{Comparisons with semi-structured pruning on LLaMA-2.}
\label{tab:Semi}
\setlength\extrarowheight{2pt}
\resizebox{0.95\linewidth}{!}{
    \renewcommand{\arraystretch}{1}
    \begin{tabular}{|l|c|cc|}
    \hline
    \multicolumn{1}{|c|}{\textbf{Method}} & \textbf{Structure} & \textbf{50\% (7B)} & \textbf{50\% (13B)} \\ \hline
    SparseGPT (2:4)       & Semi-structured & 10.17 & 8.32       \\
    Wanda (2:4)           & Semi-structured & 11.02 & 8.27       \\
    Pruner-Zero (2:4)     & Semi-structured & 10.52 & 7.41       \\ 
    \hline
    ToMoE (ours)          & Structured      & \textbf{8.36} & \textbf{6.78}       \\ 
    \hline
    \end{tabular}
}
\vspace{-10pt}
\end{wraptable}

where $\Theta = [\Theta_{\text{HN}}, \Theta_{\text{Router}}, \Theta_{\text{Proj-MHA}}, \Theta_{\text{Proj-MLP}}]$, $\Theta_{\text{HN}}$ is trainable parameters for the hypernetwork in Eq.~\ref{eq:hn}, $\Theta_{\text{Router}}$ and $\Theta_{\text{Proj-MLP}}$ are trainable parameters for the router and the project module in Eq.~\ref{eq:mlp}, $\Theta_{\text{Proj-MHA}}$ is the trainable parameters of the projection modules given in Eq.~\ref{eq:MHA-STGS}, $\mathcal{R}_{\mathbf{P}}$, $\mathcal{R}_{\mathbf{U}}$, and $\mathcal{R}_{\mathbf{L}}$ are regularization terms defined in Sec.~\ref{sec:reg}. And $\alpha$, $\beta$, and $\gamma$ are hyperparameters to control the strength of these regularization terms. Here,$f$  represents the original dense model, and $f'$ is the model equipped with our designed modules for MoE construction. Under this setting, we use $\mathcal{L}(\cdot,\cdot)$ to calculate the KL divergence between the logits of $f$ and $f'$, which is used as the guidance to preserve the capacity of the dense model~\citep{hinton2015distillingknowledgeneuralnetwork}. We also found that using the KL divergence alone can lead to the best performance, and this observation complies with the experimental setup in~\citep{muralidharan2024compact}. Also, note that we perform \textbf{in-place} knowledge distillation since the original model weights are frozen. Thus, the knowledge distillation process does not introduce overheads in terms of GPU memory.

After learning how to construct the MoE, we convert the MLP layer to $N$ experts with shared weights. After pruning the MLP layer, we save $\frac{1}{N}\mathbf{1}^{\top}\mathbf{E}$ for MHA layers as the bias of the $\text{Proj}_{\text{E}}^{\text{\tiny MHA}}$, and we drop $\text{Proj}_{\text{D}}^{\text{\tiny MLP}}$ and convert Eq.~\ref{eq:MHA-STGS} into a Top-K routing function as well as use $\mathbf{s}_0$ for pruning $\mathbf{W}_Q$ and $\mathbf{W}_K$. Our construction also enables converting the MoE back into a pseudo-MoE model. The MoE model and the pseudo-MoE model are equivalent, and more details can be found in  Appendix~\ref{sec:app-pmoe}.

%, enabling easier support for efficient parallel training
\begin{table*}[t!]
\centering
\caption{Zero-shot task performance of compressed LLaMA-2 7B, LLaMA-3 8B, Qwen-2.5 7B. \label{tbl:zero_shot_main}}
\setlength{\extrarowheight}{2pt}
\resizebox{0.94\textwidth}{!}{
\begin{tabular}{c|c|l|c|c|c|c|c|c}
\cmidrule[1pt]{1-9}
% Model & Compression & Method & ARC-e & ARC-c & PIQA & WinoG. & HellaS. & Average \\ 
\multirow{2}{*}{Model} & \multirow{2}{*}{Active Parameters} & \multirow{2}{*}{Method} &  ARC-e & ARC-c & PIQA & WinoG. & HellaS. & \multirow{2}{*}{Average} \\ 
\cmidrule{4-8}
 &  &  & acc-norm & acc-norm & acc-norm & acc & acc-norm & \\
\cmidrule[1pt]{1-9}

\multirow{7}{*}{LLaMA-2 7B} 
& 100\% & Dense & 74.58 & 46.25 & 79.11 & 69.06 & 75.99 & 69.00 \\ \cmidrule{2-9}

% & \multirow{7}{*}{70\%} 
% & ShortGPT \citep{men2024shortgpt} & 48.65 & 32.85 & 64.31 & 64.33 & 56.13 & 53.25 \\ 
% & & SliceGPT \citep{ashkboos2023slicegpt} & 58.88 & 33.36 & 68.55 & 58.01 & 49.86 & 53.73 \\ 
% & & LLM Surgeon \citep{van2023llm} & {63.09} & 36.69 & {73.56} & 61.09 & 60.72 & 59.03 \\
% & & DISP-LLM ~\cite{gaodisp} & 59.81 & 33.19 & 71.82 & 62.27 & 63.43 & 58.10 \\ 
% & & DISP-LLM Alpaca~\cite{gaodisp} & 60.10 & 37.03 & 73.72 & 63.93 & 62.87 & 59.53 \\ 
% & & ModeGPT~\cite{lin2024modegpt} & 63.26 & {38.73} & 70.40 & {67.32} & {63.26} & {60.78} \\ 
% & & ModeGPT-Alpaca~\cite{lin2024modegpt} & 65.49 & 39.16 & 73.34 & 66.22 & 65.90 & 62.02 \\ 
% \cmidrule{2-9}
%62.27 63.43 59.81 33.19 71.82
%63.93 62.87 60.10 37.03 73.72 59.53 
& \multirow{6}{*}{60\%} 
& ShortGPT \citep{men2024shortgpt} & 41.16 & 29.94 & 60.12 & 60.46 & 43.67 & 47.07 \\ 
& & SliceGPT \citep{ashkboos2023slicegpt} & 36.49 & 24.57 & 54.90 & 53.43 & 34.80 & 40.84 \\ 
& & LLM Surgeon \citep{van2023llm} & {52.31} & {30.29} & {69.26} & 54.38 & 48.04 & 50.86 \\ 
& & ModeGPT~\citep{lin2024modegpt} & 49.45 & 30.03 & 64.96 & {61.96} & {53.01} & {51.88} \\ 
& & ModeGPT-Alpaca~\citep{lin2024modegpt} & 59.76 & 34.73 & 70.35 & \textbf{64.40} & 58.63 & 57.58 \\ 
& &\cellcolor{tabcolor} \textbf{ToMoE (Ours)} &\cellcolor{tabcolor} \textbf{63.64} &\cellcolor{tabcolor} \textbf{38.74} &\cellcolor{tabcolor} \textbf{72.85} &\cellcolor{tabcolor} 62.51 &\cellcolor{tabcolor} \textbf{65.84} &\cellcolor{tabcolor} \textbf{60.72} \\ 
% 65.84	62.51	63.64	38.74	72.85	60.716
% \cmidrule{2-9}
% & \multirow{4}{*}{50\%}
% &  LLM Surgeon ~\cite{van2023llm} & 44.91 & 26.28 & 64.36 & 52.57 &40.29 & 45.68 \\ 
% & & DISP-LLM ~\cite{gaodisp} & 43.06 & 25.85 & 63.93 & 54.54 & 63.43 &  46.72 \\ 
% & & DISP-LLM Alpaca~\cite{gaodisp} & 51.14 & 30.20 & 68.34 & 56.20 & 49.35 & 51.05 \\ 
% & &\cellcolor{tabcolor} \textbf{ToMoE (Ours)} &\cellcolor{tabcolor} \textbf{56.65} &\cellcolor{tabcolor} \textbf{33.87} &\cellcolor{tabcolor} \textbf{71.00} &\cellcolor{tabcolor} \textbf{58.56} &\cellcolor{tabcolor} \textbf{60.26} &\cellcolor{tabcolor} \textbf{56.07} \\ 
\cmidrule[1pt]{1-9}
%52.57 40.29 44.91 26.28 64.36 45.68
%54.54 46.33 43.06 25.85 63.93 46.72
\multirow{6}{*}{LLaMA-3 8B} 
& 100\% & Dense & 77.69 & 53.58 & 80.63 & 72.69 & 79.16 & 72.75 \\ 
\cmidrule{2-9}
& \multirow{3}{*}{75\%} 
& ShortGPT-Alpaca \citep{men2024shortgpt} & 38.13 & 31.40 & 60.94 & 54.22 & 31.52 & 43.24 \\ 
& & SliceGPT-Alpaca \citep{ashkboos2023slicegpt} & 44.44 & 29.27 & 57.56 & 58.48 & 41.08 & 46.17 \\ 
& & ModeGPT-Alpaca~\citep{lin2024modegpt} & 67.05 & 41.13 & 75.52 & \textbf{69.61} & 66.49 & 63.96 \\ 
\cmidrule{2-9}
& 70\% &\cellcolor{tabcolor}\textbf{ToMoE (Ours)} &\cellcolor{tabcolor} \textbf{71.55} &\cellcolor{tabcolor}  \textbf{44.71} &\cellcolor{tabcolor}  \textbf{76.28} &\cellcolor{tabcolor}  68.98 &\cellcolor{tabcolor}  \textbf{71.84} &\cellcolor{tabcolor}  \textbf{66.67} \\ 
\cmidrule{2-9}
&60\% &\cellcolor{tabcolor}\textbf{ToMoE (Ours)} &\cellcolor{tabcolor}  65.87 &\cellcolor{tabcolor}  41.64 &\cellcolor{tabcolor}  73.61 &\cellcolor{tabcolor}  63.30 &\cellcolor{tabcolor}  66.42 &\cellcolor{tabcolor}  62.17 \\ 
\cmidrule[1pt]{1-9}

\multirow{9}{*}{Qwen-2.5 7B} 
& 100\% & Dense & 79.42 & 50.17 & 79.54 & 71.35 & 78.36 & 71.77 \\ 
\cmidrule{2-9}
&\multirow{4}{*}{50\%}   & DISP-LLM Alpaca~\citep{gaodisp} &55.35& 34.22& 70.29& 53.59& 55.00& 53.69\\
% & \multirow{4}{*}{50\%}  & DISP-LLM Alpaca~\cite{gaodisp}& 52.65& 27.99& 65.94& 53.43& 44.11& 53.69\\
& &\cellcolor{tabcolor}\textbf{ToMoE N=8 (Ours)}&\cellcolor{tabcolor}  61.83 &\cellcolor{tabcolor}  36.77 &\cellcolor{tabcolor}  71.82 &\cellcolor{tabcolor}  57.70 &\cellcolor{tabcolor}  59.55 &\cellcolor{tabcolor}  57.53 \\
& &\cellcolor{tabcolor}\textbf{ToMoE N=16 (Ours)}&\cellcolor{tabcolor} \textbf{64.81} &\cellcolor{tabcolor} \textbf{41.72} &\cellcolor{tabcolor} \textbf{73.45} &\cellcolor{tabcolor} 58.48 &\cellcolor{tabcolor}  61.06 &\cellcolor{tabcolor}\textbf{59.90} \\ 
& &\cellcolor{tabcolor}\textbf{ToMoE N=24 (Ours)}&\cellcolor{tabcolor} 64.69 &\cellcolor{tabcolor} 39.85 &\cellcolor{tabcolor}  72.96 &\cellcolor{tabcolor} \textbf{58.56} &\cellcolor{tabcolor} \textbf{63.21} &\cellcolor{tabcolor} 59.85 \\ 
\cmidrule{2-9}
&\multirow{4}{*}{40\%}   & DISP-LLM Alpaca~\citep{gaodisp} &52.65& 27.99& 65.94& 53.43& 44.11& 48.82\\
& &\cellcolor{tabcolor}\textbf{ToMoE N=8 (Ours)} &\cellcolor{tabcolor}  53.80  &\cellcolor{tabcolor}  32.59 &\cellcolor{tabcolor}  69.59 &\cellcolor{tabcolor}  55.49 &\cellcolor{tabcolor}  52.98 &\cellcolor{tabcolor}  52.89 \\ 
& &\cellcolor{tabcolor}\textbf{ToMoE N=16 (Ours)} &\cellcolor{tabcolor}  53.87 &\cellcolor{tabcolor}  \textbf{33.70}  &\cellcolor{tabcolor}  \textbf{71.98} &\cellcolor{tabcolor}  54.38 &\cellcolor{tabcolor}  \textbf{55.65} &\cellcolor{tabcolor}  53.92 \\ 
& &\cellcolor{tabcolor}\textbf{ToMoE N=24 (Ours)} &\cellcolor{tabcolor}  \textbf{55.43} &\cellcolor{tabcolor}  32.85 &\cellcolor{tabcolor}  69.86 &\cellcolor{tabcolor}  \textbf{57.85} &\cellcolor{tabcolor}  {54.62} &\cellcolor{tabcolor}  \textbf{54.12} \\ 
\cmidrule[1pt]{1-9}

%LLaMA-38b 30%	71.84	68.98	71.55	44.71	76.28	66.672
%               66.42	63.3	65.87	41.64	73.61	62.168
%\cmidrule[1pt]{1-9}
%\midrule
\end{tabular}
}
\vspace{-1pt}
\end{table*}
\vspace{-10pt}
\section{Experiments}
\subsection{Settings}
\noindent\textbf{Models.} Our ToMoE method is evaluated using several LLMs with decoder blocks. Specifically, we choose the following models: LLaMA-2~\citep{touvron2023llama2}: LLaMA-2 7B and LLaMA-2 13B; LLaMA-3 8B~\citep{dubey2024llama}; Phi-2~\citep{javaheripi2023phi}; Qwen-2.5~\citep{yang2024qwen2}: Qwen-2.5 7B and Qwen-2.5 14B. Results for LLaMA-2 13B and Qwen-2.5 14B are presented in the Appendix~\ref{sec:app-exp}.

\noindent\textbf{Implementations.}  ToMoE is implemented by Pytorch~\citep{paszke2019pytorch} and Hugging Face transformer library~\citep{wolf-etal-2020-transformers}. 
The model weights are frozen when training the modules with learnable parameters $\Theta$ in Obj.~\ref{Eq:all}. We use the AdamW~\citep{loshchilov2018decoupled} optimizer to optimize $\Theta$, which is trained for 10,000 iterations for all models. 
For all experiments, we set $\alpha = 16$, $\beta = 2.0$, and $\gamma = 1.0$, where $\alpha$, $\beta$, and $\gamma$ are defined in Obj.~\ref{Eq:all}. Without specific descriptions, the number of experts for ToMoE is $8$ across all settings. Depending on the size of the base model, 1 to 4 NVIDIA A100 GPUs are used to train $\Theta$. More implementation details can be found in  Appendix~\ref{sec:app-imp}.

\noindent\textbf{Datasets.} Two training settings are provided for all modules with learnable parameters $\Theta$: (1) using WikiText~\cite{merity2016pointer}, and (2) using a mixed dataset comprising WikiText, Alpaca~\cite{alpaca}, and Code-Alpaca~\cite{codealpaca} (mixing ratio: 1:1:1). Based on our observation, {ToMoE} benefits from a diverse mixture of datasets to effectively construct experts. Following previous methods~\citep{ashkboos2023slicegpt, gaodisp}, our method and other methods are evaluated on five well-known zero-shot tasks: PIQA~\citep{bisk2020piqa}; WinoGrande~\citep{sakaguchi2021winogrande}; HellaSwag~\citep{zellers2019hellaswag}; ARC-e and ARC-c~\citep{clark2018think}. \revise{We further evaluate our method on the following tasks and configurations to ensure consistency with comparison baselines: 32-shot {BoolQ~\citep{clark2019boolq}}, {SciQ~\citep{SciQ}}, 5-shot {WinoGrande}, 25-shot {ARC-c}, 10-shot {HellaSwag}, {TruthfulQA~\cite{lin2022truthfulqa}}, and 5-shot {MMLU}~\citep{hendryckstest2021}.} We use llm-eval-harness~\citep{eval-harness} to evaluate the compressed models.

\noindent\textbf{Baselines.}  ToMoE is compared to  baselines from structural pruning methods~\citep{ma2023llm,ashkboos2023slicegpt,men2024shortgpt,songsleb,van2023llm,lin2024modegpt,gaodisp}, semi-structural pruning methods~\citep{frantar2023sparsegpt,sun2024a,dongpruner} and MoE construction \revise{methods~\citep{zhu2024llama,lee2024breaking,qu2024llama,pei2025cmoe}.}

% like LLM-Pruner~\cite{ma2023llm}, SliceGPT~\cite{ashkboos2023slicegpt}, ShortGPT~\cite{men2024shortgpt}, SLEB~\cite{songsleb}, LLMSurgeon~\cite{van2023llm}, ModeGPT~\cite{lin2024modegpt}, and DISP-LLM~\cite{gaodisp}. We also include semi-structure pruning baselines like SparseGPT~\cite{frantar2023sparsegpt}, Wanda~\cite{sun2024a}, and Pruner-Zero~\cite{dongpruner}.

\begin{table*}[t!]
\centering
\caption{Compassion against MoE construction methods. \label{tbl:zero_shot_dense_to_moe}}
\setlength{\extrarowheight}{2pt}
\resizebox{0.94\textwidth}{!}{
\begin{tabular}{c|c|l|c|c|c|c|c|c}
\cmidrule[1pt]{1-9}
\multirow{2}{*}{Model} & \multirow{2}{*}{Active Parameters} & \multirow{2}{*}{Method} &  ARC-e & ARC-c & PIQA & WinoG. & HellaS. & \multirow{2}{*}{Average} \\ 
\cmidrule{4-8}
 &  &  & acc-norm & acc-norm & acc-norm & acc & acc-norm & \\
\cmidrule[1pt]{1-9}

\multirow{6}{*}{LLaMA-2 7B} 
& 100\% & Dense & 74.58 & 46.25 & 79.11 & 69.06 & 75.99 & 69.00 \\ \cmidrule{2-9}
&\multirow{5}{*}{50\%} 
& LLaMA-MoE E16A4~\citep{zhu2024llama} & 27.02 & 29.53 & 49.13 & 49.49 & 26.19 & 36.27 \\
& & + fine-tuning & 35.52 & 26.88 & 49.13 & 55.71 & 34.07 & 40.55 \\
& & LLaMA-MoE E8A2~\citep{zhu2024llama} & 25.76 & 27.22 & 50.92 & 47.51 & 26.02 & 35.49 \\
& & + fine-tuning & 37.71 & 27.56 & 58.11 & 51.78 & 36.37 & 42.31 \\
& & \cellcolor{tabcolor} \textbf{ToMoE (Ours)} & \cellcolor{tabcolor} \textbf{56.65} & \cellcolor{tabcolor} \textbf{33.87} & \cellcolor{tabcolor} \textbf{71.00} & \cellcolor{tabcolor} \textbf{58.56} & \cellcolor{tabcolor} \textbf{60.26} & \cellcolor{tabcolor} \textbf{56.07} \\
\cmidrule[1pt]{1-9}
\multirow{4}{*}{Phi-2} 
& 100\%   & Dense & 78.24 & 54.01 & 79.11 & 75.61 & 73.86 & 72.17 \\ \cmidrule{2-9}
& 78\% & G-MoEfication~\citep{lee2024breaking} & 66.46 & 38.91 & 70.29 & 65.75 & 57.90 & 59.86 \\
\cmidrule[1pt]{2-9}
& \multirow{2}{*}{70\%} 
& G-MoEfication~\citep{lee2024breaking} & 57.95 & 35.24 & 64.80 & 60.30 & 49.07 & 53.47 \\
&      & \cellcolor{tabcolor} \textbf{ToMoE (Ours)} & \cellcolor{tabcolor} \textbf{70.79} & \cellcolor{tabcolor} \textbf{43.86} & \cellcolor{tabcolor} \textbf{77.00} & \cellcolor{tabcolor} \textbf{66.38} & \cellcolor{tabcolor} \textbf{62.68} & \cellcolor{tabcolor} \textbf{64.16} \\
\cmidrule[1pt]{1-9}

\end{tabular}
}
\vspace{-5pt}
\end{table*}

\begin{table*}[t!]
\centering
\caption{\revise{Comparison against MoE construction methods on LLaMA-2 7B.} \label{tab:come_comparasion}}
\setlength{\extrarowheight}{2pt}
\resizebox{0.97\textwidth}{!}{
\begin{tabular}{c|l|c|c|c|c|c|c|c|c}
\cmidrule[1pt]{1-10}
\multirow{2}{*}{Active Parameters} & \multirow{2}{*}{Method} & \multirow{2}{*}{\# Tokens} & 
BoolQ (32) & SciQ & PIQA & WinoG. (5) & ARC-C (25) & HellaS. (10) & \multirow{2}{*}{Average} \\
\cmidrule{4-9}
 &  &  &  acc & acc & acc & acc & acc-norm & acc-norm &  \\
\cmidrule[1pt]{1-10}

100\% & LLaMA-2 7B & 2T & 82.04 & 90.80 & 78.78 & 73.95 & 53.15 & 78.55 & 76.21 \\ 
\cmidrule{1-10}
\multirow{5}{*}{50\%} 
& LLaMA-MoE E8A2~\citep{zhu2024llama} & 1.2B &  37.83 & 20.00 & 49.73 & 50.12 & 25.79 & 26.18 & 34.27 \\
& LLaMA-MoE-v2 E8A2~\citep{qu2024llama} & 1.2B &  51.25 & 67.00 & 56.64 & 52.88 & 25.68 & 35.10 & 48.59 \\
& CMoE E8S1A1~\citep{pei2025cmoe} & \textbf{0} &   46.09 & 65.30 & 52.77 & 48.70 & 23.80 & 30.12 & 44.13 \\
& + fine-tuning & 1.2B &{55.04} & {77.50} & {57.12} & {54.06} & {27.56} & {38.79} & {51.68} \\
%\cmidrule[1pt]{1-10}
& \cellcolor{tabcolor}\textbf{ToMoE (Ours)} & \cellcolor{tabcolor}{0.02B} & %\cellcolor{tabcolor}\checkmark &
\cellcolor{tabcolor}\textbf{62.42} &
\cellcolor{tabcolor}\textbf{81.70} &
\cellcolor{tabcolor}\textbf{71.00} &
\cellcolor{tabcolor}\textbf{60.41} &
\cellcolor{tabcolor}\textbf{34.39} &
\cellcolor{tabcolor}\textbf{60.54} &
\cellcolor{tabcolor}\textbf{61.39} \\
\cmidrule[1pt]{1-10}
\end{tabular}
}
\vspace{-5pt}
\end{table*}

\begin{table*}[t!]
\centering
\caption{\revise{Comparison against LLaMA-MoE-v2 on LLaMA-3 8B.} \label{tab:vs_llama_moe_v2}}
\setlength{\extrarowheight}{2pt}
\resizebox{0.97\textwidth}{!}{
\begin{tabular}{c|l|c|c|c|c|c|c|c|c|c}
\cmidrule[1pt]{1-11}
\multirow{2}{*}{Active Parameters} & \multirow{2}{*}{Method} & \multirow{2}{*}{\# Tokens} &
BoolQ (32) & SciQ & PIQA & ARC-C (25) & TruthfulQA & HellaS. (10) & MMLU (5) & \multirow{2}{*}{Average} \\
\cmidrule{4-10}
 &  &  & acc & acc-norm & acc & acc-norm & acc & acc-norm & acc &  \\
\cmidrule[1pt]{1-11}

100\% & LLaMA-3 8B & 15T & 83.00 & 93.20 & 78.51 & 61.86 & 51.71 & 78.79 & 67.22 & 73.33 \\ 
\cmidrule{1-11}
\multirow{3}{*}{50\%} 
& LLaMA-MoE-v2 E8A2~\citep{qu2024llama} & 7B & 74.62 & \textbf{90.60} & 69.26 & \textbf{42.83} & 45.62 & {58.95} & 37.41 & \textbf{59.61} \\
& LLaMA-MoE-v2 E8A1S1~\citep{qu2024llama} & 7B & \textbf{76.88} & 88.80 & 67.90 & 40.19 & \textbf{46.85} & 53.67 & \textbf{40.89} & {59.31} \\
& \cellcolor{tabcolor}\textbf{ToMoE (Ours)} & \cellcolor{tabcolor}\textbf{0.02B} &
\cellcolor{tabcolor}63.67 &
\cellcolor{tabcolor}{87.50} &
\cellcolor{tabcolor}\textbf{72.63} &
\cellcolor{tabcolor}37.29 &
\cellcolor{tabcolor}40.48 &
\cellcolor{tabcolor}\textbf{63.91} &
\cellcolor{tabcolor}36.31 &
\cellcolor{tabcolor}57.40 \\
\cmidrule[1pt]{1-11}
\end{tabular}
}
\vspace{-5pt}
\end{table*}

\begin{table*}[t]
\centering
\caption{ToMoE Visualization for the last layer of the LLaMA-2 7B model with 50\% active parameters}\label{tab:gen_sample}
\setlength\extrarowheight{2pt}
\resizebox{0.85\linewidth}{!}{
\begin{tabular}{|p{1.35cm}|p{16cm}|}
\hline
% \textbf{LLaMA-2 7B} &
Expert Color:
{\setlength{\fboxsep}{0pt}\sethlcolor{color0}\hl{Expert 1}
\sethlcolor{color1}\hl{Expert 2} \sethlcolor{color2}\hl{Expert 3}  \sethlcolor{color3}\hl{Expert 4}  \sethlcolor{color4}\hl{Expert 5} \sethlcolor{color5}\hl{Expert 6} \sethlcolor{color6}\hl{Expert 7} \sethlcolor{color7}\hl{Expert 8}
}
&
{\setlength{\fboxsep}{0pt}
\sethlcolor{color2}\hl{\texttt{<s>}}\sethlcolor{color1}\hl{ Grand}\sethlcolor{color7}\hl{ The}\sethlcolor{color6}\hl{ft}\sethlcolor{color2}\hl{ Auto}\sethlcolor{color1}\hl{ VI}\sethlcolor{color6}\hl{ is}\sethlcolor{color5}\hl{ an}\sethlcolor{color2}\hl{ up}\sethlcolor{color5}\hl{coming}\sethlcolor{color5}\hl{ video}\sethlcolor{color6}\hl{ game}\sethlcolor{color5}\hl{ in}\sethlcolor{color2}\hl{ development}\sethlcolor{color2}\hl{ by}\sethlcolor{color7}\hl{ Rock}\sethlcolor{color2}\hl{star}\sethlcolor{color3}\hl{ Games}\sethlcolor{color0}\hl{.}\sethlcolor{color2}\hl{ It}\sethlcolor{color6}\hl{ is}\sethlcolor{color2}\hl{ due}\sethlcolor{color4}\hl{ to}\sethlcolor{color6}\hl{ be}\sethlcolor{color5}\hl{ the}\sethlcolor{color7}\hl{ e}\sethlcolor{color4}\hl{ighth}\sethlcolor{color5}\hl{ main}\sethlcolor{color6}\hl{ Grand}\sethlcolor{color7}\hl{ The}\sethlcolor{color6}\hl{ft}\sethlcolor{color5}\hl{ Auto}\sethlcolor{color2}\hl{ game}\sethlcolor{color4}\hl{,}\sethlcolor{color4}\hl{ following}\sethlcolor{color6}\hl{ Grand}\sethlcolor{color6}\hl{ The}\sethlcolor{color6}\hl{ft}\sethlcolor{color6}\hl{ Auto}\sethlcolor{color3}\hl{ V}\sethlcolor{color1}\hl{ (}\sethlcolor{color2}\hl{2}\sethlcolor{color7}\hl{0}\sethlcolor{color7}\hl{1}\sethlcolor{color2}\hl{3}\sethlcolor{color4}\hl{),}\sethlcolor{color2}\hl{ and}\sethlcolor{color6}\hl{ the}\sethlcolor{color5}\hl{ six}\sethlcolor{color2}\hl{teenth}\sethlcolor{color2}\hl{ entry}\sethlcolor{color3}\hl{ overall}\sethlcolor{color6}\hl{.}\sethlcolor{color2}\hl{ Set}\sethlcolor{color4}\hl{ within}\sethlcolor{color5}\hl{ the}\sethlcolor{color7}\hl{ fict}\sethlcolor{color4}\hl{ional}\sethlcolor{color2}\hl{ open}\sethlcolor{color5}\hl{ world}\sethlcolor{color0}\hl{ state}\sethlcolor{color4}\hl{ of}\sethlcolor{color6}\hl{ Leon}\sethlcolor{color6}\hl{id}\sethlcolor{color2}\hl{ab}\sethlcolor{color2}\hl{ased}\sethlcolor{color4}\hl{ on}\sethlcolor{color2}\hl{ Florida}\sethlcolor{color4}\hl{and}\sethlcolor{color4}\hl{ its}\sethlcolor{color2}\hl{ Miami}\sethlcolor{color1}\hl{-}\sethlcolor{color7}\hl{in}\sethlcolor{color7}\hl{sp}\sethlcolor{color5}\hl{ired}\sethlcolor{color7}\hl{ Vice}\sethlcolor{color1}\hl{ City}\sethlcolor{color4}\hl{,}\sethlcolor{color5}\hl{ the}\sethlcolor{color6}\hl{ story}\sethlcolor{color6}\hl{ is}\sethlcolor{color2}\hl{ expected}\sethlcolor{color4}\hl{ to}\sethlcolor{color4}\hl{ follow}\sethlcolor{color4}\hl{ the}\sethlcolor{color1}\hl{ criminal}\sethlcolor{color7}\hl{ du}\sethlcolor{color6}\hl{o}\sethlcolor{color4}\hl{ of}\sethlcolor{color7}\hl{ Lu}\sethlcolor{color1}\hl{cia}\sethlcolor{color6}\hl{ and}\sethlcolor{color5}\hl{ her}\sethlcolor{color5}\hl{ male}\sethlcolor{color1}\hl{ partner}\sethlcolor{color1}\hl{.}\sethlcolor{color1}\hl{ \textbackslash}\sethlcolor{color1}\hl{n}\sethlcolor{color5}\hl{F}\sethlcolor{color7}\hl{ollow}\sethlcolor{color4}\hl{ing}\sethlcolor{color2}\hl{ years}\sethlcolor{color4}\hl{ of}\sethlcolor{color7}\hl{ spec}\sethlcolor{color2}\hl{ulation}\sethlcolor{color4}\hl{ and}\sethlcolor{color7}\hl{ anticip}\sethlcolor{color2}\hl{ation}\sethlcolor{color4}\hl{,}\sethlcolor{color2}\hl{ Rock}\sethlcolor{color6}\hl{star}\sethlcolor{color2}\hl{ confirmed}\sethlcolor{color4}\hl{ in}\sethlcolor{color2}\hl{ February}\sethlcolor{color1}\hl{ }\sethlcolor{color1}\hl{ 2}\sethlcolor{color7}\hl{0}\sethlcolor{color7}\hl{2}\sethlcolor{color2}\hl{2}\sethlcolor{color4}\hl{ that}\sethlcolor{color5}\hl{ the}\sethlcolor{color2}\hl{ game}\sethlcolor{color6}\hl{ was}\sethlcolor{color5}\hl{ in}\sethlcolor{color3}\hl{ development}\sethlcolor{color1}\hl{.}\sethlcolor{color4}\hl{ That}\sethlcolor{color2}\hl{ September}\sethlcolor{color4}\hl{,}\sethlcolor{color5}\hl{ foot}\sethlcolor{color6}\hl{age}\sethlcolor{color4}\hl{ from}\sethlcolor{color7}\hl{ un}\sethlcolor{color7}\hl{fin}\sethlcolor{color2}\hl{ished}\sethlcolor{color6}\hl{ versions}\sethlcolor{color6}\hl{ was}\sethlcolor{color7}\hl{ le}\sethlcolor{color2}\hl{aked}\sethlcolor{color3}\hl{ online}\sethlcolor{color4}\hl{ in}\sethlcolor{color2}\hl{ what}\sethlcolor{color2}\hl{ journal}\sethlcolor{color6}\hl{ists}\sethlcolor{color2}\hl{ described}\sethlcolor{color5}\hl{ as}\sethlcolor{color6}\hl{ one}\sethlcolor{color4}\hl{ of}\sethlcolor{color5}\hl{ the}\sethlcolor{color5}\hl{ biggest}\sethlcolor{color7}\hl{ le}\sethlcolor{color2}\hl{aks}\sethlcolor{color5}\hl{ in}\sethlcolor{color5}\hl{ the}\sethlcolor{color3}\hl{ history}\sethlcolor{color5}\hl{ of}\sethlcolor{color5}\hl{ the}\sethlcolor{color6}\hl{ video}\sethlcolor{color5}\hl{ game}\sethlcolor{color3}\hl{ industry}\sethlcolor{color0}\hl{.}\sethlcolor{color5}\hl{ The}\sethlcolor{color2}\hl{ game}\sethlcolor{color6}\hl{ was}\sethlcolor{color6}\hl{ formally}\sethlcolor{color2}\hl{ revealed}\sethlcolor{color4}\hl{ in}\sethlcolor{color2}\hl{ December}\sethlcolor{color6}\hl{ }\sethlcolor{color6}\hl{ 2}\sethlcolor{color1}\hl{0}\sethlcolor{color7}\hl{2}\sethlcolor{color3}\hl{3}\sethlcolor{color6}\hl{ and}\sethlcolor{color6}\hl{ is}\sethlcolor{color2}\hl{ scheduled}\sethlcolor{color4}\hl{ to}\sethlcolor{color6}\hl{ be}\sethlcolor{color2}\hl{ released}\sethlcolor{color2}\hl{ in}\sethlcolor{color6}\hl{ late}\sethlcolor{color6}\hl{ }\sethlcolor{color7}\hl{ 2}\sethlcolor{color7}\hl{0}\sethlcolor{color2}\hl{2}\sethlcolor{color2}\hl{5}\sethlcolor{color4}\hl{ for}\sethlcolor{color2}\hl{ the}\sethlcolor{color7}\hl{ Play}\sethlcolor{color2}\hl{Station}\sethlcolor{color1}\hl{ }\sethlcolor{color2}\hl{ 5}\sethlcolor{color2}\hl{ and}\sethlcolor{color7}\hl{ X}\sethlcolor{color2}\hl{box}\sethlcolor{color2}\hl{ Series}\sethlcolor{color2}\hl{ X}\sethlcolor{color7}\hl{/}\sethlcolor{color2}\hl{S}\sethlcolor{color1}\hl{.}\sethlcolor{color1}\hl{ \textbackslash}\sethlcolor{color1}\hl{n}\sethlcolor{color6}\hl{Gr}\sethlcolor{color6}\hl{and}\sethlcolor{color6}\hl{ The}\sethlcolor{color6}\hl{ft}\sethlcolor{color2}\hl{ Auto}\sethlcolor{color2}\hl{ VI}\sethlcolor{color6}\hl{ is}\sethlcolor{color2}\hl{ set}\sethlcolor{color4}\hl{ in}\sethlcolor{color4}\hl{ the}\sethlcolor{color7}\hl{ fict}\sethlcolor{color4}\hl{ional}\sethlcolor{color6}\hl{ open}\sethlcolor{color6}\hl{ world}\sethlcolor{color2}\hl{ state}\sethlcolor{color6}\hl{ of}\sethlcolor{color6}\hl{ Leon}\sethlcolor{color6}\hl{id}\sethlcolor{color6}\hl{ab}\sethlcolor{color2}\hl{ased}\sethlcolor{color6}\hl{ on}\sethlcolor{color3}\hl{ Florida}\sethlcolor{color4}\hl{which}\sethlcolor{color4}\hl{ includes}\sethlcolor{color6}\hl{ Vice}\sethlcolor{color2}\hl{ City}\sethlcolor{color4}\hl{,}\sethlcolor{color5}\hl{ a}\sethlcolor{color7}\hl{ fict}\sethlcolor{color5}\hl{ional}\sethlcolor{color4}\hl{ised}\sethlcolor{color3}\hl{ version}\sethlcolor{color4}\hl{ of}\sethlcolor{color6}\hl{ Miami}\sethlcolor{color1}\hl{.}\sethlcolor{color6}\hl{ Vice}\sethlcolor{color2}\hl{ City}\sethlcolor{color6}\hl{ was}\sethlcolor{color6}\hl{ previously}\sethlcolor{color2}\hl{ featured}\sethlcolor{color4}\hl{ in}\sethlcolor{color6}\hl{ Grand}\sethlcolor{color6}\hl{ The}\sethlcolor{color6}\hl{ft}\sethlcolor{color2}\hl{ Auto}\sethlcolor{color1}\hl{ (}\sethlcolor{color7}\hl{1}\sethlcolor{color7}\hl{9}\sethlcolor{color7}\hl{9}\sethlcolor{color2}\hl{7}\sethlcolor{color2}\hl{)}\sethlcolor{color6}\hl{ and}\sethlcolor{color4}\hl{ as}\sethlcolor{color4}\hl{ the}\sethlcolor{color5}\hl{ main}\sethlcolor{color2}\hl{ setting}\sethlcolor{color4}\hl{ of}\sethlcolor{color6}\hl{ Grand}\sethlcolor{color6}\hl{ The}\sethlcolor{color6}\hl{ft}\sethlcolor{color2}\hl{ Auto}\sethlcolor{color1}\hl{:}\sethlcolor{color7}\hl{ Vice}\sethlcolor{color3}\hl{ City}\sethlcolor{color1}\hl{ (}\sethlcolor{color7}\hl{2}\sethlcolor{color7}\hl{0}\sethlcolor{color7}\hl{0}\sethlcolor{color7}\hl{2}\sethlcolor{color2}\hl{)}\sethlcolor{color6}\hl{ and}\sethlcolor{color6}\hl{ Grand}\sethlcolor{color6}\hl{ The}\sethlcolor{color6}\hl{ft}\sethlcolor{color2}\hl{ Auto}\sethlcolor{color4}\hl{:}\sethlcolor{color6}\hl{ Vice}\sethlcolor{color7}\hl{ City}\sethlcolor{color7}\hl{ St}\sethlcolor{color2}\hl{ories}\sethlcolor{color1}\hl{ (}\sethlcolor{color7}\hl{2}\sethlcolor{color7}\hl{0}\sethlcolor{color7}\hl{0}\sethlcolor{color7}\hl{6}\sethlcolor{color1}\hl{).}\sethlcolor{color5}\hl{ The}\sethlcolor{color2}\hl{ game}\sethlcolor{color2}\hl{ world}\sethlcolor{color7}\hl{ par}\sethlcolor{color7}\hl{od}\sethlcolor{color4}\hl{ies}\sethlcolor{color1}\hl{ }\sethlcolor{color7}\hl{ 2}\sethlcolor{color7}\hl{0}\sethlcolor{color7}\hl{2}\sethlcolor{color5}\hl{0}\sethlcolor{color5}\hl{s}\sethlcolor{color5}\hl{ American}\sethlcolor{color3}\hl{ culture}\sethlcolor{color4}\hl{,}\sethlcolor{color4}\hl{ with}\sethlcolor{color7}\hl{ sat}\sethlcolor{color7}\hl{ir}\sethlcolor{color5}\hl{ical}\sethlcolor{color7}\hl{ dep}\sethlcolor{color7}\hl{ict}\sethlcolor{color2}\hl{ions}\sethlcolor{color4}\hl{ of}\sethlcolor{color5}\hl{ social}\sethlcolor{color5}\hl{ media}\sethlcolor{color5}\hl{ and}\sethlcolor{color7}\hl{ influen}\sethlcolor{color5}\hl{cer}\sethlcolor{color2}\hl{ culture}\sethlcolor{color4}\hl{,}\sethlcolor{color4}\hl{ and}\sethlcolor{color4}\hl{ references}\sethlcolor{color4}\hl{ to}\sethlcolor{color5}\hl{ Internet}\sethlcolor{color7}\hl{ mem}\sethlcolor{color3}\hl{es}\sethlcolor{color0}\hl{ such}\sethlcolor{color5}\hl{ as}\sethlcolor{color5}\hl{ Florida}\sethlcolor{color7}\hl{ Man}\sethlcolor{color1}\hl{.}\sethlcolor{color5}\hl{ The}\sethlcolor{color2}\hl{ story}\sethlcolor{color4}\hl{ follows}\sethlcolor{color6}\hl{ a}\sethlcolor{color5}\hl{ criminal}\sethlcolor{color7}\hl{ du}\sethlcolor{color6}\hl{o}\sethlcolor{color5}\hl{:}\sethlcolor{color6}\hl{ Lu}\sethlcolor{color2}\hl{cia}\sethlcolor{color4}\hl{,}\sethlcolor{color4}\hl{ the}\sethlcolor{color5}\hl{ series}\sethlcolor{color5}\hl{'}\sethlcolor{color5}\hl{ first}\sethlcolor{color5}\hl{ female}\sethlcolor{color7}\hl{ protagon}\sethlcolor{color2}\hl{ist}\sethlcolor{color4}\hl{ since}\sethlcolor{color1}\hl{ }\sethlcolor{color7}\hl{ 2}\sethlcolor{color7}\hl{0}\sethlcolor{color7}\hl{0}\sethlcolor{color2}\hl{0}\sethlcolor{color2}\hl{,}\sethlcolor{color6}\hl{and}\sethlcolor{color5}\hl{ her}\sethlcolor{color5}\hl{ male}\sethlcolor{color6}\hl{ partner}\sethlcolor{color4}\hl{;}\sethlcolor{color5}\hl{ the}\sethlcolor{color5}\hl{ first}\sethlcolor{color7}\hl{ tra}\sethlcolor{color6}\hl{iler}\sethlcolor{color7}\hl{ dep}\sethlcolor{color6}\hl{ict}\sethlcolor{color4}\hl{s}\sethlcolor{color6}\hl{ Lu}\sethlcolor{color6}\hl{cia}\sethlcolor{color5}\hl{ as}\sethlcolor{color4}\hl{ a}\sethlcolor{color5}\hl{ prison}\sethlcolor{color5}\hl{ in}\sethlcolor{color6}\hl{mate}\sethlcolor{color4}\hl{,}\sethlcolor{color4}\hl{ and}\sethlcolor{color4}\hl{ later}\sethlcolor{color7}\hl{ ev}\sethlcolor{color4}\hl{ading}\sethlcolor{color7}\hl{ cust}\sethlcolor{color5}\hl{ody}\sethlcolor{color4}\hl{ with}\sethlcolor{color6}\hl{ her}\sethlcolor{color2}\hl{ partner}\sethlcolor{color1}\hl{.}
}
\\
\hline
\end{tabular}
}
\vspace{-5pt}
\end{table*} 

\begin{figure*}[t]
	\subfloat[Loss $\mathcal{L}$]{
	\begin{minipage}[b]{.22\linewidth}
			\centering
			\includegraphics[width=.99\textwidth]{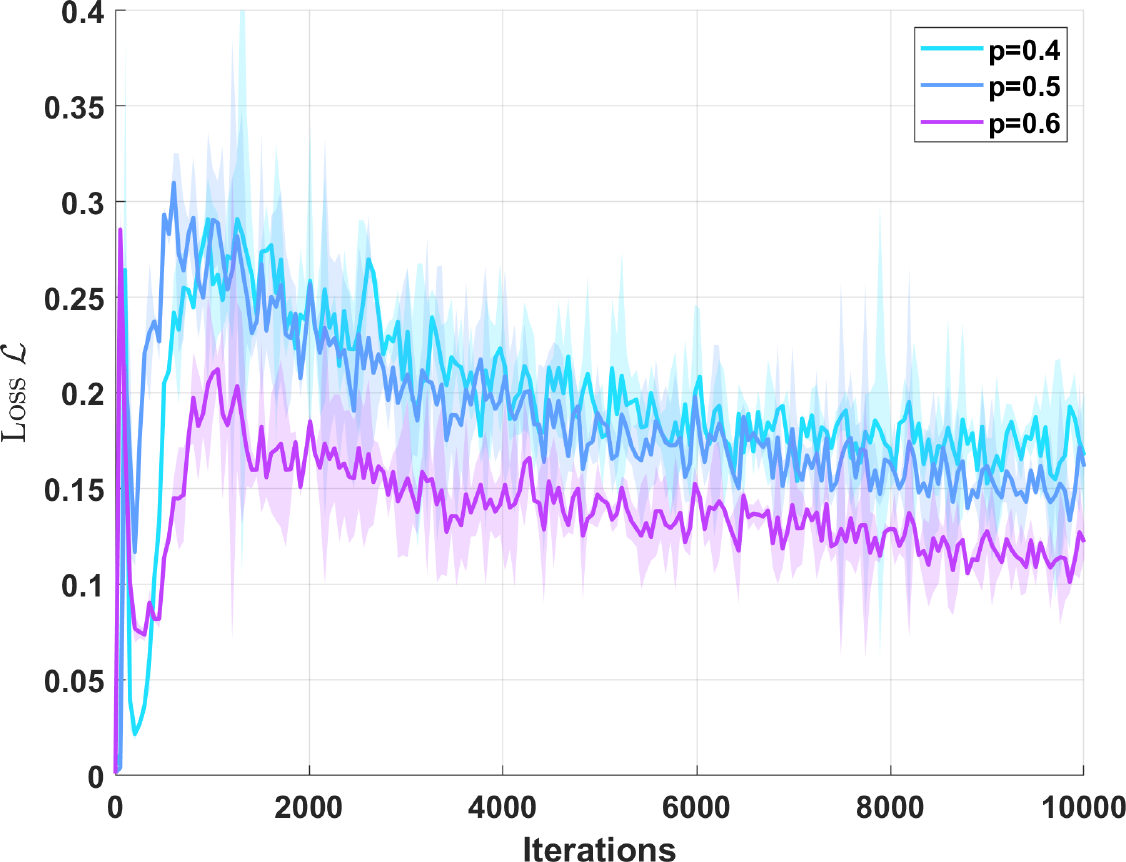}
	\end{minipage}~\label{p-l}}
        \hfill
	\subfloat[Loss $\mathcal{R}_\mathbf{P}$]{
		\begin{minipage}[b]{.22\linewidth}
			\centering
			\includegraphics[width=.99\textwidth]{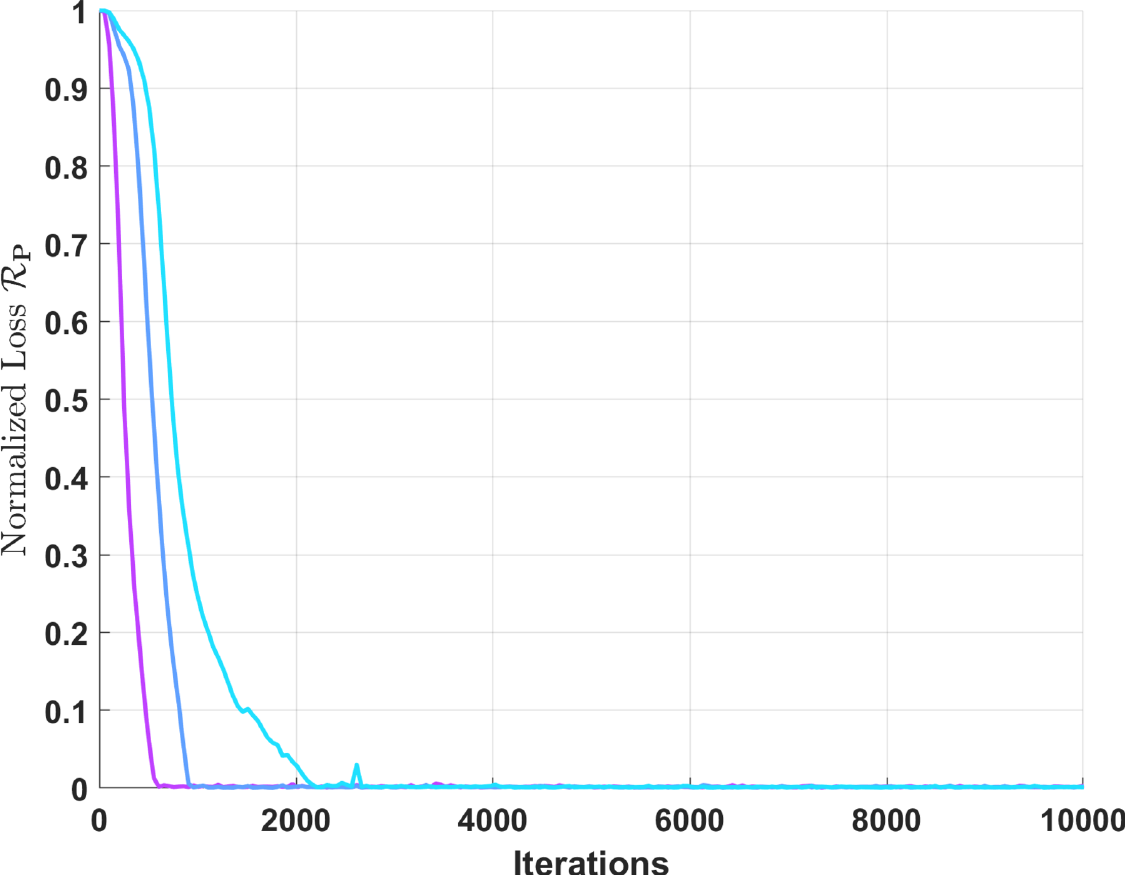}
	\end{minipage}~\label{p-r}}
        \hfill
	\subfloat[Loss $\mathcal{R}_\mathbf{U}$]{
		\begin{minipage}[b]{.22\linewidth}
			\centering
			\includegraphics[width=.99\textwidth]{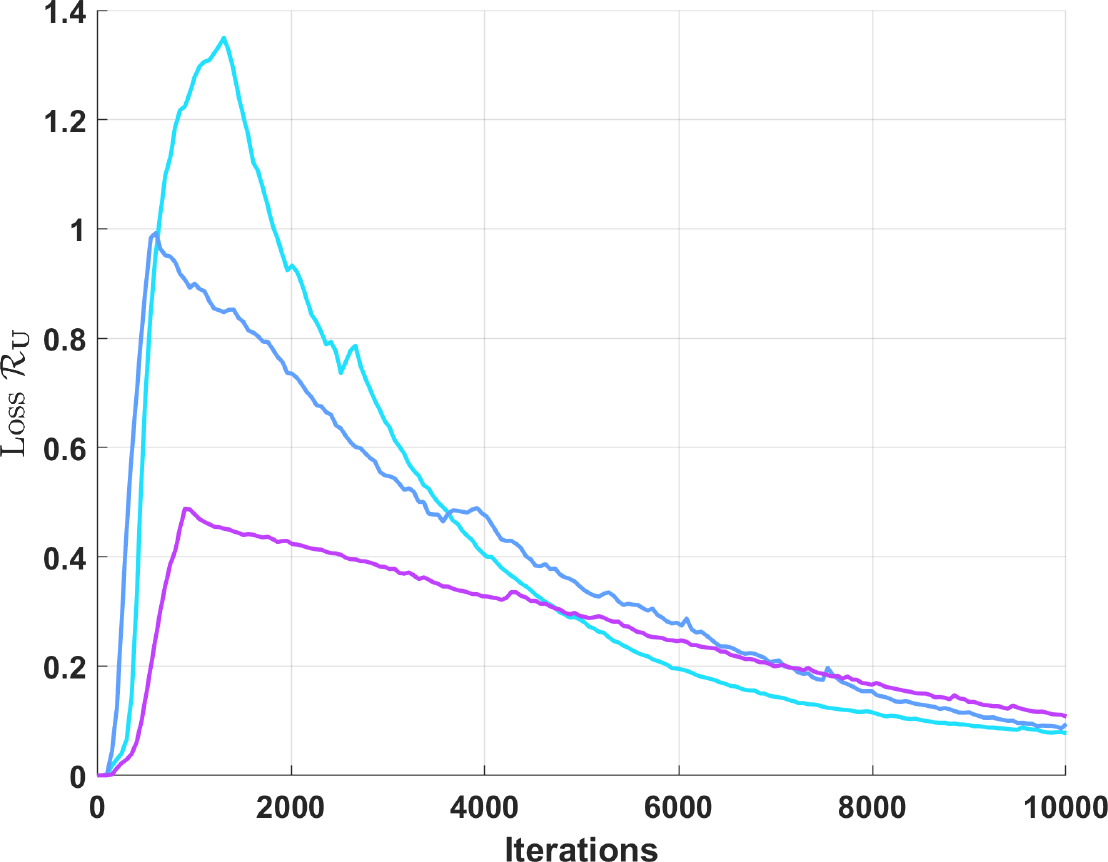}
	\end{minipage}~\label{throughput}}
        \hfill
	\subfloat[Loss $\mathcal{R}_\mathbf{L}$]{
		\begin{minipage}[b]{.22\linewidth}
			\centering
			\includegraphics[width=.99\textwidth]{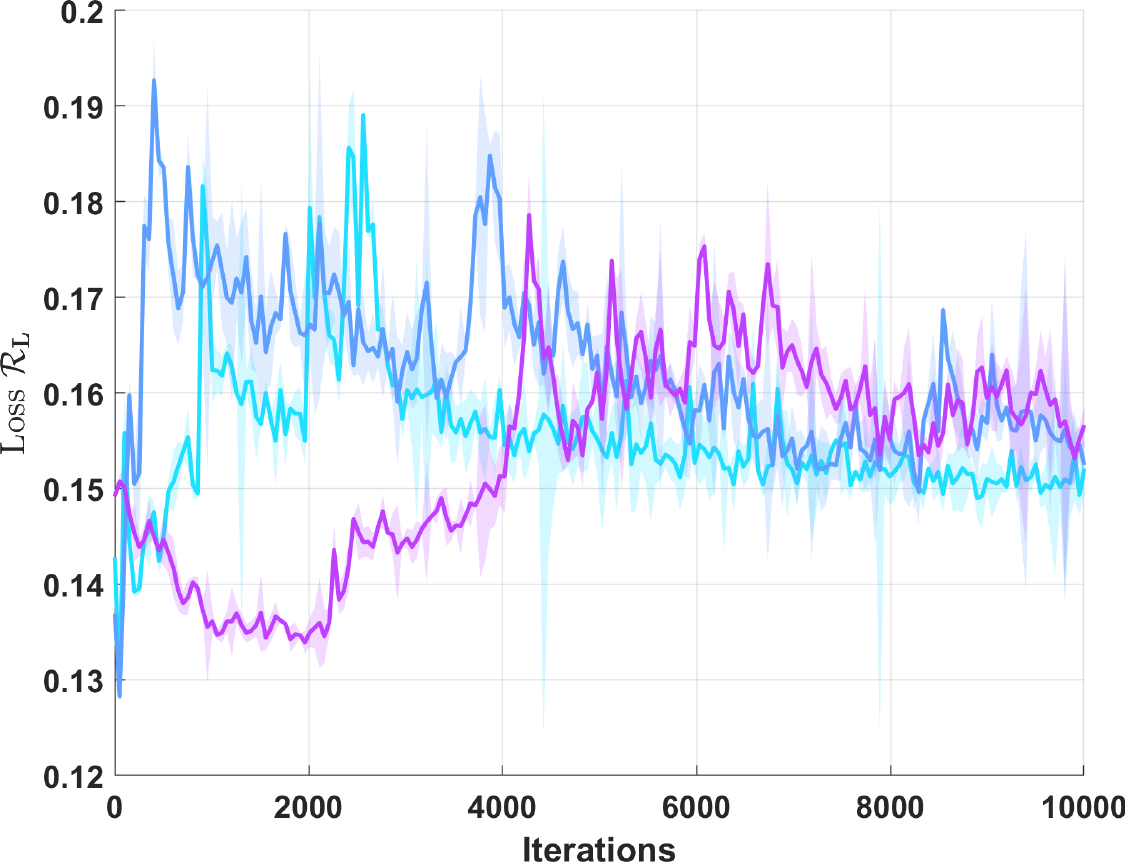}
	\end{minipage}~\label{costs}}
 
	\caption{The training dynamics give different ratios $p$ of active parameters on the Qwen-2.5 7B model.}
	\label{fig:diff-p}
    \vspace{-10pt}
\end{figure*}
\subsection{Language Modeling}

\begin{figure*}[ht]
	\centering
    \includegraphics[width=.85\textwidth]{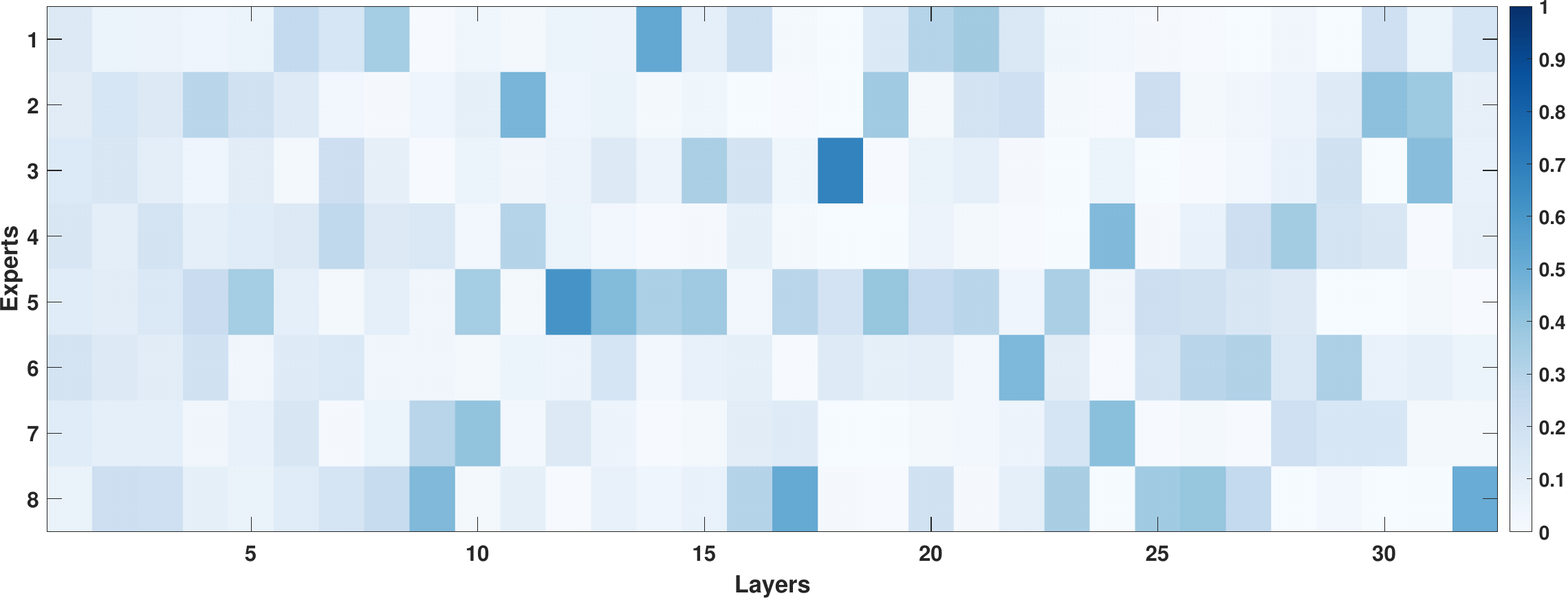}
    \caption{\revise{Experts token allocation of ToMoE for the LLaMA-3 8B model collected on the WikiText dataset.}}
    \label{fig:load_balance_map}
  \vspace{-5pt}
\end{figure*}

\begin{figure*}[ht]
	\centering
    \includegraphics[width=.9\textwidth]{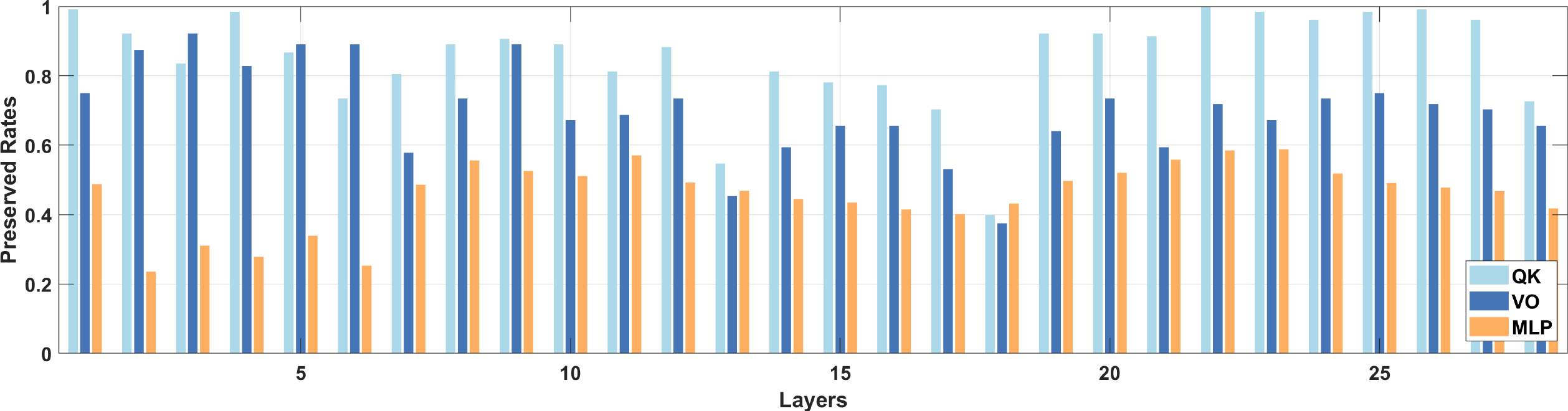}
    \caption{Model width after ToMoE for the Qwen-2.5 7B model when the number of active parameters equals 50\%.}
  \label{fig:width}
  %\vspace{-5pt}
\end{figure*}

% \begin{figure}[h]
% 	\centering
%     	\subfloat[]{
% 	\begin{minipage}[b]{.45\linewidth}
% 			\centering
% 			\includegraphics[width=.99\textwidth]{exp_figures/model_arch_qwen25_7b_uline.png}
% 	\end{minipage}}
%         \hfill
% 	\subfloat[]{
% 		\begin{minipage}[b]{.45\linewidth}
% 			\centering
% 			\includegraphics[width=.99\textwidth]{exp_figures/training_costs.png}
% 	\end{minipage}}
%     \caption{(a) Model width and the union of all experts. (b) Costs of different learning-based methods.}
%     \label{fig:add_analysis}
%     \vspace{-5pt}
% \end{figure}

% \begin{figure}[h]
% 	\centering
%     \includegraphics[width=.9\columnwidth]{exp_figures/Ablation_Study_new.pdf}
%     \caption{Ablation Study on Qwen-2.5 7B.}
%   \label{fig:ablation_study}
%   \vspace{-5pt}
% \end{figure}
\begin{figure}[ht]
\centering

\subfloat[Width and union of experts]{%
\includegraphics[width=0.31\linewidth]{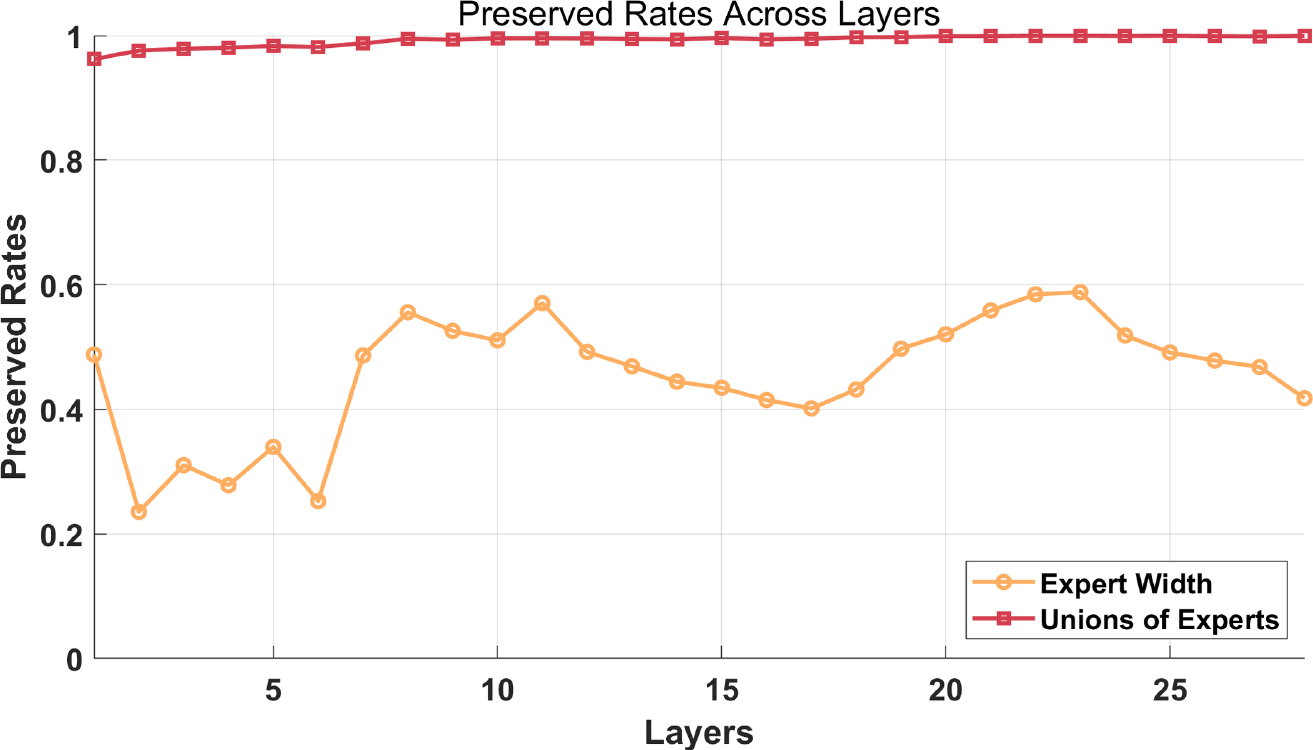}%
\label{fig:model_union}
}
\hfill
\subfloat[Training costs]{%
\includegraphics[width=0.31\linewidth]{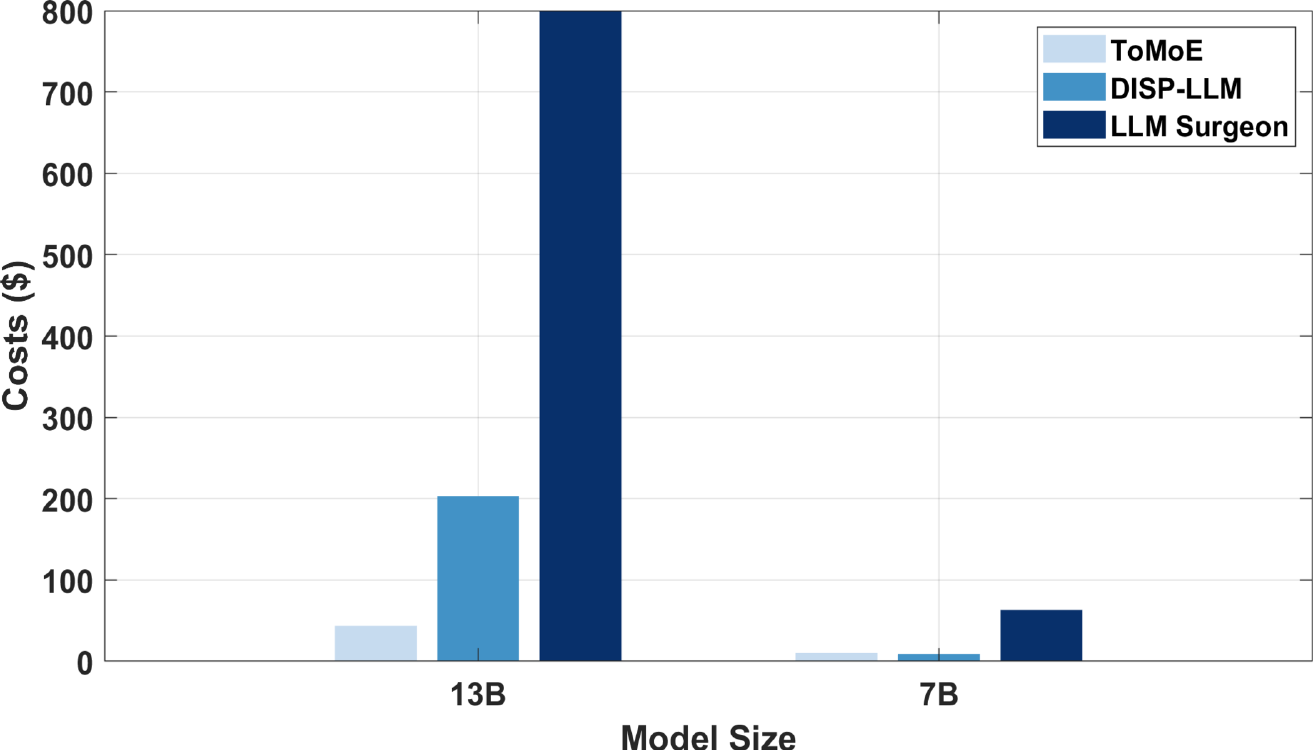}%
\label{fig:training_costs}
}
\hfill
\subfloat[Ablation study on Qwen-2.5 7B]{%
\includegraphics[width=0.31\linewidth]{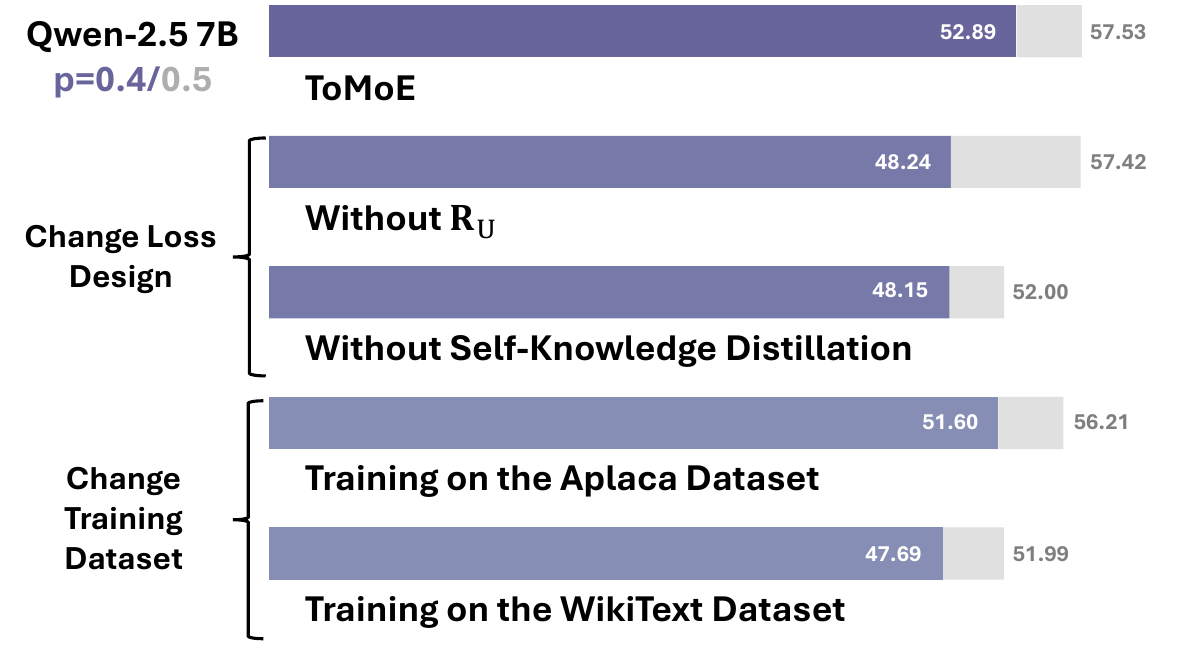}%
\label{fig:ablation_study}
}

\caption{(a) Model width and union of experts. (b) Costs of different learning-based methods. (c) Ablation study on Qwen-2.5 7B.}
\label{fig:combined_analysis}
\vspace{-5pt}
\end{figure}

% at pruning ratios of 30\%, 40\%, and 50\%, which correspond to active parameter percentages of 70\%, 60\%, and 50\%, respectively
Tab.~\ref{tab:lm} presents the perplexity results of structured pruning methods applied to LLaMA-2 models of sizes 7B and 13B on the WikiText-2 dataset, comparing various methods with 70\%, 60\%, and 50\% of active parameters setting—corresponding to pruning ratios of 30\%, 40\%, and 50\% for pruning, respectively. Across all pruning ratios, ToMoE consistently achieves the lowest perplexity compared to other methods, even outperforming many approaches with significantly larger numbers of active parameters. For instance, ToMoE with 50\% active parameters achieves a perplexity of 8.36, which is superior to LLM-Pruner, ShortGPT, SLEB, and SliceGPT at a 30\% pruning ratio. Furthermore, ToMoE with 50\% active parameters surpasses ModeGPT and LLM Surgeon at a 40\% pruning ratio. While the gap between ToMoE and DISP-LLM is smaller, it is still obvious at a 50\% pruning ratio: ToMoE achieves a perplexity that is 1.48 points lower than DISP-LLM. 
ToMoE also exhibits superior performance with the LLaMA-2 13B model, maintaining a similar advantage over other methods as observed with the LLaMA-2 7B model. This demonstrates the effectiveness of ToMoE in maintaining strong language modeling performance, even with much fewer active parameters. 
% ToMoE also demonstrates superior performance with the LLaMA-2 13B model, showing a similar advantage over other methods as observed with the LLaMA-2 7B model. demonstrate the effectiveness of ToMoE in maintaining strong language modeling performance, even with much fewer active parameters. 
Tab.~\ref{tab:Semi} presents a comparison of our method against semi-structural pruning techniques. Our approach consistently achieves the lowest perplexity while retaining 50\% of the active parameters. Moreover, the performance gap between our method and the semi-structural pruning methods is also obvious. On the LLaMA-2 7B model, SparseGPT achieves the second-best performance, with our method improving upon it by 1.81 in terms of perplexity. For the LLaMA-2 13B model, Pruner-Zero shows the second-best performance, while ToMoE further reduces the perplexity by 0.63. The comparison against semi-structural pruning methods further demonstrates the advantage of our method on the language modeling task.
\vspace{-5pt}
\subsection{Zero-Shot and Few-Shot Performance}
% In Tab.~\ref{tbl:zero_shot_main}, we present the zero-shot performance of different methods on LLaMA-2 7B, LLaMA-3 8B, and Qwen-2.5 7B. It is obvious that our method achieves the best average performance on all models. Compared to relatively weak methods like ShortGPT and SliceGPT, our method can achieve a higher average performance (ToMoE 50\%: 56.07, SliceGPT and ShortGPT 30\%: 53.73, 53.25) while having 20\% less active parameters on the LLaMA-2 7B model.  Compared to stronger LLM Surgeon and DISP-LLM, ToMoE achieves a better performance with 10\% less active parameters on the LLaMA-2 7B model, for example, ToMoE 50\% vs LLM Surgeon 40\%, ToMoE 40\% vs LLM Surgeon 30\% and DISP-LLM 30\%. ModeGPT is closer to ToMoE, but the gap is still large. When removing 40\% active parameters, ToMoE is 3.14 better than ModeGPT. When removing 50\% of active parameters, ToMoE is 5.02 and 10.39 better than DISP-LLM and LLM Surgeon respectively. The performance gap between our method and other baselines on LLaMA-3 8B is even larger, ToMoE reduces 5\% more active parameters than ModeGPT while it is still 2.71 better than it. When removing 15\% more active parameters than ShortGPT and SliceGPT, the average performance of ToMoE is 18.93 and 16 higher than them. 
In Tab.~\ref{tbl:zero_shot_main}, we present the zero-shot performance of various methods on LLaMA-2 7B, LLaMA-3 8B, and Qwen-2.5 7B. Our method consistently achieves the best average performance across all models. For \textbf{LLaMA-2 7B}, compared to weaker methods like ShortGPT and SliceGPT, our approach demonstrates significant advantages (ToMoE 50\%: 60.72 vs. SliceGPT 40.84 and ShortGPT 70\%: 47.07). The advantage against stronger baselines is also obvious. Although ModeGPT performs closer to ToMoE, the gap remains significant. With 60\% active parameters, ToMoE is 3.14 times better than ModeGPT. For \textbf{LLaMA-3 8B}, the performance advantage of ToMoE is even larger, where it reduces 5\% more active parameters than ModeGPT while still achieving a 2.71 performance gain. Furthermore, when removing 15\% more active parameters compared to ShortGPT and SliceGPT, ToMoE exceeds their average performance by 18.93 and 16 points, respectively. For \textbf{Qwen-2.5 7B}, ToMoE significantly outperforms DISP-LLM, consistent with previous findings on other models. We further investigate the effect of the number of experts $N$ when 40\% to 50\% of the parameters are active. The results indicate that increasing the number of experts to 16 is beneficial. However, further increasing $N$ to 24 provides only marginal or no improvement, likely because a too-large number of experts burdens the learning process. \revise{Thus, we recommend choosing the number of experts $N$ to be smaller than 16.}

In Tab.~\ref{tbl:zero_shot_dense_to_moe}, our method demonstrated superb advantages compared to existing MoE construction methods. In ``+fine-tuning" setting of LLaMA-MoE, the resulting model is trained for the same number of iterations as ToMoE for updating model weights.  \revise{In Tab.~\ref{tab:come_comparasion} and Tab.~\ref{tab:vs_llama_moe_v2}, we further compare our method with {CMoE}, {LLaMA-MoE}, and {LLaMA-MoE-v2}, following the experimental settings in their papers. Our approach consistently surpasses all baselines while requiring significantly fewer tokens. Notably, in Tab.~\ref{tab:vs_llama_moe_v2}, when compared against the fully trained {LLaMA-MoE-v2}, our method achieves comparable performance even without additional fine-tuning of the model weights. In summary, our method shows that learning routers and experts together is a more promising solution compared to existing works. 
}

%Compared to relatively weaker methods like ShortGPT and SliceGPT, our approach demonstrates higher average performance (ToMoE 50\%: 56.07 vs. SliceGPT 70\%: 53.73 and ShortGPT 70\%: 53.25) while utilizing 20\% fewer active parameters on the LLaMA-2 7B model. Against stronger baselines such as LLM Surgeon and DISP-LLM, ToMoE achieves better performance with 10\% fewer active parameters on the same model. For instance, ToMoE (50\%) outperforms LLM Surgeon (60\%), and ToMoE (60\%) outperforms both LLM Surgeon (70\%) and DISP-LLM (70\%). Although ModeGPT performs closer to ToMoE, the gap remains significant. With 60\% active parameters, ToMoE is 3.14 better than ModeGPT. In addition, when keeping 50\% active parameters, ToMoE outperforms DISP-LLM and LLM Surgeon by 5.02 and 10.39, respectively. The performance advantage of ToMoE is even larger on LLaMA-3 8B, where it reduces 5\% more active parameters than ModeGPT while still achieving a 2.71 performance gain. Furthermore, when removing 15\% more active parameters compared to ShortGPT and SliceGPT, ToMoE exceeds their average performance by 18.93 and 16 points, respectively.

\subsection{Analysis of ToMoE}
\textbf{Training Dynamics.} In Fig.\ref{fig:diff-p}, we visualize the training dynamics under different values of $p$. Across all $p$, the knowledge distillation loss $\mathcal{L}$ (Fig.\ref{fig:diff-p}(a)), the parameter regularization loss $\mathcal{R}_{\text{P}}$, and the union of experts regularization loss $\mathcal{R}_{\text{U}}$ decrease over the course of training. Notably, the parameter regularization loss quickly drops to $0$ in the early stages of training, while using a smaller $p$ requires more iterations. The peak of the union of experts regularization loss increases when using smaller values of $p$, indicating that the initial solution tends to only cover a small portion of the dense model. Regarding the load balancing loss, it oscillates around 0.15, demonstrating that ToMoE maintains a relatively balanced load distribution during the training process.

\textbf{Ablation Study.} In Fig.~\ref{fig:ablation_study},  we present the average zero-shot task performance under different settings. For $p=0.4$ and $p=0.5$, replacing the knowledge distillation loss with the language modeling loss significantly impacts performance. At $p=0.4$, removing $\mathcal{R}_{\text{U}}$ also results in a substantial performance drop, whereas the impact is much smaller at $p=0.5$. We hypothesize that this difference arises because reducing $p$ makes the learning process more challenging. Without the guidance provided by $\mathcal{R}_{\text{U}}$, the model struggles to effectively utilize the parameters of the original model. Additionally, the choice of dataset affects performance, particularly when switching to the WikiText dataset. This demonstrates that a mixing dataset is beneficial to the overall performance. 

\textbf{Other Analysis.} \textbf{(1).} Fig.~\ref{fig:width} presents the width of our ToMoE model for Qwen-2.5 7B, which shows the layer-wise configuration is highly non-uniform. It demonstrates that our method can flexibly set the width of different layers and operations. \textbf{(2).} Fig.~\ref{fig:model_union} shows that the union of experts is close to the full model capacity, even though the width of experts across different layers is highly non-uniform, demonstrating the effectiveness of our loss design. \textbf{(3).}  Fig.~\ref{fig:training_costs} plots the costs of different learning-based methods in terms of US dollars. ToMoE costs similarly compared to DISP-LLM with LLaMA-2 7B and 13B models, and both of them are much cheaper than LLM Surgeon. \revise{\textbf{(4).} Fig.~\ref{fig:load_balance_map} shows the token allocation across experts on the Wikitext dataset. We observe that the early and late layers exhibit relatively balanced expert utilization, while the middle layers have certain experts activated more frequently.
}   \textbf{(5).} Finally, we visualize the expert selection for LLaMA-2 7B in Tab.~\ref{tab:gen_sample}. We can observe that each expert aligns syntax rather than semantic meanings, resembling the observations in~\citep{jiang2024mixtral}.
\section{Conclusion}
In this paper, we propose a novel algorithm, ToMoE, for converting dense models into MoE models through dynamic structural pruning. The resulting MoE models significantly outperform state-of-the-art structural pruning methods while using similar or lower training costs compared to other learning-based pruning methods. Our findings reveal the presence of meaningful experts within the MLP layers of dense models, even without fine-tuning the model weights. ToMoE serves as a powerful tool for uncovering these experts within the original dense LLM.

\bibliography{main}
\bibliographystyle{tmlr}

\newpage
\appendix
\section{Details of trainable modules\label{sec:app-modules}}
\subsection{Module Configurations}
We present the details of trainable modules in Tab.~\ref{tab:modules}. In short, we project the input tokens to a low-dimensional space and add them to the output of the HyperNetwork. The inputs $z$ to the HyperNetwork are fixed random vectors of size $N\times32$ sampled from a Normal Distribution. Except for the HyperNetwork, other individual trainable modules are created for each MHA and MLP layer. If we have $L$ blocks, then we will have $L$ $\text{Proj}_{\text{E}}^{\text{\tiny MHA}}$, $L$ $\text{Proj}_{\text{D}}^{\text{\tiny MHA}}$ with output size of $\frac{d}{H}$, $L$ $\text{Proj}_{\text{D}}^{\text{\tiny MHA}}$ with output size of $\frac{d}{2H}$ $L$, $\text{Proj}_{\text{D}}^{\text{\tiny MLP}}$ and $L$ $\text{Router}$ layers. Notations of $d$, $H$, $d_{\text{mid}}$, and $N$ are already defined in Sec.~\ref{sec:arch_basics}.

\begin{table}[h]
      \centering
      \resizebox{0.7\textwidth}{!}{
        \begin{tabular}{l|c|c}
            \toprule
            Module Types & Removed? & Structures\\
            \midrule
            HyperNetwork & \cmark &Input $z$ $\rightarrow$ Bi-GRU(32,64)\\
            \midrule 
                  $\text{Proj}_{\text{E}}^{\text{\tiny MHA}}$ & \xmark & Linear($d$, $d_e$ = 128) \\
            \midrule
                  $\text{Proj}_{\text{D}}^{\text{\tiny MHA}}$ & \xmark &  LayerNorm($d_e$ = 128)$\rightarrow$ GeLU$\rightarrow$Linear($d_e$ = 128, $\frac{d}{H}$)  \\
            \midrule
                  $\text{Proj}_{\text{D}}^{\text{\tiny MLP}}$ & \cmark & LayerNorm($d_e$ = 128)$\rightarrow$ GeLU$\rightarrow$Linear($d_e$ = 128, $d_{\text{mid}}$)\\
            \midrule
                  $\text{Router}$ & \xmark  & Linear($d$, $N$) \\
            \bottomrule
          \end{tabular}
          }
          \caption{Detailed configuration of trainable modules.}
          \label{tab:modules}
\end{table}

After we complete the training of ToMoE, we do not have to preserve all modules. The embeddings from the HyperNetwork will be saved, so the HyperNetwork can be removed without impacting the model. $\text{Proj}_{\text{D}}^{\text{\tiny MLP}}$ brings most additional parameters, fortunately, it can also be removed. After the training of ToMoE, $\mathbf{E}$ and $\text{Proj}_{\text{D}}^{\text{\tiny MLP}}$ can be used to directly generate experts:
\begin{equation}~\label{eq:final_expert}
    \mathbf{s}_{\scriptstyle{\mathbf{e}}} = \text{ST-GSig}(\text{Proj}_{\text{D}}^{\text{\tiny MLP}}(\mathbf{E})),
\end{equation}
where $\text{ST-GSig}$ again is the Straight-Through Gumbel-Sigmoid function. $ \mathbf{s}_{\scriptstyle{\mathbf{e}}} \in \{0,1\}^{N\times d_{\text{mid}}}$ is the resulting binary vectors to select experts from the dense model. Once $\mathbf{s}_{\scriptstyle{\mathbf{e}}}$ is generated, it can be reused, and thus we no longer need $\text{Proj}_{\text{D}}^{\text{\tiny MLP}}$. Let $\mathbf{S}_{\scriptstyle{\mathbf{e}}}^i = \text{Diag}(\mathbf{s}_{\scriptstyle{\mathbf{e}}}^i),\ i=1,\cdots,N$. Similarly, we use $\hat{\mathbf{s}}_{\scriptstyle{\mathbf{e}}}^i\in \Re^{d_{\text{mid}}\times d_{\text{mid}}'}$ to represent the actual column or row selection matrix by removing zero columns or rows, where $d_{\text{mid}}' < d_{\text{mid}}$ and it is the width of each expert. The $i$th expert can be represented as:
\begin{equation}
    f_{\text{MLP}}^i(\mathbf{X}) = 
\sigma(\mathbf{X}\mathbf{W}_G\hat{\mathbf{S}}_{\scriptstyle{\mathbf{e}}}^i)\odot (\mathbf{X}\mathbf{W}_U\hat{\mathbf{S}}_{\scriptstyle{\mathbf{e}}}^i) \hat{\mathbf{S}}_{\scriptstyle{\mathbf{e}}}^i\mathbf{W}_D.\\
\end{equation}~\label{eq:single_expert}
After ToMoE, given the result of the routing function $\mathbf{G} = {\text{ST-GSmax}}(\text{Router}(\mathbf{X}))$, the MLP calculation with MoE can be written as:
\begin{equation}
    \mathbf{Y}_t = \mathbf{G}_{t,i} f_{\text{MLP}}^i(\mathbf{X}_t),
    ~\label{eq:mlp_moe_actual}
\end{equation}
where $\mathbf{X}_t$ is the feature map of $t$th token, and $i$ represents the index where $\mathbf{G}_{t,i}=1$. Note that Eq.~\ref{eq:mlp_moe_actual} is still differentiable with respect to the parameters of the $\text{Router}$.

Another question is how many parameters we need after introducing Top-K routing for MHA layers and Top-1 routing for MLP layers. Analytically, the additional parameters can be calculated by $L \times d \times 128 +  L \times 128 \times (\frac{d}{H}) + L\times d \times N$. Let's use LLaMA-2 7B as an example, $L=32$, $d=4096$, $N=8$, $H=32$, the additional parameters are $1\times 32 \times 4096 \times 128 +  32 \times 128 \times(128) +32\times 4096 \times 8 = 0.0184$B. This is equivalent to $0.27\%$ of the total parameters of the LLaMA-2 7B model, and thus, the additional parameter is not significant compared to the original number of parameters.
%The term $(\frac{d}{H}+\frac{d}{2H})$ will be explained in the Sec.~\ref{sec:app-rope}.

\subsection{Head Dimension Pruning vs. RoPE}~\label{sec:app-rope}
Rotary Position Embedding (RoPE)~\citep{su2024roformer} is a popular positional encoding method, and it is regularly used in LLMs like LLaMA~\citep{touvron2023llama2}. RoPE divides the $\frac{d}{H}$ dimensional space into $\frac{d}{2H}$ sub-spaces, and they are applied on query and key. This means that if we want to perform head dimension pruning for query and key matrices, we need to follow the sub-spaces resulting from RoPE and make these two sub-spaces share the same pruning mask $\mathbf{s}_{0}$, and the final pruning mask for query and key is $\mathbf{s}_0' = [{\mathbf{s}_{0}}_{[1:\frac{d}{2H}]}, {\mathbf{s}_{0}}_{[1:\frac{d}{2H}]}]$, and clearly the size of ${\mathbf{s}_{0}}_{[1:\frac{d}{2H}]}$ is $\frac{d}{2H}$. In short, we simply select the first half of elements from $\mathbf{s}_0$ and repeat it twice to make the final pruning decision. We also found that applying dynamic pruning for query and key matrices along the head dimension is difficult and unreasonable since different tokens may have different positions after pruning. It becomes a problem when calculating the inner product between the query and key matrices given different tokens.

By applying head dimension pruning, our method also does not need to be modified when facing different attention mechanisms like GQA (Grouped-Query Attention)~\citep{ainslie2023gqa} and MQA (Multi-Query Attention)~\citep{shazeer2019fast}.  
%This also explains why $\text{Proj}_{\text{D}}^{\text{\tiny MHA}}$ in Tab.~\ref{tab:modules} has two output dimensions: $\frac{d}{H}$ for output and value matrices and $\frac{d}{2H}$ for query and key matrices.
% \begin{table}[t!]
% \centering
% \caption{Zero-shot task performance of compressed LLaMA-2 13B,  Qwen-2.5 7B. \label{tbl:zero_shot_main}}
% \setlength{\extrarowheight}{2pt}
% \resizebox{\textwidth}{!}{
% \begin{tabular}{c|c|l|c|c|c|c|c|c}
% \cmidrule[1pt]{1-9}
% % Model & Compression & Method & ARC-e & ARC-c & PIQA & WinoG. & HellaS. & Average \\ 
% \multirow{2}{*}{Model} & \multirow{2}{*}{Compression} & \multirow{2}{*}{Method} &  ARC-e & ARC-c & PIQA & WinoG. & HellaS. & \multirow{2}{*}{Average} \\ 
% \cmidrule{4-8}
%  &  &  & acc-norm & acc-norm & acc-norm & acc & acc-norm & \\
% \cmidrule[1pt]{1-9}

% \multirow{6}{*}{LLaMA-2 13B} 
% & 0\% & Dense & 74.58 & 46.25 & 79.11 & 69.06 & 75.99 & 69.00 \\ \cmidrule{2-9}

% \end{tabular}
% }
% \end{table}

\begin{table}[t!]
\centering
\caption{Zero-shot task performance of compressed LLaMA-2 13B and Qwen-2.5 14B. \label{tbl:zero_shot_large}}
\setlength{\extrarowheight}{2pt}
\resizebox{\textwidth}{!}{
\begin{tabular}{c|c|l|c|c|c|c|c|c}
\cmidrule[1pt]{1-9}
\multirow{2}{*}{Model} & \multirow{2}{*}{Active Parameters} & \multirow{2}{*}{Method} & ARC-e & ARC-c & PIQA & WinoG. & HellaS. & \multirow{2}{*}{Average} \\ 
\cmidrule{4-8}
 &  &  & acc-norm & acc-norm & acc-norm & acc & acc-norm & \\
\cmidrule[1pt]{1-9}

\multirow{18}{*}{LLaMA-2 13B} 
& 100\%  & Dense                & 77.48 & 49.23 & 80.47 & 72.22 & 79.39 & 71.76 \\ \cmidrule{2-9}

& \multirow{6}{*}{70\%} 
    & SliceGPT~\citep{ashkboos2023slicegpt}        & 60.27 & 36.18 & 69.42 & 64.09 & 49.74 & 55.94 \\
&   & LLM Surgeon~\citep{van2023llm}               & 69.74 & 40.27 & 76.50 & 68.67 & 71.52 & 65.34 \\
&   & DISP-LLM~\citep{gaodisp}                     & 63.80 & 39.42 & 74.43 & 66.85 & 70.86 & 63.07 \\ 
&   & DISP-LLM Alpaca~\citep{gaodisp}              & 68.98 & 44.28 & 77.31 & 67.32 & 68.98 & 65.59 \\
&   & MoDeGPT-Alpaca~\citep{lin2024modegpt}        & 70.24 & 41.47 & 77.15 & 71.27 & 71.84 & 66.39 \\
&   &\cellcolor{tabcolor}\textbf{ToMoE (Ours)}    &\cellcolor{tabcolor}\textbf{74.58} &\cellcolor{tabcolor}\textbf{45.14} 
&\cellcolor{tabcolor}\textbf{77.97} &\cellcolor{tabcolor}\textbf{75.44} &\cellcolor{tabcolor}\textbf{68.19} &\cellcolor{tabcolor}\textbf{68.26} \\ \cmidrule{2-9}

& \multirow{6}{*}{60\%} 
    & SliceGPT~\citep{ashkboos2023slicegpt}        & 48.99 & 32.51 & 63.17 & 56.75 & 39.85 & 48.25 \\
&   & LLM Surgeon~\citep{van2023llm}               & 63.80 & 37.12 & 73.16 & 65.75 & 65.64 & 60.94 \\
&   & DISP-LLM~\citep{gaodisp}                     & 62.67 & 35.63 & 73.39 & 62.67 & 65.86 & 60.04 \\
&   & DISP-LLM Alpaca~\citep{gaodisp}              & 66.79 & 42.75 & 75.30 & 64.25 & 67.52 & 63.32 \\
&   & MoDeGPT-Alpaca~\citep{lin2024modegpt}        & 63.72 & 38.82 & 71.87 & 66.30 & 62.10 & 60.56 \\
&   & \cellcolor{tabcolor}\textbf{ToMoE (Ours)}   &\cellcolor{tabcolor}\textbf{67.63} &\cellcolor{tabcolor}\textbf{41.81}
&\cellcolor{tabcolor}\textbf{75.79} &\cellcolor{tabcolor}\textbf{66.38} &\cellcolor{tabcolor}\textbf{73.41} &\cellcolor{tabcolor}\textbf{65.00} \\ \cmidrule{2-9}

& \multirow{5}{*}{50\%} 
    & LLM Surgeon~\citep{van2023llm}               & 56.19 & 37.12 & 68.87 & 63.22 & 56.19 & 56.32 \\
&   & DISP-LLM~\citep{gaodisp}                     & 58.27 & 36.87 & 68.67 & 59.27 & 57.18 & 54.50 \\
&   & DISP-LLM Alpaca~\citep{gaodisp}              & 55.72 & 37.54 & 72.20 & 59.59 & 62.39 & 57.49 \\
&   &\cellcolor{tabcolor}\textbf{ToMoE (Ours)}    &\cellcolor{tabcolor}\textbf{63.51} &\cellcolor{tabcolor}\textbf{37.97} 
&\cellcolor{tabcolor}\textbf{73.29} &\cellcolor{tabcolor}\textbf{62.59} &\cellcolor{tabcolor}\textbf{68.30} &\cellcolor{tabcolor}\textbf{61.13} \\ \cmidrule[1pt]{1-9}

\multirow{5}{*}{Qwen-2.5 14B} 
& 100\%  & Dense                                     & 79.42 & 59.13 & 83.09 & 74.98 & 82.10 & 75.74 \\ \cmidrule{2-9}
& \multirow{2}{*}{50\%} 
    & DISP-LLM~\citep{gaodisp}                     & 65.87 & 37.99 & 72.20 & 58.51 & 60.63 & 59.04 \\
&   & \cellcolor{tabcolor}\textbf{ToMoE (Ours)}   & \cellcolor{tabcolor}\textbf{66.37} & \cellcolor{tabcolor}\textbf{39.16} 
& \cellcolor{tabcolor}\textbf{74.59} & \cellcolor{tabcolor}\textbf{60.69} & \cellcolor{tabcolor}\textbf{65.57} & \cellcolor{tabcolor}\textbf{61.28} \\ \cmidrule{2-9}

& \multirow{2}{*}{40\%} 
    & DISP-LLM~\citep{gaodisp}                     & 56.36 & 32.62 & 70.08 & 54.22 & 51.50 & 52.96 \\
&   & \cellcolor{tabcolor}\textbf{ToMoE (Ours)}   & \cellcolor{tabcolor}\textbf{59.85} & \cellcolor{tabcolor}\textbf{34.73} 
& \cellcolor{tabcolor}\textbf{71.49} & \cellcolor{tabcolor}\textbf{57.62} & \cellcolor{tabcolor}\textbf{56.60} & \cellcolor{tabcolor}\textbf{56.06} \\
\cmidrule[1pt]{1-9}
\end{tabular}
}
\end{table}
\subsection{Details of Gumbel-Softmax and Gumbel-Sigmoid}
The Gumbel-Softmax function~\citep{jang2016categorical} allows for differentiable sampling from a categorical distribution. Given logits $\mathbf{x}$, the Gumbel-Softmax sample $\mathbf{y}$ is computed as:
\[
\mathbf{y} = \text{softmax}\left(\frac{\mathbf{x} + \mathbf{g}}{\tau}\right),
\]
where each element of $\mathbf{g}$ is drawn from $\text{Gumbel}(0, 1)$, and $\tau$ is the temperature parameter that controls the smoothness of the distribution. Combining Gumbel-Softmax with the Straight-Through gradient Estimator~\citep{bengio2013estimating}, we have the following equation:
\begin{equation}~\label{eq:STG}
    \text{ST-GSmax}(\mathbf{x}) = \text{one-hot}\left(\arg\max_{i\in D} \left[ \frac{x_i + g_i}{\tau} \right]\right)
\end{equation}
where $D=\{1,2,\cdots,N\}$, $N$ again is the number of experts in our setting, and $\text{one-hot}$ will assign $1$ corresponding to the position of the maximum value in $\frac{\mathbf{x} + \mathbf{g}}{\tau}$ and assign $0$ to other positions. 

The Gumbel-Sigmoid function is a special case of the Gumbel-Softmax function, designed for binary distributions. Given logits $\mathbf{x}$, the Gumbel-Sigmoid sample $\mathbf{y}$ is computed as:
\[
\mathbf{y} = \text{sigmoid}\left(\frac{\mathbf{x} + \mathbf{g}}{\tau}\right),
\]
where $\mathbf{g}$ is sampled from $\text{Gumbel}(0, 1)$ and $\tau$ again is the temperature parameter. Combining with the Straight-Through gradient Estimator, we have the following equation:
\begin{equation}~\label{eq:STGS}
    \text{ST-GSig}(\mathbf{x}) = \text{round}( \text{sigmoid}\left(\frac{\mathbf{x} + \mathbf{g}+b}{\tau}\right)),
\end{equation}
where $b$ is a constant bias in our implementation and it ensures that all experts start from the whole model, $\text{round}(\cdot)$ will round the input values to the nearest integer, and in our case, it rounds inputs to $0$ or $1$. For all experiments, we set $b=3.0$ in Eq.~\ref{eq:STGS}, and we set $\tau=0.4$ for Eq.~\ref{eq:STG} and Eq.~\ref{eq:STGS}.

\begin{table}[t!]
\centering
\caption{Zero-shot task performance of the compressed Phi-2. \label{tbl:zero_shot_phi2}}
\setlength{\extrarowheight}{2pt}
\resizebox{0.8\textwidth}{!}{
\begin{tabular}{c|l|c|c|c|c|c|c}
\cmidrule[1pt]{1-8}
\multirow{2}{*}{Active Parameters} & \multirow{2}{*}{Method} & ARC-e & ARC-c & PIQA & WinoG. & HellaS. & \multirow{2}{*}{Avg} \\ 
\cmidrule{3-7}
 &  & acc-norm & acc-norm & acc-norm & acc & acc-norm & \\
\cmidrule[1pt]{1-8}

\multirow{1}{*}{100\%} 
     & Dense                           & 78.24 & 54.01 & 79.11 & 75.61 & 73.86 & 72.17 \\ \cmidrule{1-8}

\multirow{3}{*}{80\%} 
     & SliceGPT~\citep{ashkboos2023slicegpt}    & 58.00 & 35.32 & 71.87 & 67.80 & 57.76 & 58.15 \\
     & +Fine-tuning                   & 56.61 & 38.91 & 71.27 & 67.17 & 54.86 & 57.76 \\
     & DISP-LLM~\citep{gaodisp}        & 68.18 & \textbf{44.11} & 74.86 & 67.09 & \textbf{62.93} & 63.43 \\ \cmidrule{1-8}

\multirow{3}{*}{75\%} 
     & SliceGPT~\citep{ashkboos2023slicegpt}    & 53.70 & 31.66 & 69.21 & 65.35 & 52.40 & 54.46 \\
     & +Fine-tuning                   & 52.78 & 35.49 & 69.91 & 65.19 & 52.48 & 55.17 \\
     & DISP-LLM~\citep{gaodisp}        & 65.93 & 43.34 & 74.27 & 65.11 & 59.95 & 61.72 \\ \cmidrule{1-8}
     
\multirow{4}{*}{70\%} 
     & SliceGPT~\citep{ashkboos2023slicegpt}    & 53.03 & 30.29 & 65.94 & 63.14 & 47.56 & 51.99 \\
     & +Fine-tuning                   & 46.38 & 32.68 & 66.16 & 63.54 & 49.72 & 51.70 \\
     & DISP-LLM~\citep{gaodisp}        & 63.59 & 38.48 & 73.34 & 65.19 & 54.43 & 59.00 \\
     & \cellcolor{tabcolor}\textbf{ToMoE (Ours)}  & \cellcolor{tabcolor}\textbf{70.79} & \cellcolor{tabcolor}43.86 & \cellcolor{tabcolor}\textbf{77.09} & \cellcolor{tabcolor}\textbf{66.38} & \cellcolor{tabcolor}62.68 & \cellcolor{tabcolor}\textbf{64.16} \\ 
\cmidrule[1pt]{1-8}
\end{tabular}
}
\end{table}

\begin{lstlisting}[language=Python, keywords={model, helper, torch, self, None, True, False}, xleftmargin=0.1\linewidth,                   
linewidth=0.9\linewidth,
caption={Pseudo-code for self-knowledge distillation.}, label={lst:self-kd}]
with torch.no_grad():
    # Disable trainable modules for ToMoE
    helper.set_module_status(model, False)
    
    # Get logits from the teacher (original model)
    teacher_output = model(inputs)
    teacher_logits = teacher_output.logits
    
    # Enable trainable modules for ToMoE
    helper.set_module_status(model, True)

# Get logits from the student model from ToMoE learning process
model_output = model(inputs)
logits = model_output.logits
\end{lstlisting}

\section{More Details of the Loss Design}
\subsection{Implementation of the Self-Knowledge Distillation}
During the ToMoE learning process, we freeze the parameters of the original model. This approach offers the additional benefit of enabling self-knowledge distillation without the need to load an extra model. In Lst.~\ref{lst:self-kd}, we present the pseudo-code for the self-knowledge distillation process. In summary, we first disable the trainable modules associated with ToMoE and compute the output logits from the original model. Next, we re-enable the trainable modules for ToMoE and perform a regular forward pass. The logits from the original model are then used to guide the learning of ToMoE.

\begin{table}[t]
\centering
\caption{ToMoE Visualization of LLaMA-2 7B with 50\% active parameters}\label{tab:moe_sample}
\setlength\extrarowheight{2pt}
\resizebox{1\linewidth}{!}{
\begin{tabular}{|l|p{16cm}|}
\hline
Expert Color & {\setlength{\fboxsep}{0pt}\sethlcolor{color0}\hl{Expert 1} \sethlcolor{color1}\hl{Expert 2} \sethlcolor{color2}\hl{Expert 3}  \sethlcolor{color3}\hl{Expert 4}  \sethlcolor{color4}\hl{Expert 5} \sethlcolor{color5}\hl{Expert 6} \sethlcolor{color6}\hl{Expert 7} \sethlcolor{color7}\hl{Expert 8}
}
\\
\hline
MLP 1 &  
{\setlength{\fboxsep}{0pt}
\hl{\texttt{<s>}}\sethlcolor{color2}\hl{ Hom}\sethlcolor{color2}\hl{arus}\sethlcolor{color4}\hl{ gam}\sethlcolor{color4}\hl{mar}\sethlcolor{color4}\hl{us}\sethlcolor{color2}\hl{,}\sethlcolor{color7}\hl{ known}\sethlcolor{color2}\hl{ as}\sethlcolor{color2}\hl{ the}\sethlcolor{color2}\hl{ European}\sethlcolor{color2}\hl{ lo}\sethlcolor{color0}\hl{b}\sethlcolor{color0}\hl{ster}\sethlcolor{color2}\hl{ or}\sethlcolor{color6}\hl{ common}\sethlcolor{color5}\hl{ lo}\sethlcolor{color3}\hl{b}\sethlcolor{color2}\hl{ster}\sethlcolor{color1}\hl{,}\sethlcolor{color2}\hl{ is}\sethlcolor{color0}\hl{ a}\sethlcolor{color2}\hl{ species}\sethlcolor{color2}\hl{ of}\sethlcolor{color7}\hl{ cla}\sethlcolor{color1}\hl{wed}\sethlcolor{color2}\hl{ lo}\sethlcolor{color7}\hl{b}\sethlcolor{color2}\hl{ster}\sethlcolor{color6}\hl{ from}\sethlcolor{color2}\hl{ the}\sethlcolor{color2}\hl{ eastern}\sethlcolor{color0}\hl{ Atlantic}\sethlcolor{color2}\hl{ Ocean}\sethlcolor{color4}\hl{,}\sethlcolor{color6}\hl{ Mediter}\sethlcolor{color7}\hl{rane}\sethlcolor{color7}\hl{an}\sethlcolor{color6}\hl{ Sea}\sethlcolor{color2}\hl{ and}\sethlcolor{color4}\hl{ parts}\sethlcolor{color3}\hl{ of}\sethlcolor{color2}\hl{ the}\sethlcolor{color7}\hl{ Black}\sethlcolor{color6}\hl{ Sea}\sethlcolor{color0}\hl{.}\sethlcolor{color6}\hl{ It}\sethlcolor{color1}\hl{ is}\sethlcolor{color5}\hl{ closely}\sethlcolor{color7}\hl{ related}\sethlcolor{color5}\hl{ to}\sethlcolor{color4}\hl{ the}\sethlcolor{color3}\hl{ American}\sethlcolor{color3}\hl{ lo}\sethlcolor{color1}\hl{b}\sethlcolor{color7}\hl{ster}\sethlcolor{color2}\hl{,}\sethlcolor{color3}\hl{ H}\sethlcolor{color5}\hl{.}\sethlcolor{color6}\hl{ amer}\sethlcolor{color2}\hl{ican}\sethlcolor{color2}\hl{us}\sethlcolor{color5}\hl{.}\sethlcolor{color1}\hl{ It}\sethlcolor{color4}\hl{ may}\sethlcolor{color2}\hl{ grow}\sethlcolor{color4}\hl{ to}\sethlcolor{color5}\hl{ a}\sethlcolor{color7}\hl{ length}\sethlcolor{color4}\hl{ of}\sethlcolor{color3}\hl{ }\sethlcolor{color3}\hl{ 6}\sethlcolor{color0}\hl{0}\sethlcolor{color6}\hl{ cm}\sethlcolor{color0}\hl{ (}\sethlcolor{color5}\hl{2}\sethlcolor{color7}\hl{4}\sethlcolor{color4}\hl{ in}\sethlcolor{color5}\hl{)}\sethlcolor{color2}\hl{ and}\sethlcolor{color5}\hl{ a}\sethlcolor{color7}\hl{ mass}\sethlcolor{color3}\hl{ of}\sethlcolor{color2}\hl{ }\sethlcolor{color0}\hl{ 6}\sethlcolor{color2}\hl{ kil}\sethlcolor{color1}\hl{og}\sethlcolor{color7}\hl{rams}\sethlcolor{color0}\hl{ (}\sethlcolor{color5}\hl{1}\sethlcolor{color4}\hl{3}\sethlcolor{color1}\hl{ lb}\sethlcolor{color3}\hl{),}\sethlcolor{color1}\hl{ and}\sethlcolor{color5}\hl{ be}\sethlcolor{color2}\hl{ars}\sethlcolor{color0}\hl{ a}\sethlcolor{color7}\hl{ consp}\sethlcolor{color6}\hl{ic}\sethlcolor{color6}\hl{uous}\sethlcolor{color7}\hl{ pair}\sethlcolor{color3}\hl{ of}\sethlcolor{color7}\hl{ cla}\sethlcolor{color4}\hl{ws}\sethlcolor{color3}\hl{.}\sethlcolor{color2}\hl{ In}\sethlcolor{color1}\hl{ life}\sethlcolor{color3}\hl{,}\sethlcolor{color5}\hl{ the}\sethlcolor{color4}\hl{ lo}\sethlcolor{color4}\hl{bst}\sethlcolor{color3}\hl{ers}\sethlcolor{color3}\hl{ are}\sethlcolor{color7}\hl{ blue}\sethlcolor{color3}\hl{,}\sethlcolor{color6}\hl{ only}\sethlcolor{color4}\hl{ becoming}\sethlcolor{color3}\hl{ '}\sethlcolor{color5}\hl{lob}\sethlcolor{color7}\hl{ster}\sethlcolor{color4}\hl{ red}\sethlcolor{color2}\hl{'}\sethlcolor{color3}\hl{ on}\sethlcolor{color6}\hl{ cook}\sethlcolor{color6}\hl{ing}\sethlcolor{color5}\hl{.}\sethlcolor{color6}\hl{ M}\sethlcolor{color2}\hl{ating}\sethlcolor{color5}\hl{ occurs}\sethlcolor{color3}\hl{ in}\sethlcolor{color5}\hl{ the}\sethlcolor{color6}\hl{ summer}\sethlcolor{color4}\hl{,}\sethlcolor{color6}\hl{ producing}\sethlcolor{color0}\hl{ eggs}\sethlcolor{color0}\hl{ which}\sethlcolor{color0}\hl{ are}\sethlcolor{color7}\hl{ carried}\sethlcolor{color6}\hl{ by}\sethlcolor{color5}\hl{ the}\sethlcolor{color5}\hl{ females}\sethlcolor{color5}\hl{ for}\sethlcolor{color3}\hl{ up}\sethlcolor{color3}\hl{ to}\sethlcolor{color4}\hl{ a}\sethlcolor{color2}\hl{ year}\sethlcolor{color0}\hl{ before}\sethlcolor{color6}\hl{ h}\sethlcolor{color3}\hl{atch}\sethlcolor{color7}\hl{ing}\sethlcolor{color4}\hl{ into}\sethlcolor{color2}\hl{ pl}\sethlcolor{color7}\hl{ank}\sethlcolor{color6}\hl{ton}\sethlcolor{color7}\hl{ic}\sethlcolor{color6}\hl{ lar}\sethlcolor{color7}\hl{va}\sethlcolor{color0}\hl{e}\sethlcolor{color3}\hl{.}\sethlcolor{color3}\hl{ Hom}\sethlcolor{color0}\hl{arus}\sethlcolor{color7}\hl{ gam}\sethlcolor{color6}\hl{mar}\sethlcolor{color2}\hl{us}\sethlcolor{color2}\hl{ is}\sethlcolor{color1}\hl{ a}\sethlcolor{color3}\hl{ highly}\sethlcolor{color3}\hl{ este}\sethlcolor{color3}\hl{emed}\sethlcolor{color6}\hl{ food}\sethlcolor{color1}\hl{,}\sethlcolor{color0}\hl{ and}\sethlcolor{color2}\hl{ is}\sethlcolor{color1}\hl{ widely}\sethlcolor{color6}\hl{ caught}\sethlcolor{color7}\hl{ using}\sethlcolor{color1}\hl{ lo}\sethlcolor{color5}\hl{b}\sethlcolor{color7}\hl{ster}\sethlcolor{color6}\hl{ p}\sethlcolor{color7}\hl{ots}\sethlcolor{color2}\hl{,}\sethlcolor{color2}\hl{ mostly}\sethlcolor{color4}\hl{ around}\sethlcolor{color5}\hl{ the}\sethlcolor{color2}\hl{ British}\sethlcolor{color7}\hl{ Is}\sethlcolor{color1}\hl{les}\sethlcolor{color3}\hl{.}
}
\\
\hline
MLP 16 &  
{\setlength{\fboxsep}{0pt}
\hl{\texttt{<s>}}\sethlcolor{color7}\hl{ Hom}\sethlcolor{color7}\hl{arus}\sethlcolor{color7}\hl{ gam}\sethlcolor{color7}\hl{mar}\sethlcolor{color3}\hl{us}\sethlcolor{color3}\hl{,}\sethlcolor{color3}\hl{ known}\sethlcolor{color5}\hl{ as}\sethlcolor{color5}\hl{ the}\sethlcolor{color7}\hl{ European}\sethlcolor{color7}\hl{ lo}\sethlcolor{color7}\hl{b}\sethlcolor{color7}\hl{ster}\sethlcolor{color3}\hl{ or}\sethlcolor{color7}\hl{ common}\sethlcolor{color7}\hl{ lo}\sethlcolor{color7}\hl{b}\sethlcolor{color0}\hl{ster}\sethlcolor{color3}\hl{,}\sethlcolor{color1}\hl{ is}\sethlcolor{color3}\hl{ a}\sethlcolor{color2}\hl{ species}\sethlcolor{color1}\hl{ of}\sethlcolor{color7}\hl{ cla}\sethlcolor{color0}\hl{wed}\sethlcolor{color7}\hl{ lo}\sethlcolor{color7}\hl{b}\sethlcolor{color0}\hl{ster}\sethlcolor{color1}\hl{ from}\sethlcolor{color0}\hl{ the}\sethlcolor{color7}\hl{ eastern}\sethlcolor{color7}\hl{ Atlantic}\sethlcolor{color2}\hl{ Ocean}\sethlcolor{color1}\hl{,}\sethlcolor{color7}\hl{ Mediter}\sethlcolor{color7}\hl{rane}\sethlcolor{color7}\hl{an}\sethlcolor{color2}\hl{ Sea}\sethlcolor{color1}\hl{ and}\sethlcolor{color1}\hl{ parts}\sethlcolor{color1}\hl{ of}\sethlcolor{color1}\hl{ the}\sethlcolor{color7}\hl{ Black}\sethlcolor{color1}\hl{ Sea}\sethlcolor{color2}\hl{.}\sethlcolor{color0}\hl{ It}\sethlcolor{color0}\hl{ is}\sethlcolor{color6}\hl{ closely}\sethlcolor{color5}\hl{ related}\sethlcolor{color3}\hl{ to}\sethlcolor{color3}\hl{ the}\sethlcolor{color7}\hl{ American}\sethlcolor{color7}\hl{ lo}\sethlcolor{color7}\hl{b}\sethlcolor{color7}\hl{ster}\sethlcolor{color0}\hl{,}\sethlcolor{color7}\hl{ H}\sethlcolor{color7}\hl{.}\sethlcolor{color7}\hl{ amer}\sethlcolor{color7}\hl{ican}\sethlcolor{color3}\hl{us}\sethlcolor{color3}\hl{.}\sethlcolor{color2}\hl{ It}\sethlcolor{color2}\hl{ may}\sethlcolor{color1}\hl{ grow}\sethlcolor{color3}\hl{ to}\sethlcolor{color1}\hl{ a}\sethlcolor{color1}\hl{ length}\sethlcolor{color1}\hl{ of}\sethlcolor{color1}\hl{ }\sethlcolor{color1}\hl{ 6}\sethlcolor{color1}\hl{0}\sethlcolor{color1}\hl{ cm}\sethlcolor{color7}\hl{ (}\sethlcolor{color7}\hl{2}\sethlcolor{color7}\hl{4}\sethlcolor{color1}\hl{ in}\sethlcolor{color3}\hl{)}\sethlcolor{color5}\hl{ and}\sethlcolor{color1}\hl{ a}\sethlcolor{color7}\hl{ mass}\sethlcolor{color7}\hl{ of}\sethlcolor{color1}\hl{ }\sethlcolor{color1}\hl{ 6}\sethlcolor{color7}\hl{ kil}\sethlcolor{color7}\hl{og}\sethlcolor{color7}\hl{rams}\sethlcolor{color7}\hl{ (}\sethlcolor{color7}\hl{1}\sethlcolor{color7}\hl{3}\sethlcolor{color1}\hl{ lb}\sethlcolor{color3}\hl{),}\sethlcolor{color2}\hl{ and}\sethlcolor{color5}\hl{ be}\sethlcolor{color2}\hl{ars}\sethlcolor{color6}\hl{ a}\sethlcolor{color7}\hl{ consp}\sethlcolor{color5}\hl{ic}\sethlcolor{color5}\hl{uous}\sethlcolor{color2}\hl{ pair}\sethlcolor{color0}\hl{ of}\sethlcolor{color7}\hl{ cla}\sethlcolor{color3}\hl{ws}\sethlcolor{color4}\hl{.}\sethlcolor{color0}\hl{ In}\sethlcolor{color7}\hl{ life}\sethlcolor{color2}\hl{,}\sethlcolor{color5}\hl{ the}\sethlcolor{color7}\hl{ lo}\sethlcolor{color7}\hl{bst}\sethlcolor{color2}\hl{ers}\sethlcolor{color5}\hl{ are}\sethlcolor{color6}\hl{ blue}\sethlcolor{color2}\hl{,}\sethlcolor{color5}\hl{ only}\sethlcolor{color5}\hl{ becoming}\sethlcolor{color5}\hl{ '}\sethlcolor{color7}\hl{lob}\sethlcolor{color7}\hl{ster}\sethlcolor{color1}\hl{ red}\sethlcolor{color2}\hl{'}\sethlcolor{color5}\hl{ on}\sethlcolor{color7}\hl{ cook}\sethlcolor{color5}\hl{ing}\sethlcolor{color4}\hl{.}\sethlcolor{color0}\hl{ M}\sethlcolor{color2}\hl{ating}\sethlcolor{color6}\hl{ occurs}\sethlcolor{color5}\hl{ in}\sethlcolor{color6}\hl{ the}\sethlcolor{color7}\hl{ summer}\sethlcolor{color3}\hl{,}\sethlcolor{color6}\hl{ producing}\sethlcolor{color2}\hl{ eggs}\sethlcolor{color5}\hl{ which}\sethlcolor{color5}\hl{ are}\sethlcolor{color5}\hl{ carried}\sethlcolor{color5}\hl{ by}\sethlcolor{color6}\hl{ the}\sethlcolor{color3}\hl{ females}\sethlcolor{color5}\hl{ for}\sethlcolor{color7}\hl{ up}\sethlcolor{color5}\hl{ to}\sethlcolor{color5}\hl{ a}\sethlcolor{color7}\hl{ year}\sethlcolor{color5}\hl{ before}\sethlcolor{color7}\hl{ h}\sethlcolor{color7}\hl{atch}\sethlcolor{color7}\hl{ing}\sethlcolor{color7}\hl{ into}\sethlcolor{color7}\hl{ pl}\sethlcolor{color7}\hl{ank}\sethlcolor{color7}\hl{ton}\sethlcolor{color7}\hl{ic}\sethlcolor{color7}\hl{ lar}\sethlcolor{color7}\hl{va}\sethlcolor{color7}\hl{e}\sethlcolor{color7}\hl{.}\sethlcolor{color7}\hl{ Hom}\sethlcolor{color7}\hl{arus}\sethlcolor{color7}\hl{ gam}\sethlcolor{color7}\hl{mar}\sethlcolor{color2}\hl{us}\sethlcolor{color5}\hl{ is}\sethlcolor{color6}\hl{ a}\sethlcolor{color0}\hl{ highly}\sethlcolor{color7}\hl{ este}\sethlcolor{color0}\hl{emed}\sethlcolor{color1}\hl{ food}\sethlcolor{color0}\hl{,}\sethlcolor{color4}\hl{ and}\sethlcolor{color0}\hl{ is}\sethlcolor{color0}\hl{ widely}\sethlcolor{color0}\hl{ caught}\sethlcolor{color0}\hl{ using}\sethlcolor{color7}\hl{ lo}\sethlcolor{color7}\hl{b}\sethlcolor{color7}\hl{ster}\sethlcolor{color7}\hl{ p}\sethlcolor{color2}\hl{ots}\sethlcolor{color0}\hl{,}\sethlcolor{color0}\hl{ mostly}\sethlcolor{color0}\hl{ around}\sethlcolor{color0}\hl{ the}\sethlcolor{color7}\hl{ British}\sethlcolor{color7}\hl{ Is}\sethlcolor{color2}\hl{les}\sethlcolor{color4}\hl{.}
}
\\
\hline
MLP 32 &  
{\setlength{\fboxsep}{0pt}
\sethlcolor{color2}\hl{\texttt{<s>}}\sethlcolor{color7}\hl{ Hom}\sethlcolor{color1}\hl{arus}\sethlcolor{color7}\hl{ gam}\sethlcolor{color1}\hl{mar}\sethlcolor{color1}\hl{us}\sethlcolor{color1}\hl{,}\sethlcolor{color2}\hl{ known}\sethlcolor{color1}\hl{ as}\sethlcolor{color5}\hl{ the}\sethlcolor{color1}\hl{ European}\sethlcolor{color7}\hl{ lo}\sethlcolor{color7}\hl{b}\sethlcolor{color2}\hl{ster}\sethlcolor{color5}\hl{ or}\sethlcolor{color5}\hl{ common}\sethlcolor{color7}\hl{ lo}\sethlcolor{color7}\hl{b}\sethlcolor{color2}\hl{ster}\sethlcolor{color4}\hl{,}\sethlcolor{color6}\hl{ is}\sethlcolor{color5}\hl{ a}\sethlcolor{color6}\hl{ species}\sethlcolor{color5}\hl{ of}\sethlcolor{color7}\hl{ cla}\sethlcolor{color5}\hl{wed}\sethlcolor{color7}\hl{ lo}\sethlcolor{color7}\hl{b}\sethlcolor{color2}\hl{ster}\sethlcolor{color5}\hl{ from}\sethlcolor{color5}\hl{ the}\sethlcolor{color5}\hl{ eastern}\sethlcolor{color2}\hl{ Atlantic}\sethlcolor{color2}\hl{ Ocean}\sethlcolor{color4}\hl{,}\sethlcolor{color7}\hl{ Mediter}\sethlcolor{color7}\hl{rane}\sethlcolor{color2}\hl{an}\sethlcolor{color2}\hl{ Sea}\sethlcolor{color4}\hl{ and}\sethlcolor{color0}\hl{ parts}\sethlcolor{color4}\hl{ of}\sethlcolor{color5}\hl{ the}\sethlcolor{color2}\hl{ Black}\sethlcolor{color2}\hl{ Sea}\sethlcolor{color0}\hl{.}\sethlcolor{color2}\hl{ It}\sethlcolor{color6}\hl{ is}\sethlcolor{color6}\hl{ closely}\sethlcolor{color3}\hl{ related}\sethlcolor{color4}\hl{ to}\sethlcolor{color5}\hl{ the}\sethlcolor{color5}\hl{ American}\sethlcolor{color7}\hl{ lo}\sethlcolor{color7}\hl{b}\sethlcolor{color2}\hl{ster}\sethlcolor{color4}\hl{,}\sethlcolor{color6}\hl{ H}\sethlcolor{color1}\hl{.}\sethlcolor{color7}\hl{ amer}\sethlcolor{color7}\hl{ican}\sethlcolor{color3}\hl{us}\sethlcolor{color0}\hl{.}\sethlcolor{color2}\hl{ It}\sethlcolor{color4}\hl{ may}\sethlcolor{color2}\hl{ grow}\sethlcolor{color4}\hl{ to}\sethlcolor{color5}\hl{ a}\sethlcolor{color3}\hl{ length}\sethlcolor{color6}\hl{ of}\sethlcolor{color1}\hl{ }\sethlcolor{color7}\hl{ 6}\sethlcolor{color7}\hl{0}\sethlcolor{color3}\hl{ cm}\sethlcolor{color7}\hl{ (}\sethlcolor{color7}\hl{2}\sethlcolor{color7}\hl{4}\sethlcolor{color7}\hl{ in}\sethlcolor{color3}\hl{)}\sethlcolor{color2}\hl{ and}\sethlcolor{color5}\hl{ a}\sethlcolor{color2}\hl{ mass}\sethlcolor{color6}\hl{ of}\sethlcolor{color7}\hl{ }\sethlcolor{color7}\hl{ 6}\sethlcolor{color7}\hl{ kil}\sethlcolor{color6}\hl{og}\sethlcolor{color2}\hl{rams}\sethlcolor{color6}\hl{ (}\sethlcolor{color1}\hl{1}\sethlcolor{color7}\hl{3}\sethlcolor{color7}\hl{ lb}\sethlcolor{color4}\hl{),}\sethlcolor{color2}\hl{ and}\sethlcolor{color6}\hl{ be}\sethlcolor{color5}\hl{ars}\sethlcolor{color5}\hl{ a}\sethlcolor{color7}\hl{ consp}\sethlcolor{color7}\hl{ic}\sethlcolor{color5}\hl{uous}\sethlcolor{color6}\hl{ pair}\sethlcolor{color5}\hl{ of}\sethlcolor{color7}\hl{ cla}\sethlcolor{color3}\hl{ws}\sethlcolor{color1}\hl{.}\sethlcolor{color5}\hl{ In}\sethlcolor{color5}\hl{ life}\sethlcolor{color5}\hl{,}\sethlcolor{color5}\hl{ the}\sethlcolor{color7}\hl{ lo}\sethlcolor{color7}\hl{bst}\sethlcolor{color4}\hl{ers}\sethlcolor{color4}\hl{ are}\sethlcolor{color3}\hl{ blue}\sethlcolor{color4}\hl{,}\sethlcolor{color4}\hl{ only}\sethlcolor{color5}\hl{ becoming}\sethlcolor{color1}\hl{ '}\sethlcolor{color7}\hl{lob}\sethlcolor{color5}\hl{ster}\sethlcolor{color3}\hl{ red}\sethlcolor{color2}\hl{'}\sethlcolor{color5}\hl{ on}\sethlcolor{color5}\hl{ cook}\sethlcolor{color3}\hl{ing}\sethlcolor{color5}\hl{.}\sethlcolor{color5}\hl{ M}\sethlcolor{color5}\hl{ating}\sethlcolor{color2}\hl{ occurs}\sethlcolor{color5}\hl{ in}\sethlcolor{color5}\hl{ the}\sethlcolor{color2}\hl{ summer}\sethlcolor{color4}\hl{,}\sethlcolor{color5}\hl{ producing}\sethlcolor{color3}\hl{ eggs}\sethlcolor{color4}\hl{ which}\sethlcolor{color6}\hl{ are}\sethlcolor{color2}\hl{ carried}\sethlcolor{color5}\hl{ by}\sethlcolor{color5}\hl{ the}\sethlcolor{color2}\hl{ females}\sethlcolor{color5}\hl{ for}\sethlcolor{color2}\hl{ up}\sethlcolor{color2}\hl{ to}\sethlcolor{color5}\hl{ a}\sethlcolor{color2}\hl{ year}\sethlcolor{color5}\hl{ before}\sethlcolor{color5}\hl{ h}\sethlcolor{color7}\hl{atch}\sethlcolor{color3}\hl{ing}\sethlcolor{color6}\hl{ into}\sethlcolor{color7}\hl{ pl}\sethlcolor{color6}\hl{ank}\sethlcolor{color5}\hl{ton}\sethlcolor{color5}\hl{ic}\sethlcolor{color7}\hl{ lar}\sethlcolor{color3}\hl{va}\sethlcolor{color3}\hl{e}\sethlcolor{color5}\hl{.}\sethlcolor{color6}\hl{ Hom}\sethlcolor{color5}\hl{arus}\sethlcolor{color6}\hl{ gam}\sethlcolor{color7}\hl{mar}\sethlcolor{color2}\hl{us}\sethlcolor{color6}\hl{ is}\sethlcolor{color5}\hl{ a}\sethlcolor{color5}\hl{ highly}\sethlcolor{color7}\hl{ este}\sethlcolor{color5}\hl{emed}\sethlcolor{color5}\hl{ food}\sethlcolor{color6}\hl{,}\sethlcolor{color5}\hl{ and}\sethlcolor{color6}\hl{ is}\sethlcolor{color6}\hl{ widely}\sethlcolor{color6}\hl{ caught}\sethlcolor{color5}\hl{ using}\sethlcolor{color7}\hl{ lo}\sethlcolor{color7}\hl{b}\sethlcolor{color5}\hl{ster}\sethlcolor{color6}\hl{ p}\sethlcolor{color3}\hl{ots}\sethlcolor{color5}\hl{,}\sethlcolor{color2}\hl{ mostly}\sethlcolor{color4}\hl{ around}\sethlcolor{color5}\hl{ the}\sethlcolor{color2}\hl{ British}\sethlcolor{color7}\hl{ Is}\sethlcolor{color3}\hl{les}\sethlcolor{color1}\hl{.}
}
\\
\hline
\end{tabular}
}
\end{table} 

\begin{table*}[t!]
\centering
\caption{Zero-shot task performance of compressed LLaMA-2 7B with more settings. \label{tbl:zero_shot_llama2}}
\setlength{\extrarowheight}{2pt}
\resizebox{0.9\textwidth}{!}{
\begin{tabular}{c|l|c|c|c|c|c|c}
\cmidrule[1pt]{1-8}
% Active Params & Method & ARC-e & ARC-c & PIQA & WinoG. & HellaS. & Average \\ 
\multirow{2}{*}{Active Parameters} & \multirow{2}{*}{Method} & ARC-e & ARC-c & PIQA & WinoG. & HellaS. & \multirow{2}{*}{Average} \\ 
\cmidrule{3-7}
 &  & acc-norm & acc-norm & acc-norm & acc & acc-norm & \\
\cmidrule[1pt]{1-8}

100\% & Dense & 74.58 & 46.25 & 79.11 & 69.06 & 75.99 & 69.00 \\ 
\cmidrule{1-8}

\multirow{7}{*}{70\%} 
& ShortGPT \citep{men2024shortgpt} & 48.65 & 32.85 & 64.31 & 64.33 & 56.13 & 53.25 \\ 
& SliceGPT \citep{ashkboos2023slicegpt} & 58.88 & 33.36 & 68.55 & 58.01 & 49.86 & 53.73 \\ 
& LLM Surgeon \citep{van2023llm} & 63.09 & 36.69 & 73.56 & 61.09 & 60.72 & 59.03 \\
& DISP-LLM ~\citep{gaodisp} & 59.81 & 33.19 & 71.82 & 62.27 & 63.43 & 58.10 \\ 
& DISP-LLM Alpaca~\citep{gaodisp} & 60.10 & 37.03 & 73.72 & 63.93 & 62.87 & 59.53 \\ 
& ModeGPT~\citep{lin2024modegpt} & 63.26 & 38.73 & 70.40 & 67.32 & 63.26 & 60.78 \\ 
& ModeGPT-Alpaca~\citep{lin2024modegpt} & 65.49 & 39.16 & 73.34 & 66.22 & 65.90 & 62.02 \\ 
\cmidrule{1-8}

\multirow{6}{*}{60\%} 
& ShortGPT \citep{men2024shortgpt} & 41.16 & 29.94 & 60.12 & 60.46 & 43.67 & 47.07 \\ 
& SliceGPT \citep{ashkboos2023slicegpt} & 36.49 & 24.57 & 54.90 & 53.43 & 34.80 & 40.84 \\ 
& LLM Surgeon \citep{van2023llm} & 52.31 & 30.29 & 69.26 & 54.38 & 48.04 & 50.86 \\ 
& ModeGPT~\citep{lin2024modegpt} & 49.45 & 30.03 & 64.96 & 61.96 & 53.01 & 51.88 \\ 
& ModeGPT-Alpaca~\citep{lin2024modegpt} & 59.76 & 34.73 & 70.35 & \textbf{64.40} & 58.63 & 57.58 \\ 
& \cellcolor{tabcolor} \textbf{ToMoE (Ours)} & \cellcolor{tabcolor} \textbf{63.64} & \cellcolor{tabcolor} \textbf{38.74} & \cellcolor{tabcolor} \textbf{72.85} & \cellcolor{tabcolor} 62.51 & \cellcolor{tabcolor} \textbf{65.84} & \cellcolor{tabcolor} \textbf{60.72} \\ 
\cmidrule{1-8}

\multirow{4}{*}{50\%}
& LLM Surgeon ~\citep{van2023llm} & 44.91 & 26.28 & 64.36 & 52.57 & 40.29 & 45.68 \\ 
& DISP-LLM ~\citep{gaodisp} & 43.06 & 25.85 & 63.93 & 54.54 & 63.43 & 46.72 \\ 
& DISP-LLM Alpaca~\citep{gaodisp} & 51.14 & 30.20 & 68.34 & 56.20 & 49.35 & 51.05 \\ 
& \cellcolor{tabcolor} \textbf{ToMoE (Ours)} & \cellcolor{tabcolor} \textbf{56.65} & \cellcolor{tabcolor} \textbf{33.87} & \cellcolor{tabcolor} \textbf{71.00} & \cellcolor{tabcolor} \textbf{58.56} & \cellcolor{tabcolor} \textbf{60.26} & \cellcolor{tabcolor} \textbf{56.07} \\ 
\cmidrule[1pt]{1-8}
\end{tabular}
}
\vspace{-10pt}
\end{table*}

\subsection{Efficient Implementation of $\mathcal{R}_\mathbf{u}$}
Recall from Eq.~\ref{eq:union} that the union regularization for MLP and MHA layers is defined as:
\begin{equation*}
    \mathcal{R}_\mathbf{u} = \bigcup_{i=1}^T \mathbf{s}_i = 1 - \prod_{i=1}^T (1 - \mathbf{s}_i).
\end{equation*}
For MLP layers, this equation incurs high computational costs since $\mathbf{s} \in \mathbb{R}^{T \times d_{\text{mid}}}$, whereas for MHA layers, the cost is significantly lower because $\frac{d}{H} \ll d_{\text{mid}}$. To simplify Eq.~\ref{eq:union}, note that all $\mathbf{s}_i$ (\(i=1,\ldots,T\)) are derived from \(N\) experts. Using embeddings from the hypernetwork, we calculate the configuration of \(N\) experts as:
\begin{equation*}
    \mathbf{s}_{\mathbf{e}} = \text{ST-GSig}(\text{Proj}_{\text{D}}^{\text{\tiny MLP}}(\mathbf{E})),
\end{equation*}
and substitute $\mathbf{s}_{\mathbf{e}}$ into Eq.~\ref{eq:union}:
\begin{equation}~\label{eq:union-mlp}
    \mathcal{R}_\mathbf{u}^{\text{\tiny MLP}} = \bigcup_{i=1}^N \mathbf{s}_{\scriptstyle{\mathbf{e}}}^i = 1 - \prod_{i=1}^N (1 - \mathbf{s}_{\scriptstyle{\mathbf{e}}}^i).
\end{equation}
This reduces computation by a factor of $\frac{T}{N}$. For example, in LLaMA-2, the computational cost is reduced by $\frac{2048}{8} = 256$ times.
\begin{table}[h]
\caption{Ablation study on design choices of ToMoE and the impact of temperature $\tau$ on performance.}
\centering
\begin{tabular}{lcccccc}
\toprule
\textbf{Settings} & \textbf{ARC-e} & \textbf{ARC-c} & \textbf{PIQA} & \textbf{WinoG.} & \textbf{HellaS.} & \textbf{Avg} \\
\midrule
Local Emb & 55.22 & 32.94 & 66.32 & 53.12 & 57.85 & 53.09 \\
Head Pruning & 45.12 & 25.85 & 64.82 & 49.49 & 39.65 & 44.99 \\
w/o VO Routing & 55.51 & 32.34 & 70.40 & 57.14 & 59.12 & 54.90 \\
$\tau=0.3$ & 57.28 & 33.53 & 70.35 & 57.54 & 60.22 & 55.78 \\
$\tau=0.4$ & 56.65 & 33.87 & 71.00 & 58.56 & 60.26 & 56.07 \\
$\tau=0.5$ & 55.89 & 33.36 & 71.00 & 56.75 & 59.31 & 55.16 \\
\midrule
ToMoE & 56.65 & 33.87 & 71.00 & 58.56 & 60.26 & 56.07 \\
\bottomrule
\end{tabular}
\label{tab:tomoe_design_choices}
\vspace{-10pt}
\end{table}
\subsection{Equivalence of MoE and pseudo-MoE}~\label{sec:app-pmoe}
One major challenge when training MoE models is maintaining an appropriate expert capacity, defined as the number of tokens each expert processes~\citep{fedus2022switch}. This is typically addressed using a load balancing loss. Without this loss, some experts may become overloaded while others remain underutilized, leading to bottlenecks where a few experts dominate the computation.

Although ToMoE also requires load balancing loss, the potential overhead introduced by load balancing is mitigated by the pseudo-MoE approach after ToMoE. After applying ToMoE, the resulting model can be trained using pseudo-MoE, which resembles the training of a dense model. This is straightforward to implement as follows:
\begin{equation}
    f_{\text{MLP}}(\mathbf{X}) = 
\sigma(\mathbf{X}\mathbf{W}_G){\mathbf{S}} \odot (\mathbf{X}\mathbf{W}_U{\mathbf{S}}) {\mathbf{S}}\mathbf{W}_D,
\end{equation}~\label{eq:pseudo-MoE}
where $\mathbf{S}_i$ in $\mathbf{S}$ represents the routed expert from $\mathbf{S}_e$ in Eq.~\ref{eq:final_expert}, as determined by the router. The pseudo-MoE is useful when the active number of parameters is relatively large. In such cases, pseudo-MoE training can be more time-efficient than conventional MoE training.

\section{More Implementation Details \label{sec:app-imp}}
During training the modules of ToMoE, we use AdamW optimizer to optimize it with a constant learning rate $10^{-3}$ and weight decay $0.05$. For different models, we always set the mini-batchsize to $1$ on each GPU. For LLaMA-2 7B, and Qwen-2.5 7B models, we use 2 NVIDIA A100 GPUs, For LLaMA-3 8B, we use 3 NVIDIA A100 GPUs. For LLaMA-2 13B and Qwen-2.5 14B models, we use 4 NVIDIA A100 GPUs. For all the rest models, we use 1 NVIDIA A100 GPU. We set $p=\{0.6,0.5,0.4,0.3\}$ when the ratios of active parameters equals to $\{40\%, 50\%,60\%,70\%\}$. 

For the Alpaca dataset~\footnote{https://huggingface.co/datasets/tatsu-lab/alpaca}, we use the `text' column within the dataset, which combines the columns of `instruction' and `output'. For the Code Alpaca dataset~\footnote{https://github.com/sahil280114/codealpaca}, we combine the `instruction', `input', and `output' columns as one training sample.

\begin{table}[t]
\centering
\caption{ToMoE Visualization of LLaMA-2 7B with 50\% active parameters (continued).}\label{tab:moe_sample_math}
\setlength\extrarowheight{2pt}
\resizebox{1\linewidth}{!}{
\begin{tabular}{|l|p{16cm}|}
\hline
Expert Color & {\setlength{\fboxsep}{0pt}\sethlcolor{color0}\hl{Expert 1} \sethlcolor{color1}\hl{Expert 2} \sethlcolor{color2}\hl{Expert 3}  \sethlcolor{color3}\hl{Expert 4}  \sethlcolor{color4}\hl{Expert 5} \sethlcolor{color5}\hl{Expert 6} \sethlcolor{color6}\hl{Expert 7} \sethlcolor{color7}\hl{Expert 8}
}
\\
\hline
MLP 1 &
{\setlength{\fboxsep}{0pt}
\sethlcolor{color4}\hl{\texttt{<s>}}\sethlcolor{color3}\hl{ Find}\sethlcolor{color2}\hl{ the}\sethlcolor{color2}\hl{ equation}\sethlcolor{color2}\hl{ of}\sethlcolor{color3}\hl{ the}\sethlcolor{color2}\hl{ line}\sethlcolor{color2}\hl{ passing}\sethlcolor{color4}\hl{ through}\sethlcolor{color0}\hl{ the}\sethlcolor{color3}\hl{ points}\sethlcolor{color2}\hl{ (}\sethlcolor{color0}\hl{3}\sethlcolor{color0}\hl{,}\sethlcolor{color3}\hl{ }\sethlcolor{color2}\hl{ 5}\sethlcolor{color5}\hl{)}\sethlcolor{color3}\hl{ and}\sethlcolor{color7}\hl{ (}\sethlcolor{color1}\hl{7}\sethlcolor{color3}\hl{,}\sethlcolor{color1}\hl{ }\sethlcolor{color1}\hl{ 9}\sethlcolor{color2}\hl{)}\sethlcolor{color4}\hl{ using}\sethlcolor{color0}\hl{ the}\sethlcolor{color1}\hl{ slope}\sethlcolor{color5}\hl{-}\sethlcolor{color4}\hl{inter}\sethlcolor{color6}\hl{cept}\sethlcolor{color3}\hl{ form}\sethlcolor{color2}\hl{ of}\sethlcolor{color4}\hl{ a}\sethlcolor{color3}\hl{ linear}\sethlcolor{color2}\hl{ equation}\sethlcolor{color2}\hl{.}\sethlcolor{color2}\hl{ To}\sethlcolor{color6}\hl{ find}\sethlcolor{color3}\hl{ the}\sethlcolor{color1}\hl{ equation}\sethlcolor{color3}\hl{ of}\sethlcolor{color0}\hl{ the}\sethlcolor{color3}\hl{ line}\sethlcolor{color6}\hl{ passing}\sethlcolor{color6}\hl{ through}\sethlcolor{color5}\hl{ the}\sethlcolor{color2}\hl{ points}\sethlcolor{color1}\hl{ (}\sethlcolor{color1}\hl{3}\sethlcolor{color6}\hl{,}\sethlcolor{color2}\hl{ }\sethlcolor{color6}\hl{ 5}\sethlcolor{color1}\hl{)}\sethlcolor{color1}\hl{ and}\sethlcolor{color4}\hl{ (}\sethlcolor{color7}\hl{7}\sethlcolor{color0}\hl{,}\sethlcolor{color0}\hl{ }\sethlcolor{color4}\hl{ 9}\sethlcolor{color4}\hl{)}\sethlcolor{color6}\hl{ using}\sethlcolor{color4}\hl{ the}\sethlcolor{color6}\hl{ slope}\sethlcolor{color0}\hl{-}\sethlcolor{color6}\hl{inter}\sethlcolor{color2}\hl{cept}\sethlcolor{color1}\hl{ form}\sethlcolor{color4}\hl{ (}\sethlcolor{color6}\hl{y}\sethlcolor{color4}\hl{ =}\sethlcolor{color2}\hl{ m}\sethlcolor{color7}\hl{x}\sethlcolor{color2}\hl{ +}\sethlcolor{color1}\hl{ b}\sethlcolor{color7}\hl{),}\sethlcolor{color0}\hl{ we}\sethlcolor{color7}\hl{ first}\sethlcolor{color2}\hl{ need}\sethlcolor{color2}\hl{ to}\sethlcolor{color6}\hl{ find}\sethlcolor{color5}\hl{ the}\sethlcolor{color7}\hl{ slope}\sethlcolor{color6}\hl{ (}\sethlcolor{color7}\hl{m}\sethlcolor{color3}\hl{)}\sethlcolor{color0}\hl{ and}\sethlcolor{color0}\hl{ the}\sethlcolor{color0}\hl{ y}\sethlcolor{color1}\hl{-}\sethlcolor{color7}\hl{inter}\sethlcolor{color6}\hl{cept}\sethlcolor{color0}\hl{ (}\sethlcolor{color6}\hl{b}\sethlcolor{color5}\hl{).}\sethlcolor{color1}\hl{ }\sethlcolor{color4}\hl{ 1}\sethlcolor{color5}\hl{.}\sethlcolor{color2}\hl{ Find}\sethlcolor{color2}\hl{ the}\sethlcolor{color3}\hl{ slope}\sethlcolor{color6}\hl{ (}\sethlcolor{color6}\hl{m}\sethlcolor{color3}\hl{):}\sethlcolor{color6}\hl{ m}\sethlcolor{color4}\hl{ =}\sethlcolor{color2}\hl{ (}\sethlcolor{color6}\hl{y}\sethlcolor{color2}\hl{2}\sethlcolor{color0}\hl{ -}\sethlcolor{color4}\hl{ y}\sethlcolor{color3}\hl{1}\sethlcolor{color1}\hl{)}\sethlcolor{color1}\hl{ /}\sethlcolor{color1}\hl{ (}\sethlcolor{color7}\hl{x}\sethlcolor{color1}\hl{2}\sethlcolor{color7}\hl{ -}\sethlcolor{color7}\hl{ x}\sethlcolor{color6}\hl{1}\sethlcolor{color4}\hl{)}\sethlcolor{color2}\hl{ m}\sethlcolor{color4}\hl{ =}\sethlcolor{color0}\hl{ (}\sethlcolor{color3}\hl{9}\sethlcolor{color0}\hl{ -}\sethlcolor{color3}\hl{ }\sethlcolor{color2}\hl{ 5}\sethlcolor{color6}\hl{)}\sethlcolor{color7}\hl{ /}\sethlcolor{color4}\hl{ (}\sethlcolor{color1}\hl{7}\sethlcolor{color0}\hl{ -}\sethlcolor{color6}\hl{ }\sethlcolor{color6}\hl{ 3}\sethlcolor{color0}\hl{)}\sethlcolor{color2}\hl{ m}\sethlcolor{color4}\hl{ =}\sethlcolor{color0}\hl{ }\sethlcolor{color7}\hl{ 4}\sethlcolor{color0}\hl{ /}\sethlcolor{color3}\hl{ }\sethlcolor{color1}\hl{ 4}\sethlcolor{color1}\hl{ m}\sethlcolor{color7}\hl{ =}\sethlcolor{color6}\hl{ }\sethlcolor{color6}\hl{ 1}\sethlcolor{color3}\hl{ }\sethlcolor{color2}\hl{ 2}\sethlcolor{color0}\hl{.}\sethlcolor{color6}\hl{ Use}\sethlcolor{color2}\hl{ one}\sethlcolor{color1}\hl{ of}\sethlcolor{color5}\hl{ the}\sethlcolor{color2}\hl{ points}\sethlcolor{color6}\hl{ to}\sethlcolor{color7}\hl{ find}\sethlcolor{color5}\hl{ the}\sethlcolor{color6}\hl{ y}\sethlcolor{color5}\hl{-}\sethlcolor{color2}\hl{inter}\sethlcolor{color6}\hl{cept}\sethlcolor{color6}\hl{ (}\sethlcolor{color6}\hl{b}\sethlcolor{color5}\hl{).}\sethlcolor{color3}\hl{ We}\sethlcolor{color4}\hl{'}\sethlcolor{color6}\hl{ll}\sethlcolor{color4}\hl{ use}\sethlcolor{color0}\hl{ (}\sethlcolor{color6}\hl{3}\sethlcolor{color4}\hl{,}\sethlcolor{color2}\hl{ }\sethlcolor{color2}\hl{ 5}\sethlcolor{color6}\hl{):}\sethlcolor{color4}\hl{ y}\sethlcolor{color4}\hl{ =}\sethlcolor{color7}\hl{ m}\sethlcolor{color3}\hl{x}\sethlcolor{color4}\hl{ +}\sethlcolor{color0}\hl{ b}\sethlcolor{color0}\hl{ }\sethlcolor{color7}\hl{ 5}\sethlcolor{color4}\hl{ =}\sethlcolor{color0}\hl{ }\sethlcolor{color6}\hl{ 1}\sethlcolor{color3}\hl{ *}\sethlcolor{color0}\hl{ }\sethlcolor{color6}\hl{ 3}\sethlcolor{color6}\hl{ +}\sethlcolor{color0}\hl{ b}\sethlcolor{color2}\hl{ }\sethlcolor{color1}\hl{ 5}\sethlcolor{color4}\hl{ =}\sethlcolor{color6}\hl{ }\sethlcolor{color7}\hl{ 3}\sethlcolor{color3}\hl{ +}\sethlcolor{color5}\hl{ b}\sethlcolor{color5}\hl{ b}\sethlcolor{color4}\hl{ =}\sethlcolor{color3}\hl{ }\sethlcolor{color1}\hl{ 2}\sethlcolor{color0}\hl{ }\sethlcolor{color6}\hl{ 3}\sethlcolor{color5}\hl{.}\sethlcolor{color4}\hl{ Write}\sethlcolor{color0}\hl{ the}\sethlcolor{color6}\hl{ equation}\sethlcolor{color0}\hl{ in}\sethlcolor{color7}\hl{ slope}\sethlcolor{color1}\hl{-}\sethlcolor{color0}\hl{inter}\sethlcolor{color7}\hl{cept}\sethlcolor{color1}\hl{ form}\sethlcolor{color3}\hl{:}\sethlcolor{color3}\hl{ y}\sethlcolor{color4}\hl{ =}\sethlcolor{color1}\hl{ m}\sethlcolor{color7}\hl{x}\sethlcolor{color6}\hl{ +}\sethlcolor{color0}\hl{ b}\sethlcolor{color6}\hl{ y}\sethlcolor{color4}\hl{ =}\sethlcolor{color0}\hl{ }\sethlcolor{color4}\hl{ 1}\sethlcolor{color3}\hl{x}\sethlcolor{color4}\hl{ +}\sethlcolor{color3}\hl{ }\sethlcolor{color1}\hl{ 2}\sethlcolor{color5}\hl{ y}\sethlcolor{color4}\hl{ =}\sethlcolor{color7}\hl{ x}\sethlcolor{color7}\hl{ +}\sethlcolor{color4}\hl{ }\sethlcolor{color6}\hl{ 2}\sethlcolor{color6}\hl{ The}\sethlcolor{color6}\hl{ equation}\sethlcolor{color1}\hl{ of}\sethlcolor{color5}\hl{ the}\sethlcolor{color7}\hl{ line}\sethlcolor{color1}\hl{ passing}\sethlcolor{color6}\hl{ through}\sethlcolor{color3}\hl{ the}\sethlcolor{color1}\hl{ points}\sethlcolor{color6}\hl{ (}\sethlcolor{color7}\hl{3}\sethlcolor{color3}\hl{,}\sethlcolor{color3}\hl{ }\sethlcolor{color6}\hl{ 5}\sethlcolor{color5}\hl{)}\sethlcolor{color3}\hl{ and}\sethlcolor{color6}\hl{ (}\sethlcolor{color5}\hl{7}\sethlcolor{color2}\hl{,}\sethlcolor{color0}\hl{ }\sethlcolor{color7}\hl{ 9}\sethlcolor{color3}\hl{)}\sethlcolor{color7}\hl{ is}\sethlcolor{color6}\hl{ y}\sethlcolor{color7}\hl{ =}\sethlcolor{color7}\hl{ x}\sethlcolor{color2}\hl{ +}\sethlcolor{color3}\hl{ }\sethlcolor{color2}\hl{ 2}\sethlcolor{color3}\hl{.}
}
\\
\hline
MLP 16 &
{\setlength{\fboxsep}{0pt}
\sethlcolor{color2}\hl{\texttt{<s>}}\sethlcolor{color3}\hl{ Find}\sethlcolor{color6}\hl{ the}\sethlcolor{color1}\hl{ equation}\sethlcolor{color1}\hl{ of}\sethlcolor{color1}\hl{ the}\sethlcolor{color1}\hl{ line}\sethlcolor{color1}\hl{ passing}\sethlcolor{color3}\hl{ through}\sethlcolor{color1}\hl{ the}\sethlcolor{color2}\hl{ points}\sethlcolor{color3}\hl{ (}\sethlcolor{color1}\hl{3}\sethlcolor{color3}\hl{,}\sethlcolor{color3}\hl{ }\sethlcolor{color1}\hl{ 5}\sethlcolor{color3}\hl{)}\sethlcolor{color3}\hl{ and}\sethlcolor{color3}\hl{ (}\sethlcolor{color1}\hl{7}\sethlcolor{color1}\hl{,}\sethlcolor{color3}\hl{ }\sethlcolor{color1}\hl{ 9}\sethlcolor{color1}\hl{)}\sethlcolor{color3}\hl{ using}\sethlcolor{color1}\hl{ the}\sethlcolor{color2}\hl{ slope}\sethlcolor{color7}\hl{-}\sethlcolor{color7}\hl{inter}\sethlcolor{color7}\hl{cept}\sethlcolor{color7}\hl{ form}\sethlcolor{color3}\hl{ of}\sethlcolor{color1}\hl{ a}\sethlcolor{color7}\hl{ linear}\sethlcolor{color1}\hl{ equation}\sethlcolor{color2}\hl{.}\sethlcolor{color5}\hl{ To}\sethlcolor{color1}\hl{ find}\sethlcolor{color1}\hl{ the}\sethlcolor{color2}\hl{ equation}\sethlcolor{color1}\hl{ of}\sethlcolor{color1}\hl{ the}\sethlcolor{color2}\hl{ line}\sethlcolor{color1}\hl{ passing}\sethlcolor{color3}\hl{ through}\sethlcolor{color1}\hl{ the}\sethlcolor{color2}\hl{ points}\sethlcolor{color3}\hl{ (}\sethlcolor{color1}\hl{3}\sethlcolor{color1}\hl{,}\sethlcolor{color1}\hl{ }\sethlcolor{color1}\hl{ 5}\sethlcolor{color3}\hl{)}\sethlcolor{color3}\hl{ and}\sethlcolor{color3}\hl{ (}\sethlcolor{color7}\hl{7}\sethlcolor{color1}\hl{,}\sethlcolor{color1}\hl{ }\sethlcolor{color1}\hl{ 9}\sethlcolor{color1}\hl{)}\sethlcolor{color1}\hl{ using}\sethlcolor{color3}\hl{ the}\sethlcolor{color7}\hl{ slope}\sethlcolor{color7}\hl{-}\sethlcolor{color7}\hl{inter}\sethlcolor{color1}\hl{cept}\sethlcolor{color1}\hl{ form}\sethlcolor{color3}\hl{ (}\sethlcolor{color1}\hl{y}\sethlcolor{color1}\hl{ =}\sethlcolor{color1}\hl{ m}\sethlcolor{color3}\hl{x}\sethlcolor{color3}\hl{ +}\sethlcolor{color3}\hl{ b}\sethlcolor{color3}\hl{),}\sethlcolor{color3}\hl{ we}\sethlcolor{color3}\hl{ first}\sethlcolor{color5}\hl{ need}\sethlcolor{color5}\hl{ to}\sethlcolor{color5}\hl{ find}\sethlcolor{color3}\hl{ the}\sethlcolor{color3}\hl{ slope}\sethlcolor{color3}\hl{ (}\sethlcolor{color3}\hl{m}\sethlcolor{color3}\hl{)}\sethlcolor{color3}\hl{ and}\sethlcolor{color3}\hl{ the}\sethlcolor{color1}\hl{ y}\sethlcolor{color7}\hl{-}\sethlcolor{color7}\hl{inter}\sethlcolor{color3}\hl{cept}\sethlcolor{color3}\hl{ (}\sethlcolor{color1}\hl{b}\sethlcolor{color3}\hl{).}\sethlcolor{color3}\hl{ }\sethlcolor{color1}\hl{ 1}\sethlcolor{color2}\hl{.}\sethlcolor{color6}\hl{ Find}\sethlcolor{color1}\hl{ the}\sethlcolor{color3}\hl{ slope}\sethlcolor{color3}\hl{ (}\sethlcolor{color1}\hl{m}\sethlcolor{color3}\hl{):}\sethlcolor{color3}\hl{ m}\sethlcolor{color3}\hl{ =}\sethlcolor{color3}\hl{ (}\sethlcolor{color1}\hl{y}\sethlcolor{color1}\hl{2}\sethlcolor{color3}\hl{ -}\sethlcolor{color1}\hl{ y}\sethlcolor{color1}\hl{1}\sethlcolor{color3}\hl{)}\sethlcolor{color3}\hl{ /}\sethlcolor{color3}\hl{ (}\sethlcolor{color1}\hl{x}\sethlcolor{color1}\hl{2}\sethlcolor{color3}\hl{ -}\sethlcolor{color1}\hl{ x}\sethlcolor{color1}\hl{1}\sethlcolor{color3}\hl{)}\sethlcolor{color3}\hl{ m}\sethlcolor{color3}\hl{ =}\sethlcolor{color3}\hl{ (}\sethlcolor{color1}\hl{9}\sethlcolor{color3}\hl{ -}\sethlcolor{color3}\hl{ }\sethlcolor{color1}\hl{ 5}\sethlcolor{color3}\hl{)}\sethlcolor{color3}\hl{ /}\sethlcolor{color3}\hl{ (}\sethlcolor{color7}\hl{7}\sethlcolor{color3}\hl{ -}\sethlcolor{color3}\hl{ }\sethlcolor{color1}\hl{ 3}\sethlcolor{color3}\hl{)}\sethlcolor{color3}\hl{ m}\sethlcolor{color3}\hl{ =}\sethlcolor{color3}\hl{ }\sethlcolor{color1}\hl{ 4}\sethlcolor{color3}\hl{ /}\sethlcolor{color3}\hl{ }\sethlcolor{color1}\hl{ 4}\sethlcolor{color3}\hl{ m}\sethlcolor{color3}\hl{ =}\sethlcolor{color3}\hl{ }\sethlcolor{color1}\hl{ 1}\sethlcolor{color3}\hl{ }\sethlcolor{color1}\hl{ 2}\sethlcolor{color4}\hl{.}\sethlcolor{color3}\hl{ Use}\sethlcolor{color6}\hl{ one}\sethlcolor{color3}\hl{ of}\sethlcolor{color6}\hl{ the}\sethlcolor{color2}\hl{ points}\sethlcolor{color5}\hl{ to}\sethlcolor{color1}\hl{ find}\sethlcolor{color1}\hl{ the}\sethlcolor{color1}\hl{ y}\sethlcolor{color1}\hl{-}\sethlcolor{color7}\hl{inter}\sethlcolor{color2}\hl{cept}\sethlcolor{color3}\hl{ (}\sethlcolor{color1}\hl{b}\sethlcolor{color3}\hl{).}\sethlcolor{color3}\hl{ We}\sethlcolor{color1}\hl{'}\sethlcolor{color5}\hl{ll}\sethlcolor{color3}\hl{ use}\sethlcolor{color3}\hl{ (}\sethlcolor{color1}\hl{3}\sethlcolor{color3}\hl{,}\sethlcolor{color3}\hl{ }\sethlcolor{color1}\hl{ 5}\sethlcolor{color3}\hl{):}\sethlcolor{color2}\hl{ y}\sethlcolor{color3}\hl{ =}\sethlcolor{color1}\hl{ m}\sethlcolor{color1}\hl{x}\sethlcolor{color3}\hl{ +}\sethlcolor{color3}\hl{ b}\sethlcolor{color3}\hl{ }\sethlcolor{color1}\hl{ 5}\sethlcolor{color3}\hl{ =}\sethlcolor{color3}\hl{ }\sethlcolor{color1}\hl{ 1}\sethlcolor{color3}\hl{ *}\sethlcolor{color3}\hl{ }\sethlcolor{color1}\hl{ 3}\sethlcolor{color3}\hl{ +}\sethlcolor{color3}\hl{ b}\sethlcolor{color3}\hl{ }\sethlcolor{color1}\hl{ 5}\sethlcolor{color3}\hl{ =}\sethlcolor{color3}\hl{ }\sethlcolor{color1}\hl{ 3}\sethlcolor{color3}\hl{ +}\sethlcolor{color2}\hl{ b}\sethlcolor{color3}\hl{ b}\sethlcolor{color3}\hl{ =}\sethlcolor{color3}\hl{ }\sethlcolor{color1}\hl{ 2}\sethlcolor{color3}\hl{ }\sethlcolor{color1}\hl{ 3}\sethlcolor{color3}\hl{.}\sethlcolor{color6}\hl{ Write}\sethlcolor{color1}\hl{ the}\sethlcolor{color6}\hl{ equation}\sethlcolor{color6}\hl{ in}\sethlcolor{color7}\hl{ slope}\sethlcolor{color7}\hl{-}\sethlcolor{color1}\hl{inter}\sethlcolor{color1}\hl{cept}\sethlcolor{color3}\hl{ form}\sethlcolor{color3}\hl{:}\sethlcolor{color1}\hl{ y}\sethlcolor{color3}\hl{ =}\sethlcolor{color3}\hl{ m}\sethlcolor{color1}\hl{x}\sethlcolor{color3}\hl{ +}\sethlcolor{color3}\hl{ b}\sethlcolor{color7}\hl{ y}\sethlcolor{color3}\hl{ =}\sethlcolor{color3}\hl{ }\sethlcolor{color1}\hl{ 1}\sethlcolor{color1}\hl{x}\sethlcolor{color3}\hl{ +}\sethlcolor{color3}\hl{ }\sethlcolor{color1}\hl{ 2}\sethlcolor{color4}\hl{ y}\sethlcolor{color3}\hl{ =}\sethlcolor{color1}\hl{ x}\sethlcolor{color1}\hl{ +}\sethlcolor{color3}\hl{ }\sethlcolor{color1}\hl{ 2}\sethlcolor{color3}\hl{ The}\sethlcolor{color2}\hl{ equation}\sethlcolor{color1}\hl{ of}\sethlcolor{color1}\hl{ the}\sethlcolor{color1}\hl{ line}\sethlcolor{color1}\hl{ passing}\sethlcolor{color3}\hl{ through}\sethlcolor{color1}\hl{ the}\sethlcolor{color1}\hl{ points}\sethlcolor{color3}\hl{ (}\sethlcolor{color1}\hl{3}\sethlcolor{color3}\hl{,}\sethlcolor{color1}\hl{ }\sethlcolor{color1}\hl{ 5}\sethlcolor{color3}\hl{)}\sethlcolor{color3}\hl{ and}\sethlcolor{color3}\hl{ (}\sethlcolor{color1}\hl{7}\sethlcolor{color7}\hl{,}\sethlcolor{color3}\hl{ }\sethlcolor{color1}\hl{ 9}\sethlcolor{color1}\hl{)}\sethlcolor{color3}\hl{ is}\sethlcolor{color1}\hl{ y}\sethlcolor{color3}\hl{ =}\sethlcolor{color1}\hl{ x}\sethlcolor{color3}\hl{ +}\sethlcolor{color3}\hl{ }\sethlcolor{color1}\hl{ 2}\sethlcolor{color3}\hl{.}
}
\\
\hline

MLP 32 &  
{\setlength{\fboxsep}{0pt}
\sethlcolor{color2}\hl{\texttt{<s>}}\sethlcolor{color6}\hl{ Find}\sethlcolor{color5}\hl{ the}\sethlcolor{color3}\hl{ equation}\sethlcolor{color5}\hl{ of}\sethlcolor{color5}\hl{ the}\sethlcolor{color6}\hl{ line}\sethlcolor{color6}\hl{ passing}\sethlcolor{color5}\hl{ through}\sethlcolor{color5}\hl{ the}\sethlcolor{color3}\hl{ points}\sethlcolor{color1}\hl{ (}\sethlcolor{color1}\hl{3}\sethlcolor{color7}\hl{,}\sethlcolor{color1}\hl{ }\sethlcolor{color7}\hl{ 5}\sethlcolor{color3}\hl{)}\sethlcolor{color3}\hl{ and}\sethlcolor{color1}\hl{ (}\sethlcolor{color7}\hl{7}\sethlcolor{color7}\hl{,}\sethlcolor{color1}\hl{ }\sethlcolor{color7}\hl{ 9}\sethlcolor{color3}\hl{)}\sethlcolor{color6}\hl{ using}\sethlcolor{color5}\hl{ the}\sethlcolor{color5}\hl{ slope}\sethlcolor{color1}\hl{-}\sethlcolor{color7}\hl{inter}\sethlcolor{color7}\hl{cept}\sethlcolor{color3}\hl{ form}\sethlcolor{color6}\hl{ of}\sethlcolor{color6}\hl{ a}\sethlcolor{color5}\hl{ linear}\sethlcolor{color3}\hl{ equation}\sethlcolor{color0}\hl{.}\sethlcolor{color4}\hl{ To}\sethlcolor{color5}\hl{ find}\sethlcolor{color5}\hl{ the}\sethlcolor{color3}\hl{ equation}\sethlcolor{color6}\hl{ of}\sethlcolor{color6}\hl{ the}\sethlcolor{color6}\hl{ line}\sethlcolor{color2}\hl{ passing}\sethlcolor{color2}\hl{ through}\sethlcolor{color5}\hl{ the}\sethlcolor{color3}\hl{ points}\sethlcolor{color1}\hl{ (}\sethlcolor{color7}\hl{3}\sethlcolor{color7}\hl{,}\sethlcolor{color1}\hl{ }\sethlcolor{color7}\hl{ 5}\sethlcolor{color2}\hl{)}\sethlcolor{color0}\hl{ and}\sethlcolor{color1}\hl{ (}\sethlcolor{color7}\hl{7}\sethlcolor{color7}\hl{,}\sethlcolor{color6}\hl{ }\sethlcolor{color7}\hl{ 9}\sethlcolor{color7}\hl{)}\sethlcolor{color6}\hl{ using}\sethlcolor{color6}\hl{ the}\sethlcolor{color5}\hl{ slope}\sethlcolor{color1}\hl{-}\sethlcolor{color7}\hl{inter}\sethlcolor{color5}\hl{cept}\sethlcolor{color3}\hl{ form}\sethlcolor{color1}\hl{ (}\sethlcolor{color1}\hl{y}\sethlcolor{color5}\hl{ =}\sethlcolor{color5}\hl{ m}\sethlcolor{color7}\hl{x}\sethlcolor{color7}\hl{ +}\sethlcolor{color7}\hl{ b}\sethlcolor{color4}\hl{),}\sethlcolor{color4}\hl{ we}\sethlcolor{color4}\hl{ first}\sethlcolor{color6}\hl{ need}\sethlcolor{color4}\hl{ to}\sethlcolor{color5}\hl{ find}\sethlcolor{color5}\hl{ the}\sethlcolor{color3}\hl{ slope}\sethlcolor{color1}\hl{ (}\sethlcolor{color7}\hl{m}\sethlcolor{color3}\hl{)}\sethlcolor{color6}\hl{ and}\sethlcolor{color5}\hl{ the}\sethlcolor{color1}\hl{ y}\sethlcolor{color1}\hl{-}\sethlcolor{color7}\hl{inter}\sethlcolor{color3}\hl{cept}\sethlcolor{color1}\hl{ (}\sethlcolor{color7}\hl{b}\sethlcolor{color7}\hl{).}\sethlcolor{color1}\hl{ }\sethlcolor{color7}\hl{ 1}\sethlcolor{color1}\hl{.}\sethlcolor{color3}\hl{ Find}\sethlcolor{color5}\hl{ the}\sethlcolor{color3}\hl{ slope}\sethlcolor{color1}\hl{ (}\sethlcolor{color7}\hl{m}\sethlcolor{color1}\hl{):}\sethlcolor{color1}\hl{ m}\sethlcolor{color1}\hl{ =}\sethlcolor{color1}\hl{ (}\sethlcolor{color7}\hl{y}\sethlcolor{color7}\hl{2}\sethlcolor{color7}\hl{ -}\sethlcolor{color7}\hl{ y}\sethlcolor{color7}\hl{1}\sethlcolor{color2}\hl{)}\sethlcolor{color7}\hl{ /}\sethlcolor{color1}\hl{ (}\sethlcolor{color7}\hl{x}\sethlcolor{color7}\hl{2}\sethlcolor{color7}\hl{ -}\sethlcolor{color7}\hl{ x}\sethlcolor{color7}\hl{1}\sethlcolor{color7}\hl{)}\sethlcolor{color7}\hl{ m}\sethlcolor{color7}\hl{ =}\sethlcolor{color1}\hl{ (}\sethlcolor{color7}\hl{9}\sethlcolor{color7}\hl{ -}\sethlcolor{color7}\hl{ }\sethlcolor{color7}\hl{ 5}\sethlcolor{color7}\hl{)}\sethlcolor{color7}\hl{ /}\sethlcolor{color7}\hl{ (}\sethlcolor{color7}\hl{7}\sethlcolor{color7}\hl{ -}\sethlcolor{color7}\hl{ }\sethlcolor{color7}\hl{ 3}\sethlcolor{color7}\hl{)}\sethlcolor{color7}\hl{ m}\sethlcolor{color7}\hl{ =}\sethlcolor{color1}\hl{ }\sethlcolor{color7}\hl{ 4}\sethlcolor{color7}\hl{ /}\sethlcolor{color7}\hl{ }\sethlcolor{color7}\hl{ 4}\sethlcolor{color2}\hl{ m}\sethlcolor{color7}\hl{ =}\sethlcolor{color1}\hl{ }\sethlcolor{color7}\hl{ 1}\sethlcolor{color1}\hl{ }\sethlcolor{color7}\hl{ 2}\sethlcolor{color7}\hl{.}\sethlcolor{color5}\hl{ Use}\sethlcolor{color5}\hl{ one}\sethlcolor{color4}\hl{ of}\sethlcolor{color5}\hl{ the}\sethlcolor{color2}\hl{ points}\sethlcolor{color4}\hl{ to}\sethlcolor{color3}\hl{ find}\sethlcolor{color5}\hl{ the}\sethlcolor{color1}\hl{ y}\sethlcolor{color1}\hl{-}\sethlcolor{color7}\hl{inter}\sethlcolor{color3}\hl{cept}\sethlcolor{color1}\hl{ (}\sethlcolor{color7}\hl{b}\sethlcolor{color7}\hl{).}\sethlcolor{color6}\hl{ We}\sethlcolor{color7}\hl{'}\sethlcolor{color4}\hl{ll}\sethlcolor{color5}\hl{ use}\sethlcolor{color1}\hl{ (}\sethlcolor{color2}\hl{3}\sethlcolor{color7}\hl{,}\sethlcolor{color1}\hl{ }\sethlcolor{color7}\hl{ 5}\sethlcolor{color1}\hl{):}\sethlcolor{color1}\hl{ y}\sethlcolor{color5}\hl{ =}\sethlcolor{color1}\hl{ m}\sethlcolor{color7}\hl{x}\sethlcolor{color7}\hl{ +}\sethlcolor{color1}\hl{ b}\sethlcolor{color1}\hl{ }\sethlcolor{color7}\hl{ 5}\sethlcolor{color2}\hl{ =}\sethlcolor{color1}\hl{ }\sethlcolor{color7}\hl{ 1}\sethlcolor{color7}\hl{ *}\sethlcolor{color2}\hl{ }\sethlcolor{color7}\hl{ 3}\sethlcolor{color7}\hl{ +}\sethlcolor{color7}\hl{ b}\sethlcolor{color1}\hl{ }\sethlcolor{color7}\hl{ 5}\sethlcolor{color7}\hl{ =}\sethlcolor{color1}\hl{ }\sethlcolor{color7}\hl{ 3}\sethlcolor{color7}\hl{ +}\sethlcolor{color7}\hl{ b}\sethlcolor{color2}\hl{ b}\sethlcolor{color7}\hl{ =}\sethlcolor{color1}\hl{ }\sethlcolor{color7}\hl{ 2}\sethlcolor{color7}\hl{ }\sethlcolor{color7}\hl{ 3}\sethlcolor{color7}\hl{.}\sethlcolor{color3}\hl{ Write}\sethlcolor{color5}\hl{ the}\sethlcolor{color3}\hl{ equation}\sethlcolor{color5}\hl{ in}\sethlcolor{color5}\hl{ slope}\sethlcolor{color1}\hl{-}\sethlcolor{color7}\hl{inter}\sethlcolor{color5}\hl{cept}\sethlcolor{color3}\hl{ form}\sethlcolor{color1}\hl{:}\sethlcolor{color6}\hl{ y}\sethlcolor{color6}\hl{ =}\sethlcolor{color5}\hl{ m}\sethlcolor{color7}\hl{x}\sethlcolor{color7}\hl{ +}\sethlcolor{color1}\hl{ b}\sethlcolor{color6}\hl{ y}\sethlcolor{color2}\hl{ =}\sethlcolor{color1}\hl{ }\sethlcolor{color7}\hl{ 1}\sethlcolor{color7}\hl{x}\sethlcolor{color7}\hl{ +}\sethlcolor{color1}\hl{ }\sethlcolor{color7}\hl{ 2}\sethlcolor{color2}\hl{ y}\sethlcolor{color2}\hl{ =}\sethlcolor{color7}\hl{ x}\sethlcolor{color7}\hl{ +}\sethlcolor{color7}\hl{ }\sethlcolor{color7}\hl{ 2}\sethlcolor{color5}\hl{ The}\sethlcolor{color2}\hl{ equation}\sethlcolor{color6}\hl{ of}\sethlcolor{color6}\hl{ the}\sethlcolor{color6}\hl{ line}\sethlcolor{color6}\hl{ passing}\sethlcolor{color2}\hl{ through}\sethlcolor{color6}\hl{ the}\sethlcolor{color2}\hl{ points}\sethlcolor{color1}\hl{ (}\sethlcolor{color7}\hl{3}\sethlcolor{color0}\hl{,}\sethlcolor{color1}\hl{ }\sethlcolor{color7}\hl{ 5}\sethlcolor{color2}\hl{)}\sethlcolor{color0}\hl{ and}\sethlcolor{color1}\hl{ (}\sethlcolor{color0}\hl{7}\sethlcolor{color0}\hl{,}\sethlcolor{color1}\hl{ }\sethlcolor{color7}\hl{ 9}\sethlcolor{color2}\hl{)}\sethlcolor{color6}\hl{ is}\sethlcolor{color1}\hl{ y}\sethlcolor{color7}\hl{ =}\sethlcolor{color7}\hl{ x}\sethlcolor{color7}\hl{ +}\sethlcolor{color7}\hl{ }\sethlcolor{color5}\hl{ 2}\sethlcolor{color1}\hl{.}
}
\\
\hline
\end{tabular}
}
\end{table} 
\section{More Experimental Results\label{sec:app-exp}}
\begin{figure}[t]
	\centering
    \includegraphics[width=.9\textwidth]{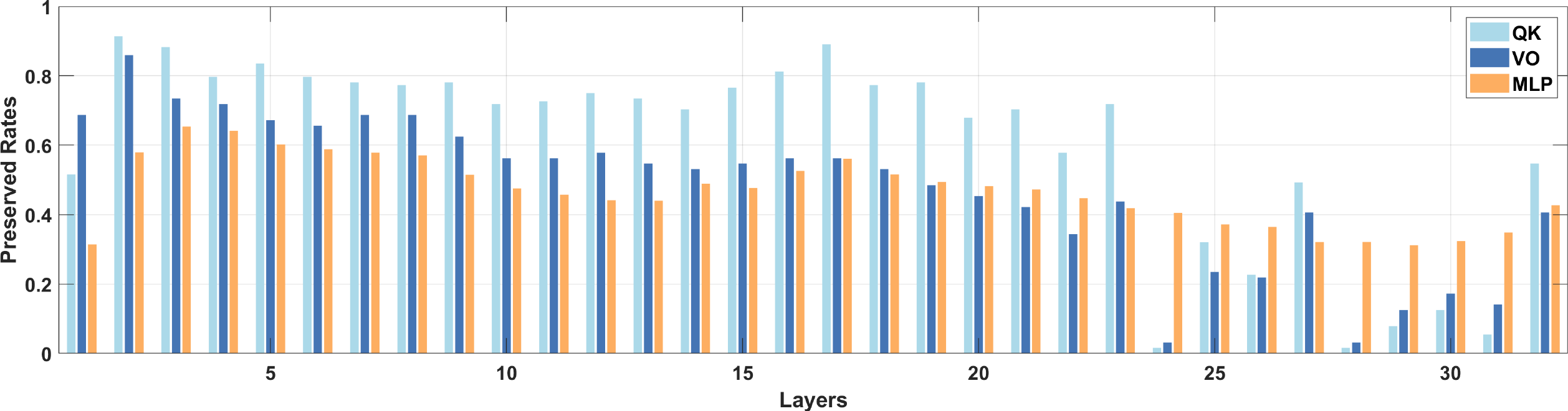}
    \caption{Model width after ToMoE for the LLaMA-2 7B model when the number of active parameters equals 50\%.}
  \label{fig:llama-width}
  %\vspace{-10pt}
\end{figure}

\begin{figure}[t]
	\centering
    \includegraphics[width=.9\textwidth]{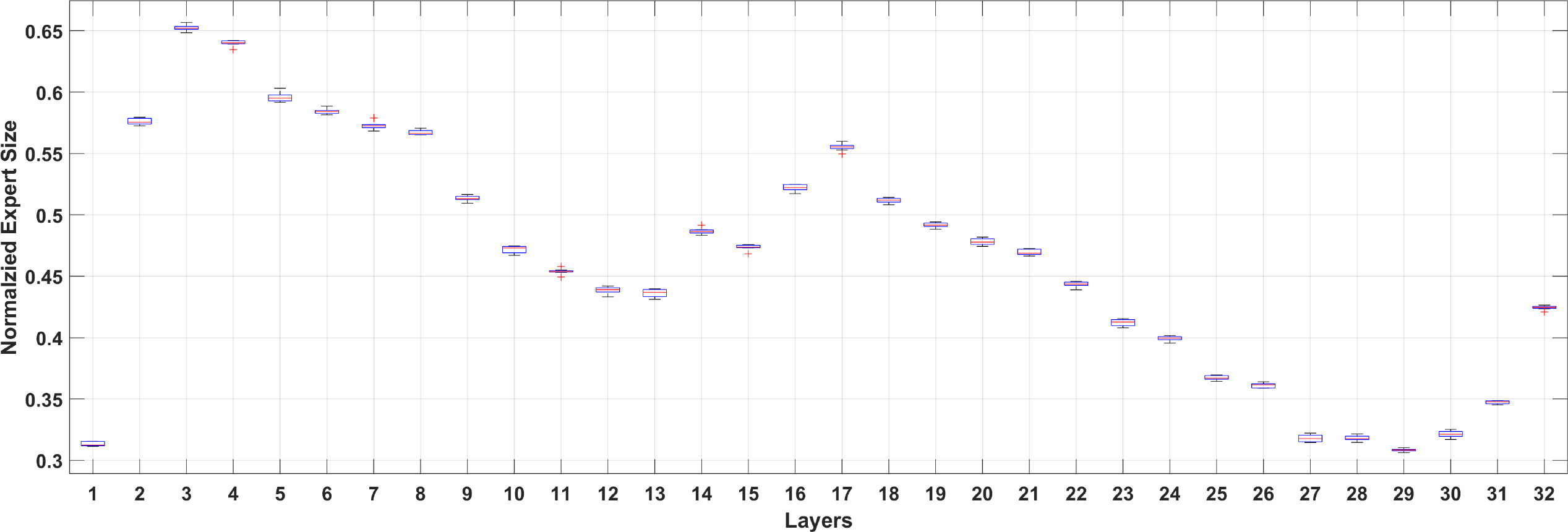}
    \caption{Box plot of widths across different experts for the LLaMA-2 7B model when the number of active parameters equals 50\%.}
  \label{fig:llama2-box}
  %\vspace{-10pt}
\end{figure}

\begin{figure}[t]
    \centering
    \includegraphics[width=.9\textwidth]{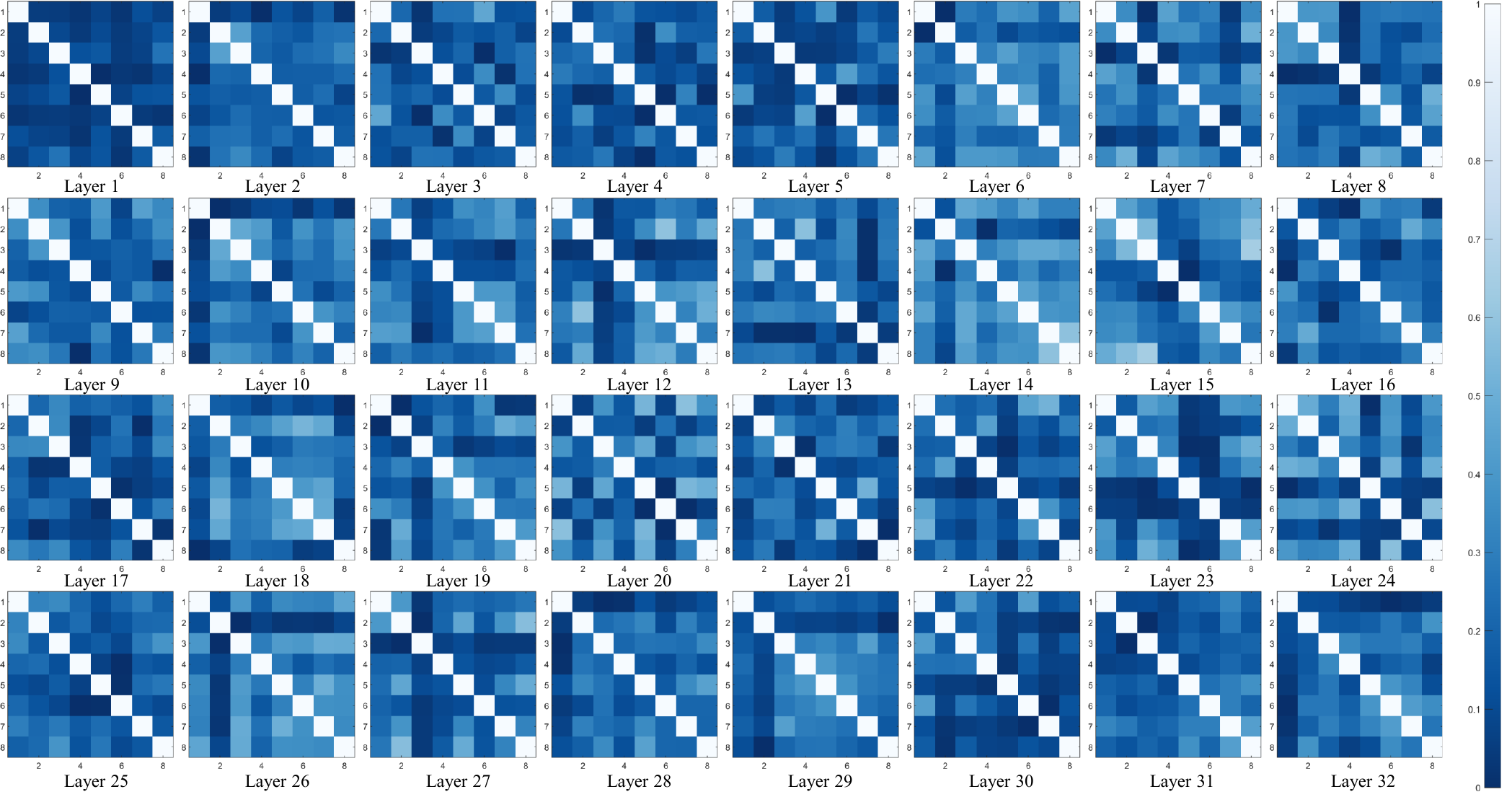}
    \caption{The similarity of different experts from different layers of ToMoE of the LLaMA-2 7B model.}
    \label{fig:expert_similarity}
    \vspace{-10pt}
\end{figure}

% In Tab.~\ref{tbl:zero_shot_large} and Tab.~\ref{tbl:zero_shot_phi2}, we present the zero-shot performance compared to different methods on LLaMA-2 13B, Qwen-2.5 14B, and Phi-2 models. From Tab.~\ref{tbl:zero_shot_large}, we can see that ToMoE consistently outperforms other comparison methods. Compared to 7B or 8B models, the gap between our methods and other comparison methods is smaller, which also applies to the gap between baseline methods, probably because the model size is larger. On the LLaMA-2 13B model, ToMoE further outperforms structural pruning methods with smaller compression rates, for example, ToMoE with 50\% active parameters performs better than MoDeGPT and LLM Surgeon with 40\% compression rate. The gap for the same number of active parameters is more obvious. From Tab.~\ref{tbl:zero_shot_phi2}, we can see that ToMoE performs better than SliceGPT and DISP-LLM. Specifically, ToMoE with 70\% active parameters achieves better performance than all three compression levels of SliceGPT and DISP-LLM. In Fig.~\ref{fig:llama-width}, we present the width of ToMoE for the LLaMA-2 7B model, we observe a highly non-uniform pattern when allocating the number of active parameters, which suggests that ToMoE can identify the ideal way to allocate active parameters even they could be highly non-uniform.

% These results again demonstrate the effectiveness of ToMoE in keeping the capacity of LLMs compared to structural pruning methods. Also, they show that ToMoE is effective across different scales of different LLMs. 

\begin{table}[h!]
\centering
\resizebox{0.85\textwidth}{!}{%
\begin{tabular}{lcccc}
\toprule
\textbf{Model} & \textbf{Active Parameters} & \textbf{Batch Size 512} & \textbf{Batch Size 1024} & \textbf{Batch Size 1536} \\
\midrule
LLaMA-2 7B      & 100\% & 1781.76 tokens/s & 1805.80 tokens/s & 1858.45 tokens/s \\
%\cmidrule[1pt]{1-5}
LLaMA-MoE E8A2~\citep{zhu2024llama}  & \multirow{2}{*}{50\%} & 2685.35 tokens/s & 2796.07 tokens/s & 2887.98 tokens/s \\
ToMoE (ours)     &    & 2360.38 tokens/s & 2702.12 tokens/s & 2919.45 tokens/s \\
\bottomrule
\end{tabular}
}
\caption{\revise{Inference throughput (tokens per second) under different mini-batchsizes.}\label{tab:throughput_params}}
\end{table}

In Tab.~\ref{tbl:zero_shot_large} and Tab.~\ref{tbl:zero_shot_phi2}, we present the zero-shot performance of various methods on LLaMA-2 13B, Qwen-2.5 14B, and Phi-2 models. From Tab.~\ref{tbl:zero_shot_large}, it is evident that ToMoE consistently outperforms other methods. Compared to 7B or 8B models, the performance gap between our method and other approaches is smaller, which also holds for the differences between baseline methods. This is likely due to the larger model sizes. Table~\ref{tbl:zero_shot_llama2} presents the zero-shot performance of the LLaMA-2 7B model across more baselines and active parameters. As shown in the table, ToMoE consistently achieves significantly better performance than all competing methods.

On the LLaMA-2 13B model, ToMoE surpasses structural pruning methods even with smaller compression rates. For instance, ToMoE with 50\% active parameters performs better than MoDeGPT and LLM Surgeon with a 40\% compression rate. The performance gap becomes even more obvious when comparing methods with the same number of active parameters. Similarly, from Tab.~\ref{tbl:zero_shot_phi2}, ToMoE demonstrates superior performance compared to SliceGPT and DISP-LLM. Specifically, ToMoE with 70\% active parameters achieves better results than all three compression levels of SliceGPT and DISP-LLM.

\begin{wrapfigure}{r}{0.25\textwidth}
    \centering
    \vspace{-\intextsep} % optional: tighten vertical space
    \includegraphics[width=\linewidth]{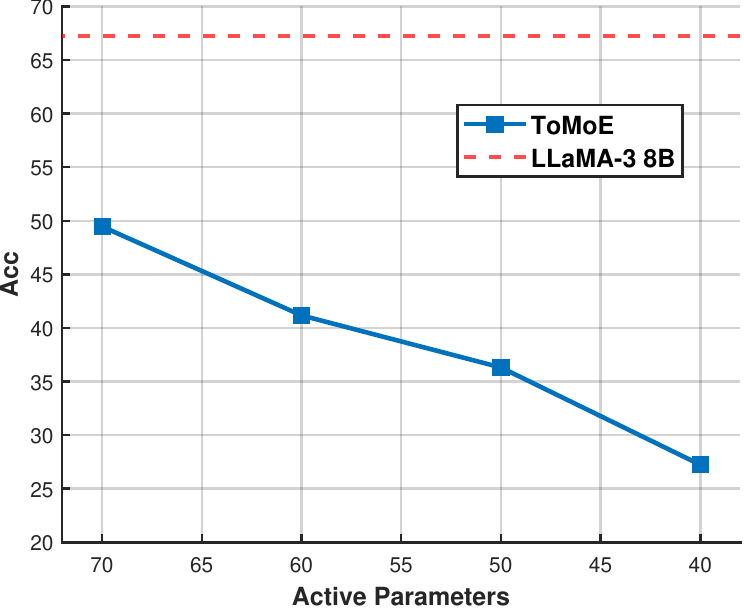} % or path/to/mmlu.pdf
    \caption{\revise{MMLU accuracy vs. active parameters.}}
    \label{fig:mmlu}
    \vspace{-\intextsep} % optional: tighten vertical space
\end{wrapfigure}

\revise{In Tab.~\ref{tab:throughput_params}, we report the inference throughput (measured in tokens per second) of different models under varying batch sizes. Compared to the dense {LLaMA-2 7B} baseline, both converted MoE models achieve higher throughput due to reduced active parameters. Our resulting model has a similar throughput to LLaMA-MoE when the batch size is large enough.}

\revise{In Fig.~\ref{fig:mmlu}, we present the accuracy–parameter trade-off on the more challenging MMLU dataset with the LLaMA-3 8B model. The results show that our method can still provide meaningful results when activating only 50\% of the model parameters.}

In Fig.~\ref{fig:llama-width}, we illustrate the width of ToMoE for the LLaMA-2 7B model. A highly non-uniform pattern emerges in the allocation of active parameters, indicating that ToMoE can effectively determine the ideal distribution of active parameters, even when the allocation is highly non-uniform.

In Tab.~\ref{tab:tomoe_design_choices}, we present an ablation study to analyze several design choices in ToMoE, focusing on architectural components and the effect of the temperature parameter $\tau$. Although ToMoE applies contextual sparsity to the value and output (VO) projections within the attention layer, their contribution to overall performance is relatively minor due to the small head dimension (128, in the case of LLaMA-2 7B). To validate this, we disable dynamic attention sparsity and instead apply only static structural pruning to the attention layer (denoted as "w/o VO routing"). This leads to only a modest performance drop of about 1\% using 50\% active parameters.

We also examine the role of global expert embeddings with GRU in conveying cross-layer architectural information. Specifically, we compare the default global expert embedding with a local-only variant ("local emb"), where expert embeddings are used only in MLP layers and removed from attention layers. Results show a slight decrease in performance, suggesting that global expert embeddings contribute to better coordination across layers.

Additionally, we evaluate the sensitivity of ToMoE to the temperature $\tau$ in the routing mechanism. The results with $\tau \in \{0.3, 0.4, 0.5\}$ show that performance remains relatively stable, indicating robustness to the choice of $\tau$ within a reasonable range.

Finally, we explore head pruning in the early stages of ToMoE development. However, this approach yielded significantly lower performance. This may be due to the distortion of attention feature maps when heads are removed, which makes it more difficult to train effective MLP experts.

These results highlight the effectiveness of ToMoE in preserving the capacity of LLMs compared to structural pruning methods. Additionally, they demonstrate that ToMoE performs robustly across various scales and types of LLMs.

\section{Visualization of Experts}
% In this section, we study the properties of resulting experts from our method. In Tab.~\ref{tab:moe_sample} and Tab.~\ref{tab:moe_sample_math}, we visualize the routed tokens with experts across different layers and texts. In Tab.~\ref{tab:moe_sample}, there is not a very clear pattern regarding different experts, which complies with our observation in Tba.~\ref{tab:gen_sample}. In addition to the semantic meanings of different experts, another interesting observation from Tab.~\ref{tab:moe_sample} and Tab.~\ref{tab:moe_sample_math} is that the first layer tends to have a more diverse allocation of tokens, and for later layers, there are more continuous tokens that are assigned to the same expert. In Tab.~\ref{tab:moe_sample_math}, the inputs belong to a math problem. Different from the visualization in Tab.~\ref{tab:moe_sample}, the MoE visualization of the math problem has a more obvious semantical meaning. For example,  \color{color2}{Expert 2} in MLP 16 is mainly active by numbers and math notations and a similar observation holds for \color{color8}{Expert 8} in MLP 32. This observation suggests that experts of ToMoE may have more clear semantic meanings than MoE trained from scratch. More studies are necessary to fully understand the exact semantic meanings of ToMoE experts. 
In this section, we analyze the properties of the experts produced by our method. Tab.~\ref{tab:moe_sample} and Tab.~\ref{tab:moe_sample_math} present visualizations of the routed tokens among experts across different layers and input texts.

In Tab.~\ref{tab:moe_sample}, we observe no distinct patterns in the allocation of tokens to specific experts, which aligns with our observations in Tab.~\ref{tab:gen_sample}. An interesting trend emerges when comparing layers: the first layer exhibits a more diverse token distribution, while subsequent layers prefer to assign continuous tokens to the same expert. Tab.~\ref{tab:moe_sample_math} focuses on inputs related to a math problem. Unlike the visualization in Tab.~\ref{tab:moe_sample}, the MoE routing for the math problem reveals clearer semantic patterns. For instance, \sethlcolor{color1}\hl{Expert 2} in MLP 16 is predominantly activated by numbers and mathematical notations, and a similar behavior is observed for \sethlcolor{color7}\hl{Expert 8} in MLP 32. This suggests that the experts in ToMoE may encode more distinct semantic meanings compared to MoE models trained from scratch. Further investigation is required to fully understand the precise semantic roles of ToMoE experts.

% In Fig.~\ref{fig:llama2-box}, we present the box plot of expert sizes of different layers. From the figure, we can see that the maximum expert size and minimum expert size are very close across different layers, which is a direct result of enforcing Eq.~\ref{eq:pr} and Eq.~\ref{eq:lb}. In addition, it is also because we penalize the largest expert in Eq.~\ref{eq:pr}. In the training process, minimizing the task loss (self-knowledge distillation loss) will encourage the expert to increase their sizes. As a result, a small expert will not remain small due to the task loss and is not penalized by the parameter regularization loss. By repeating this process, every expert will eventually have a similar size. After the ToMoE training process, we set the width of all experts to the maximum value of all experts to ensure each expert has the same computational cost.

In Fig.~\ref{fig:llama2-box}, we present a box plot showing the expert sizes across different layers. The figure reveals that the maximum and minimum expert sizes are closely aligned across layers. This outcome is a direct result of applying constraints from Eq.~\ref{eq:pr} and Eq.~\ref{eq:lb}, as well as only penalizing the largest expert in Eq.~\ref{eq:pr}. During training, minimizing the task loss (self-knowledge distillation loss) encourages experts to grow in size. Consequently, smaller experts do not remain small due to the task loss and they are not penalized by the parameter regularization loss. This iterative process leads to all experts eventually converging to similar sizes. After completing the ToMoE training process, we adjust the width of all experts to \textbf{match the maximum size among them}. This ensures uniform computational cost across all experts.

In Fig.~\ref{fig:expert_similarity}, we present a visualization of the similarity between different experts across all layers of the LLaMA-2 7B model. Within the same layer, experts generally exhibit comparable similarity values, indicating that while the experts share the same size, their weights remain distinct. Notably, certain layers, such as layer 1 and layer 30, show lower similarity values. This observation aligns with expectations, as the expert sizes in these layers are smaller.

\section{Design Choice for the MHA \label{sec:app-mha}}
Ideally, to achieve maximum flexibility, one might consider applying dynamic pruning to all projection matrices in the MHA layer, including the query ($W_Q$), key ($W_K$), value ($W_V$), and output ($W_O$) matrices. However, there is a fundamental limitation when attempting to apply dynamic pruning along the head dimension for the query and key matrices.

Suppose we generate pruning masks $S_t \in \{0, 1\}^d$ at each time step $t$ based on the input $X_t \in \mathbb{R}^{1 \times d}$, and consider two distinct time steps, $a$ and $b$. For the $i$-th attention head, the attention score between queries and keys is influenced by the pruning masks. Specifically, the effective attention score between the $a$-th query and the $a$-th key is given by:
\[
e(X_a W_{Q,i}) S_a S_a^\top e(X_a W_{K,i})^\top,
\]
while the attention score between the $a$-th query and the $b$-th key is:
\[
e(X_a W_{Q,i}) S_a S_b^\top e(X_b W_{K,i})^\top.
\]

The \emph{effective width}---that is, the dimensionality over which attention is computed—between the $a$-th query and the $a$-th key is $\|S_a S_a^\top\|_0 = \sum S_a = K$, assuming the mask has exactly $K$ active elements. However, for cross-position pairs like $(a, b)$, the effective width becomes $\|S_a S_b^\top\|_0 = \|S_a \odot S_b\|_0 \leq \min(\sum S_a, \sum S_b) = K$. The equality holds only when $S_a = S_b$, which generally does not hold for arbitrary $a \neq b$.

This observation implies that dynamically pruned query and key matrices fail to fully utilize the allocated capacity $K$ unless the pruning masks are identical across all positions. Moreover, the variability of the effective width across different query-key pairs introduces instability and inconsistent capacity utilization, making this approach less favorable compared to static pruning for the query and key matrices.

In contrast, dynamic pruning does not encounter this issue when applied to the value and output matrices, as these are not involved in pairwise comparisons like the query-key dot products. Therefore, we adopt dynamic pruning only for the value and output projections, while keeping the query and key projections pruned statically to maintain stable and full-capacity attention computation.

\end{document}